%% file: main.tex
\PassOptionsToPackage{table}{xcolor}
\documentclass{article}
\usepackage{style,times}

\usepackage{siunitx}
\usepackage{graphicx}
\usepackage{hyperref}
\usepackage{url}
\usepackage{tikz}

\usepackage{booktabs}       
\usepackage{amsfonts}       
\usepackage{nicefrac}       
\usepackage{microtype}      
\usepackage{xcolor}
\usepackage{amsmath}
\usepackage{enumitem}
\usepackage{xfrac}
\usepackage{tikz}
\usepackage{siunitx}
\usetikzlibrary{positioning}
\usepackage{graphicx}
\usepackage{subcaption}

\definecolor{mygreen}{RGB}{30,160,80}
\newcommand{\depth}{\textcolor{mygreen}{$\bullet\;$}}
\newcommand{\nodepth}{\textcolor{mygreen}{$\circ\;$}}

\title{
\includegraphics[height=5mm]{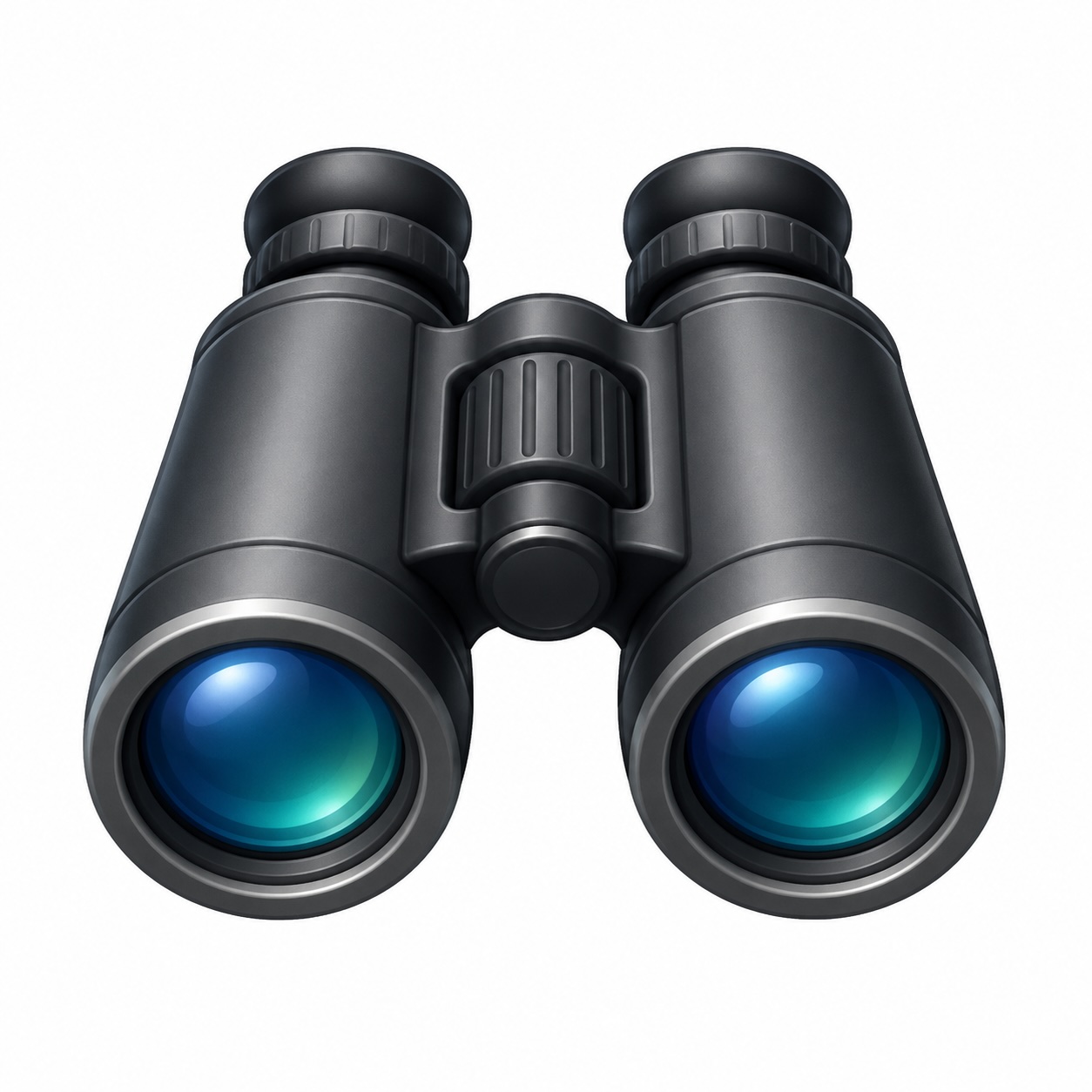}\includegraphics[height=5mm]{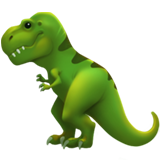}  DINOcular:\\Self-Supervised Visuospatial Representations
}

\author{Farkhat Almukhamedov\thanks{Indicates equal contribution.} ~, Sami Azirar\footnotemark[1] ~\& Hermann Blum \\
Robot Perception and Learning Lab\\
University of Bonn \& Lamarr Institute\\
\texttt{\{s94falmu,sazirar,blumh\}@uni-bonn.de}
}

\preprintfinalcopy
\begin{document}

\maketitle

\begin{abstract}
We introduce a self-supervised framework for learning joint visuospatial representations from RGB-D observations. While modern vision foundation models are trained almost exclusively on RGB images, many embodied systems have access to explicit depth sensing, which provides geometric information that monocular inputs cannot recover. Our method integrates depth-derived geometric priors with a visual backbone through inter-patch and intra-patch fusion, enabling the model to encode both appearance and spatial structure efficiently. The resulting representation shows promising improvements on 3D awareness while preserving semantic transfer: it outperforms prior methods of comparable scale on multiple 3D geometry benchmarks, and remains competitive when probed for standard RGB-D semantic segmentation tasks.
\textbf{Project page:} \url{https://heyleadro.github.io/dinocular-project/}
\end{abstract}

\section{Introduction}
Explicit depth sensing is a remarkably widespread biological solution for spatial perception: binocular stereopsis, which compares observations from two eyes to infer depth, has been demonstrated across diverse animal species, from primates and other mammals~\citep{heesy2009seeing} to falcons~\citep{fox1977stereopsis} and cuttlefish~\citep{feord2020cuttlefish}. In artificial embodied systems, however, the dominant vision models remain largely monocular. Despite the availability of stereo cameras and depth sensors in robots, autonomous vehicles, mixed-reality headsets, and consumer smartphones, modern vision foundation models are still trained almost exclusively on RGB images. This mismatch has left frontier visual AI systems with a persistent weakness in spatial and 3D understanding~\citep{feng2025seeing,el2024probing,man2024lexicon3d,azzolini2025cosmos}. Training with RGB inputs on more spatial tasks and data cannot fully solve this: Without depth as an input modality, spatial reasoning remains physically constrained by the scale ambiguity of monocular sensing.

Depth has historically been difficult to incorporate into foundation-scale vision learning for several reasons. Compared with RGB images, depth data is scarce, heterogeneous, and noisy: it may come from stereo, structured light, time-of-flight sensors, or sparse LiDAR, each with different failure modes and sampling patterns. At the same time, the field has become increasingly reliant on large RGB-only feature extractors such as DINO, making it natural to treat geometry as an output to be predicted rather than as an input to be encoded.
Recent progress in learned geometric reconstruction~\citep{wang2025vggt,keetha2026mapanything,wang2024dust3r} changes this picture. We can generate depth for training at very large scales, and we can homogenize different depth sources through completion and denoising into dense pixel-wise depth maps. These advances reopen a central question: how can models learn general-purpose representations that jointly encode visual appearance and geometric structure?

\begin{figure}[t]
    \centering
    \resizebox{\textwidth}{!}{%
    \begin{tikzpicture}[
        box/.style={
            inner sep=0pt,
            align=center
        }
    ]
    \node[box] (img1) {\includegraphics[width=3cm]{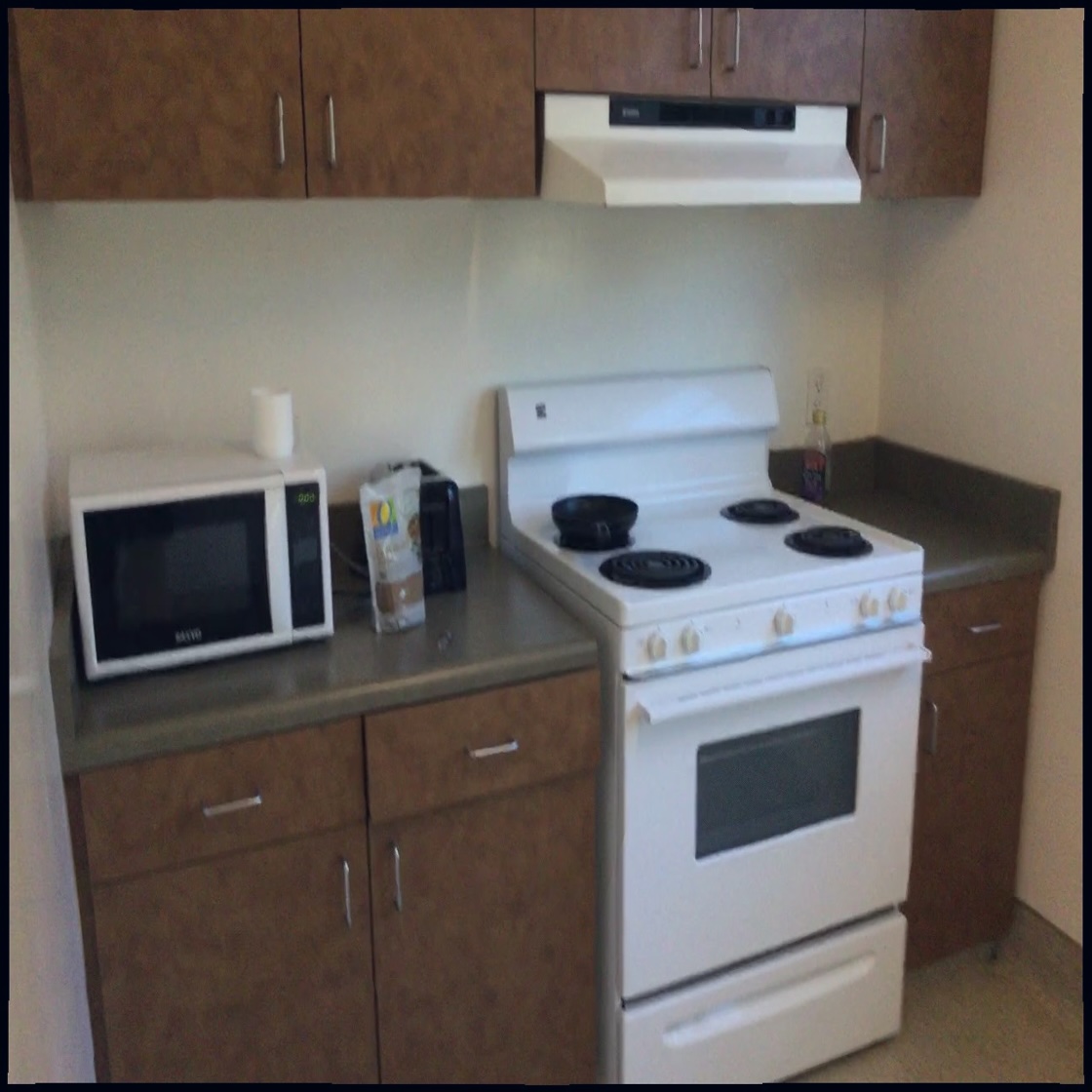}};
    \node[box, right=0cm of img1] (img2) {\includegraphics[width=3cm]{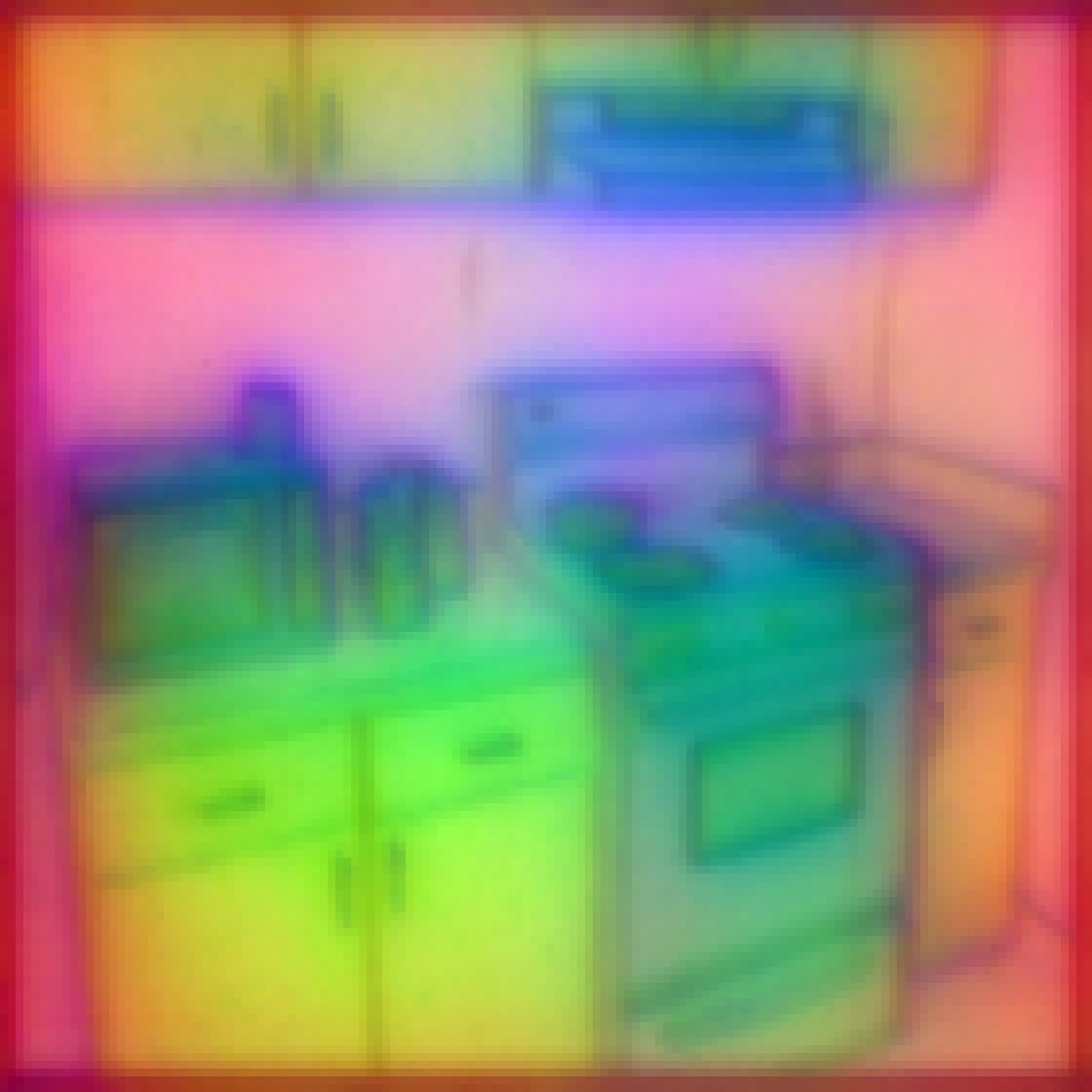}};
    \node[box, right=0cm of img2] (img3) {\includegraphics[width=3cm]{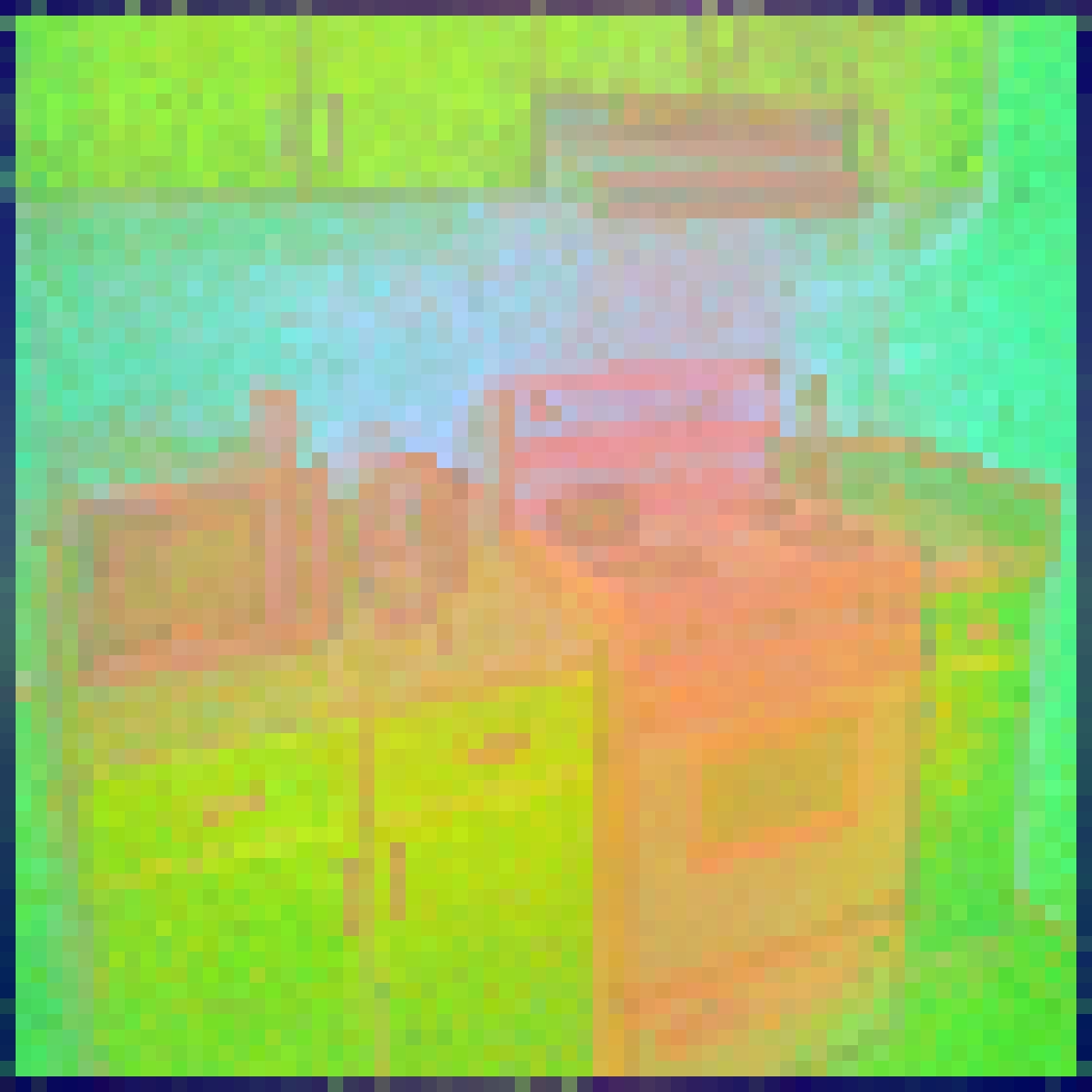}};
    \node[box, right=0cm of img3] (img4) {\includegraphics[width=3cm]{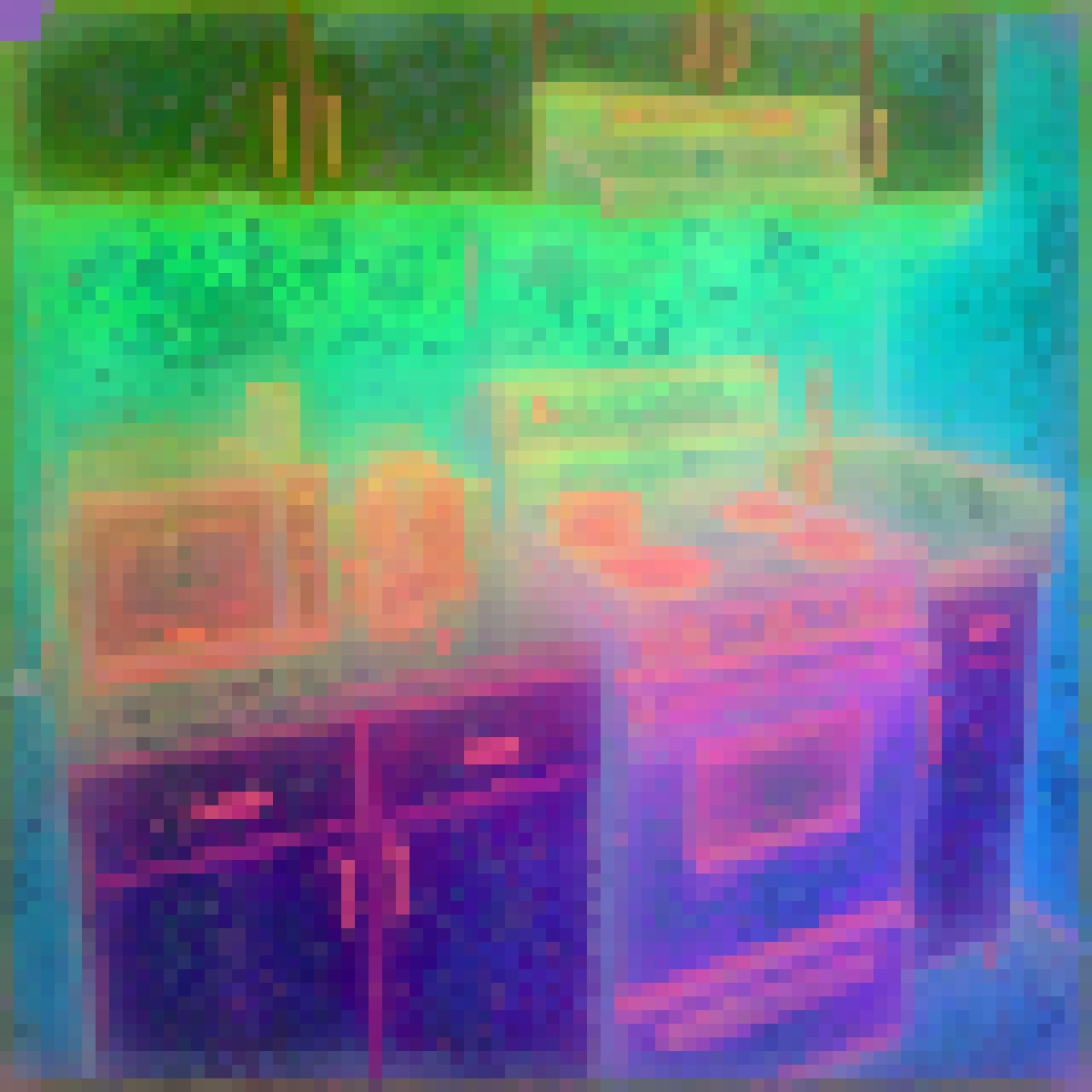}};

    \node[box, below=0cm of img1] (img6) {\includegraphics[width=3cm]{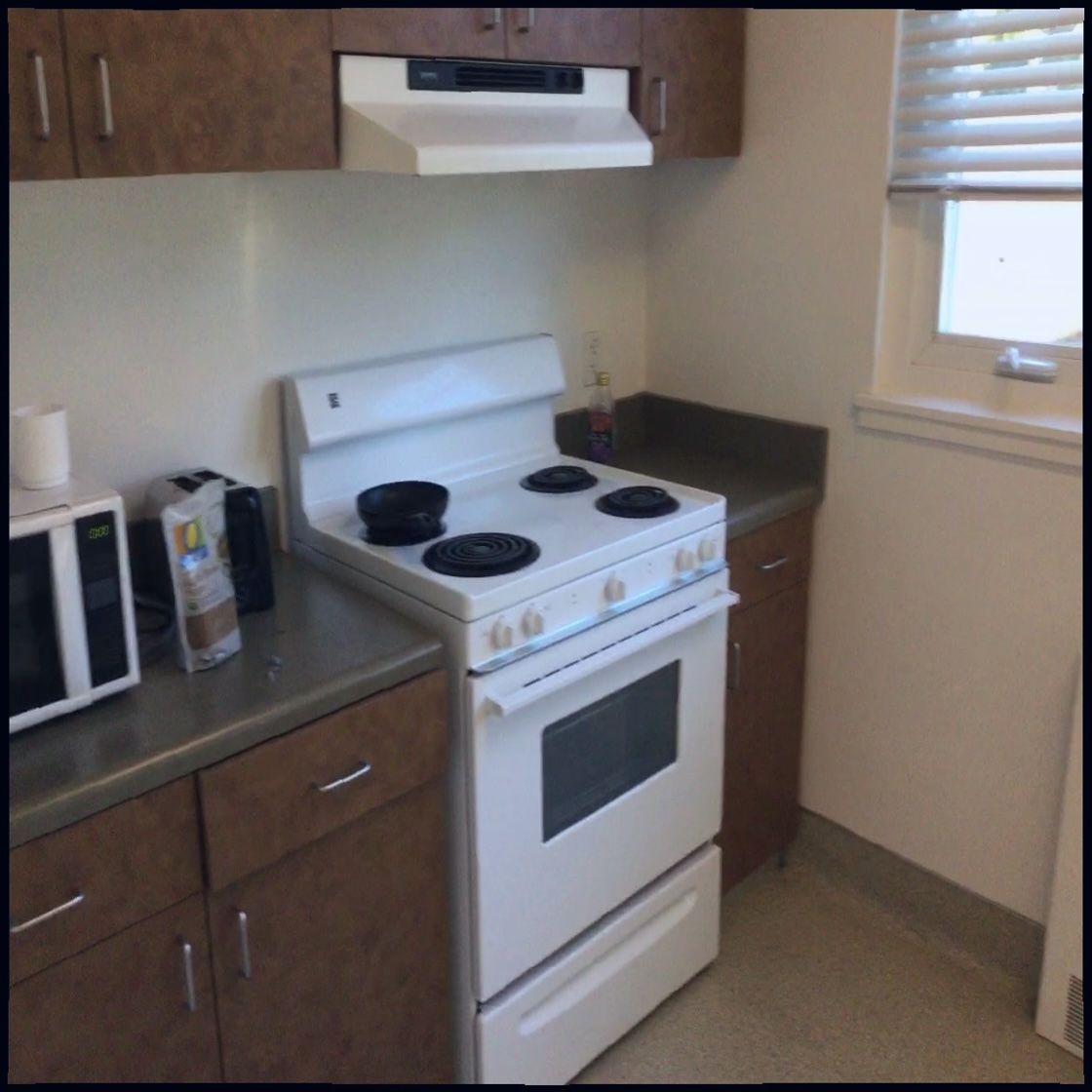}};
    \node[box, right=0cm of img6] (img7) {\includegraphics[width=3cm]{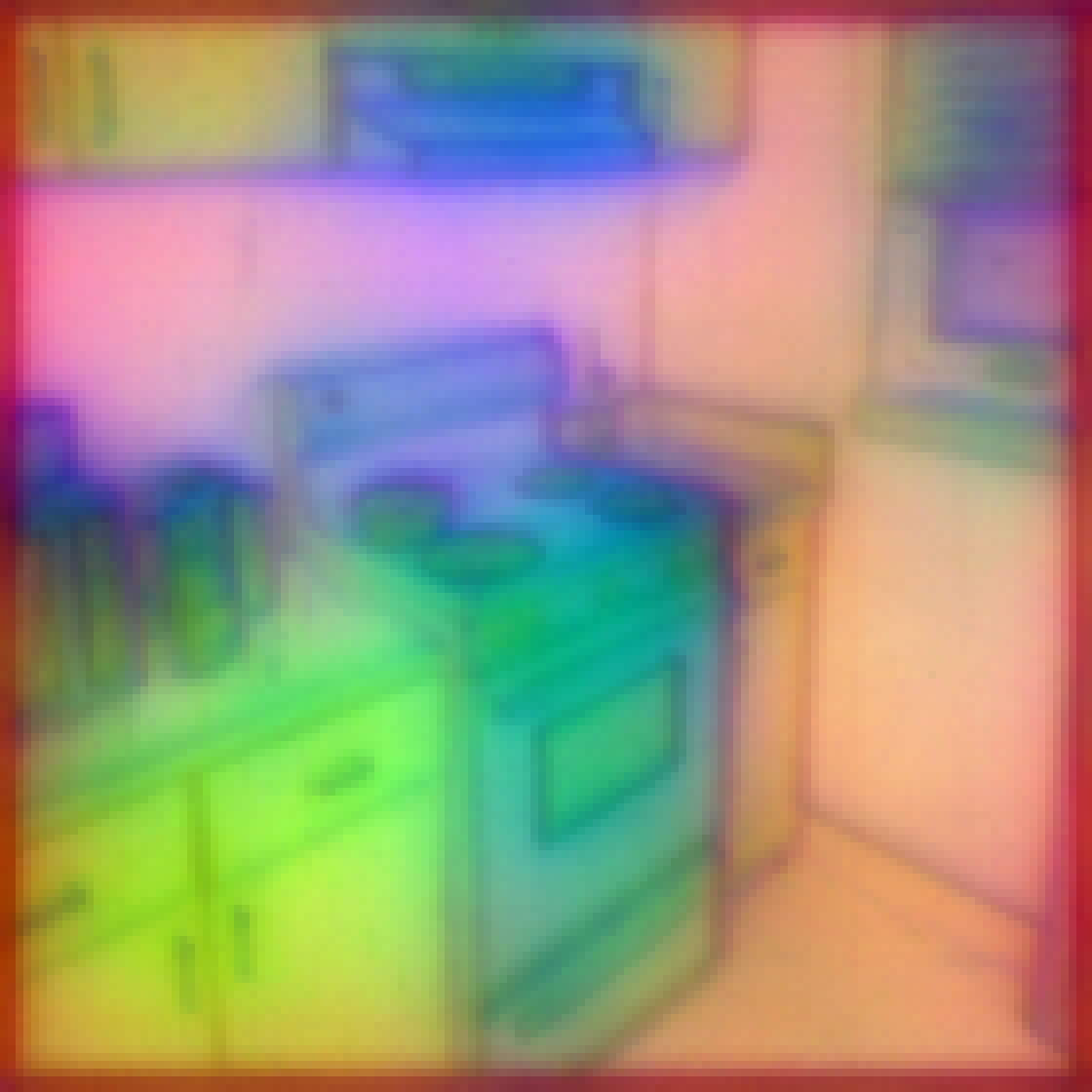}};
    \node[box, right=0cm of img7] (img8) {\includegraphics[width=3cm]{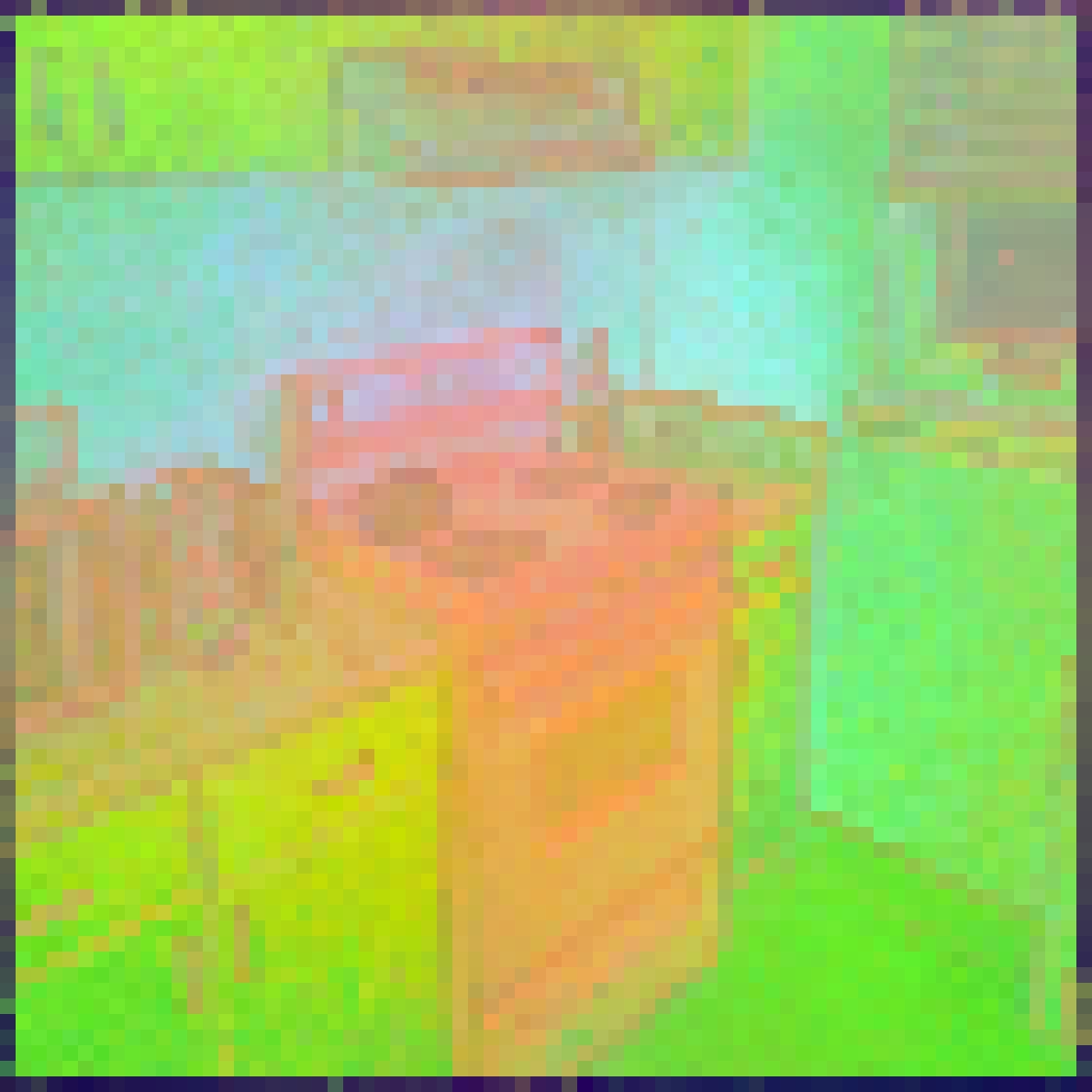}};
    \node[box, right=0cm of img8] (img9) {\includegraphics[width=3cm]{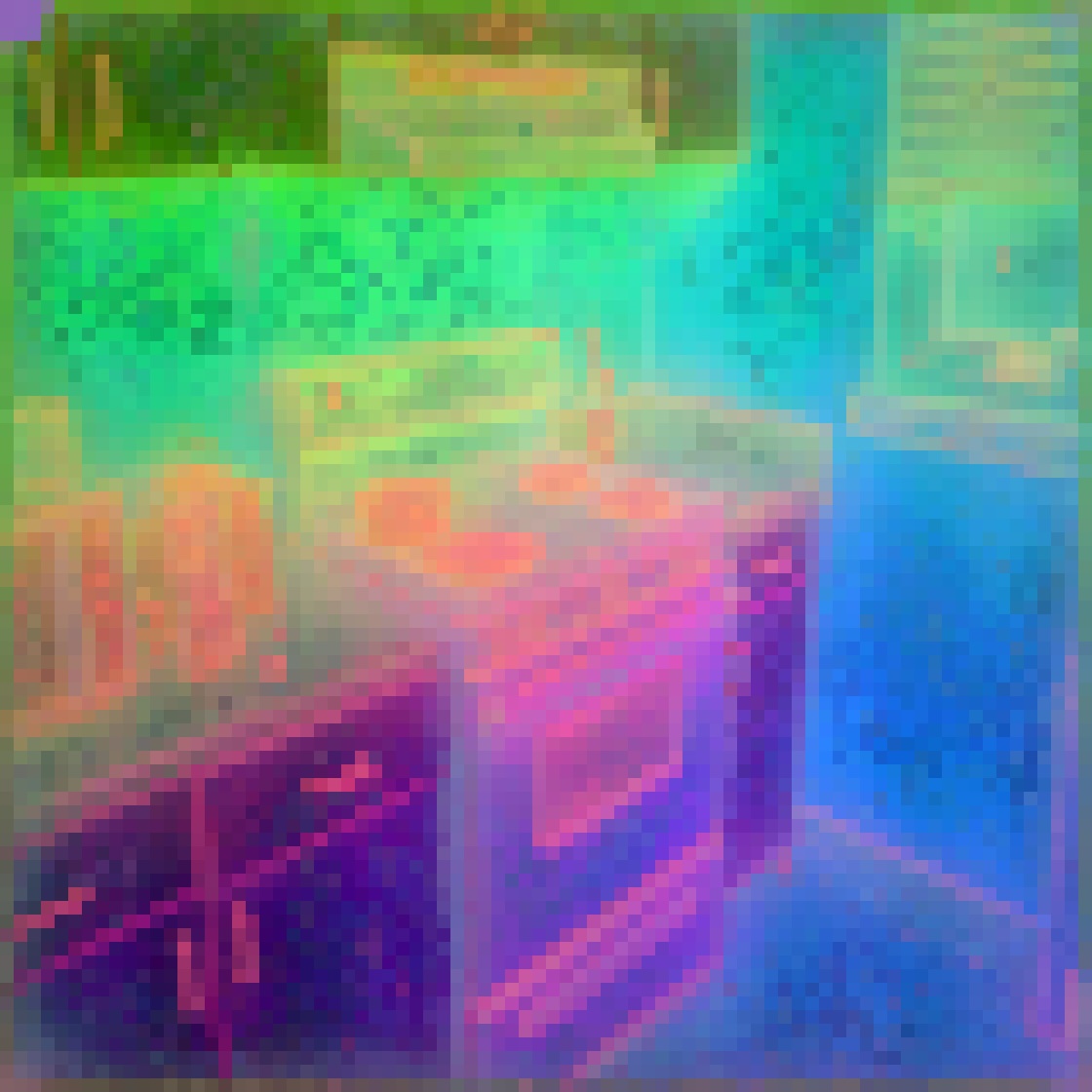}};

    \node[above=0mm of img2.north] {\footnotesize DINOcular};
    \node[above=0mm of img3.north] {\footnotesize DINO};
    \node[above=0mm of img4.north] {\footnotesize DUNE};

    \end{tikzpicture}}
    \caption{Visualization of the 3 strongest PCA components for DINOcular (ours), DINO ViT-B~\protect\citep{caron2021emerging}, and DUNE~\protect\citep{sariyildiz2025dune}. The mix of semantic and geometric information in our features is clearly visible at the edges of the kitchen objects.}
    \label{fig:qualitative}
\end{figure}
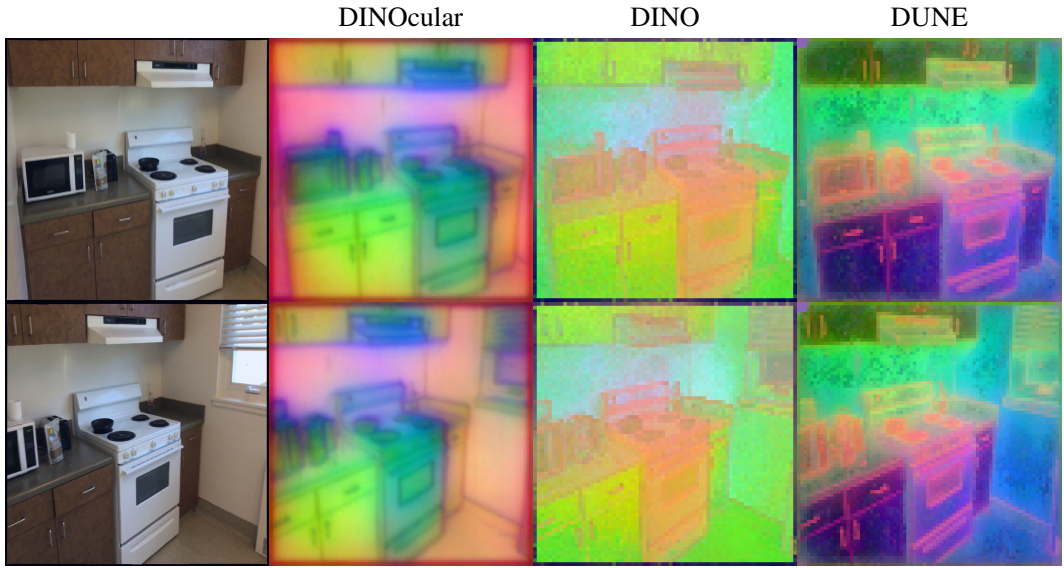

Existing approaches to visuospatial learning fall broadly into two categories. The first post-trains RGB-only foundation models for geometric consistency or reconstruction. Breakthrough systems such as VGGT~\citep{wang2025vggt} infer spatial structure from RGB observations, and recent work adapts visual representations to become more multi-view or spatially consistent. However, because these methods remain rooted in RGB-only inputs, they cannot fully exploit the complementary information provided by direct geometric sensing. The second category consists of task-specific RGB-D fusion architectures. Recent models have moved away from redundant dual-encoder designs toward more efficient mechanisms that use depth as a geometric prior. DFormerv2~\citep{yin2025dformerv2}, for example, introduces Geometry Self-Attention to guide feature interactions using depth while avoiding the cost of a fully separate depth stream. Yet, our experiments reveal that these architectures do not automatically learn generally useful visuospatial representations. 

In this paper, we propose a scalable, self-supervised framework for learning joint visuospatial representations from RGB-D observations. Our architecture efficiently integrates depth-derived geometric priors with a visual backbone, refining their interaction at inter-patch level through depth-aware 3D positional encoding and intra-patch through feature fusion. Coupled with a scalable, self-supervised training recipe over dense RGB-D observations, the resulting representation improves 3D awareness while preserving strong semantic transfer. Empirically, our model outperforms prior representations on multiple 3D benchmarks, including Probe3D~\citep{el2024probing}, and achieves competitive performance on standard RGB-D semantic segmentation benchmarks such as NYU DepthV2~\citep{Silberman:ECCV12} and SUNRGBD~\citep{song2015sun}. Our main contributions are:
\begin{itemize}
    \item We introduce an improved architecture that integrates visual and spatial information at both inter-patch and intra-patch levels.
    \item We present a self-supervised training recipe for learning generalizing visuospatial representations from RGB-D observations.
    \item We show that the learned representation improves over prior methods of comparable scale on 3D geometry benchmarks while remaining competitive on semantic segmentation tasks.
\end{itemize}

\section{Related Work}

\subsection{Spatial Understanding in Vision Foundation Models}
\label{sec:related_spatial_vfms}

Vision Transformers~\citep{dosovitskiy2020vit} established patch
tokens as the common interface for self-supervised pretraining and downstream
feature transfer. Two scalable families dominate the landscape. Self-distillation,
where DINO~\citep{caron2021emerging} and iBOT~\citep{zhou2021ibot} were
scaled to very large unlabelled corpora in DINOv2~\citep{oquab2023dinov2} and
DINOv3~\citep{simeoni2025dinov3} , yields some of the strongest available image
features. Masked reconstruction~\citep{he2022masked,wang2023videomae} forms
the second family and targets pixel-level recovery. We adopt the
self-distillation objective and extend it from RGB to RGB-D.

These features are strongly semantic but only weakly geometric.
Probe3D~\citep{el2024probing} and Lexicon3D~\citep{man2024lexicon3d}
report limited geometric consistency of frozen VFM features across viewpoints,
and \citet{you2025multiview} show that view consistency emerges only after
explicit multi-view finetuning.

A complementary line learns geometry directly from images.
CroCo~\citep{weinzaepfel2022croco} introduced this thread with self-supervised
cross-view completion, and its encoders underpin
DUSt3R~\citep{wang2024dust3r}, MASt3R~\citep{leroy2024grounding}, and
VGGT~\citep{wang2025vggt}, which cast multi-view reconstruction as
feed-forward pointmap and unified geometry prediction. Follow-up work
generalises the paradigm to dynamic scenes, persistent state, and broad metric
reconstruction~\citep{zhang2025monst3r,wang2025continuous,keetha2026mapanything}. Across this family RGB is the input and geometry is the output, so the feature space is shaped to predict 3D structure rather than to represent it.

Both families therefore leave depth outside the feature extractor, either absent
in RGB-only SSL or present only as a prediction target. What is absent are generalizing representations that combine RGB and Depth inputs.

\subsection{RGB-D Fusion and Geometric Priors}
Early RGB-D models used dual-encoder architectures with convolutional or
attention-based
fusion~\citep{hu2019acnet,chen2020bi},
transformer-based models shifted fusion to the token
level~\citep{wang2022multimodal}, and a more recent body of
work refines supervised RGB-D segmentation through primary-modality
distillation, heterogeneous branches, and robustness to degraded
depth~\citep{gong2026diffpixelformer,jaganathan2026geomprompt}.
These methods train the encoder jointly with a segmentation head, so the
representation inherits the inductive bias of the target task and does not
transfer cleanly to other objectives~\citep{yin2024dformer}.

DFormer~\citep{yin2024dformer} also targets RGB-D segmentation, but pretrains a backbone on image-depth pairs
from ImageNet under supervised classification, while
DFormerv2~\citep{yin2025dformerv2} derives a geometry prior from pooled patch
distances and relative depth, applied as an additive bias on self-attention
weights. This prior operates at the pooled-patch level and modifies attention
weights rather than token values or embeddings, so depth informs how patches
attend but does not enter the feature representation itself. We instead embed
depth at the level of token positions. Modern vision transformers encode token
position relatively through RoPE~\citep{su2024roformer}, and
\citet{schenck2025learning} extend the mechanism to a third axis for robotic
action models. We follow that direction and treat image-plane coordinates
together with mean patch depth as a joint position, so geometry shapes feature
computation rather than reweighting a 2D affinity.

Several self-supervised methods incorporate geometry without producing a native
RGB-D image backbone. MultiMAE~\citep{bachmann2022multimae} treats depth as a
reconstruction target rather than as input geometry.
Concerto~\citep{zhang2026concerto} jointly trains a 3D point cloud transformer
and a 2D image encoder, but still requires a point cloud at inference and uses
images only as a supervisory signal. Sonata~\citep{wu2025sonata} trains a
self-supervised point cloud encoder on aggregated scene-level scans and
suppresses spatial information to avoid a geometric shortcut, where
coordinate-based positional encodings cause representation collapse. In contrast, we target the more broadly available single-view RGB-D setting, avoiding assumptions about aggregated scans, fixed cameras, or scene-level point clouds.

\section{Method}
We investigate visuospatial grounding in two dimensions: model architecture and learning objectives. In the following, we first discuss how we integrate depth as an input modality to modern vision transformer architectures. Second, we describe the self-supervised learning objectives that we train these models on. Third, we describe how we modify a Swin architecture to implement these methods.

\subsection{Geometry Prior} \label{geoprior}
Inspired by the effective approach of incorporating depth in Dformerv2~\citep{yin2025dformerv2}, we explore approaches that make use of depth as a geometry prior, without creating large additional branches or complex fusion mechanisms. In particular, although~\citet{yin2025dformerv2} achieves good performance, it is highly dependent on RMT~\citep{fan2024rmt}, and simply biasing self-attention based on the weighted depth distances. It uses depth to attend to parts of the input image that are metrically close by, strictly baking into the architecture the heuristic that close-by parts matter. 

Instead of additive biasing, we extend rotary positional encoding (RoPE)~\citep{su2024roformer} into three dimensions, following ~\citet{schenck2025learning}. They already show that incorporating depth in this way improves performance in robotic action models.
Given that RoPE is a foundational component in the self-attention mechanisms of most modern Vision Transformers (ViTs), extending it to a third dimension allows the model to inherently leverage relative spatial relationships based on the depth input.
By encoding depth directly into the coordinate system (as 3rd, z-axis), we create a more principled prior that enables the model to reason about the physical proximity of features, but e.g. also to attend to patches further away. Following~\citet{schenck2025learning}, for patch coordinates $u, v \in \mathbb{N}$ on the image plane and average depth $\bar{z} \in \mathbb{R}^{\geq 0}$ of all pixels within a patch, we find the encoding $\hat{x}$ of the patch embedding $x$ as:
\begin{align*}
    \hat{x} = \textrm{exp}\left( \textbf{L}_1 u + \textbf{L}_2 v + \textbf{L}_3 \bar{z} \right) \, x
\end{align*}
where each $\textbf{L}_k \in \mathbb{R}^{d\times d}$ is a rotational positional embedding $\textbf{L}_k = \sum_{p=1}^{d/2} \left( \delta_{2p, 2p-1} - \delta_{2p-1, 2p} \theta_p\right)$ with learnable frequencies $\theta_p$. Thus, we completely change approach seen in DFormerv2 with our inter-patch mechanism - 3D RoPE.

3D positional encoding of patches gives the model access to the global geometry of an observation. However, it does not make use of the pixel-level resolution of depth input that also holds important information of the local shape, e.g. whether edges are round or sharp or whether surfaces are uneven or smooth. We therefore add this local geometric information to the patchified RGB embeddings by extracting local features from the depth within each patch. We apply lightweight patch embedding layer and fuse both embeddings through a linear projection, giving the model intra-patch geometry information.
One might suspect that scaled, metric distance is not relevant for local geometric details, and geometric intra-patch embedding would generalize better from a scale-free, normalized input. In our experiments, we therefore investigate an ablation where local shape features are instead computed based on per-pixel surface normals. Surface normals are the spatial derivative of the depth value and therefore scale-free. However, empirically we observe this variant to collapse in training.

\subsection{Learning objective}
\label{sec:objectives}
We design the learning objective in a way that it should explicitly learn fine-grained spatial information along with the visual-semantic content.
As semantic loss objective $\mathcal{L}_{sem}$, we utilize the teacher-student distillation framework of DINO and DINOv2~\citep{caron2021emerging,oquab2023dinov2}. We briefly summarize how we apply these and refer the interested reader to~\citet{oquab2023dinov2} for more details:
\begin{itemize}[leftmargin=1em,itemsep=0.1em,topsep=0em]
    \item \textbf{DINO / Image-level objective} We give student and teacher different crops of the same object-centric image and take the \texttt{cls} tokens from each. They are passed through a DINO head and finally a softmax activation.  $\mathcal{L}_\textrm{DINO}$ is then the cross-entropy between the student's output and the centered teacher's output.
    \item \textbf{iBOT / Patch-level objective} For the student, some of the input tokens are masked. Importantly, that means in our case that both RGB and intra-patch Depth are masked, but even the masked patch tokens are positionally encoded based on average patch depth. Patch tokens are then projected through an iBOT head and a softmax activation. $\mathcal{L}_\textrm{iBOT}$ is then cross-entropy between the student's output and the centered teacher's output over all masked patches.
\end{itemize}
In our experiments, DINO and iBOT heads share the same parameters, and we do not apply the untying of~\citet{oquab2023dinov2}. That is because our experiments cannot reproduce the data scale at which they observed untying to be useful. However, we follow~\citet{oquab2023dinov2} to apply the KoLeo regularizer~\citep{sablayrolles2018spreading}, arriving at the full semantic-visual objective:
\begin{align*}
    \mathcal{L}_{sem} = \mathcal{L}_\textrm{DINO} + \mathcal{L}_\textrm{iBOT} + \lambda \mathcal{L}_\textrm{KoLeo}
\end{align*}

\begin{figure}[t]
  \centering
  \includegraphics[width=.8\textwidth]{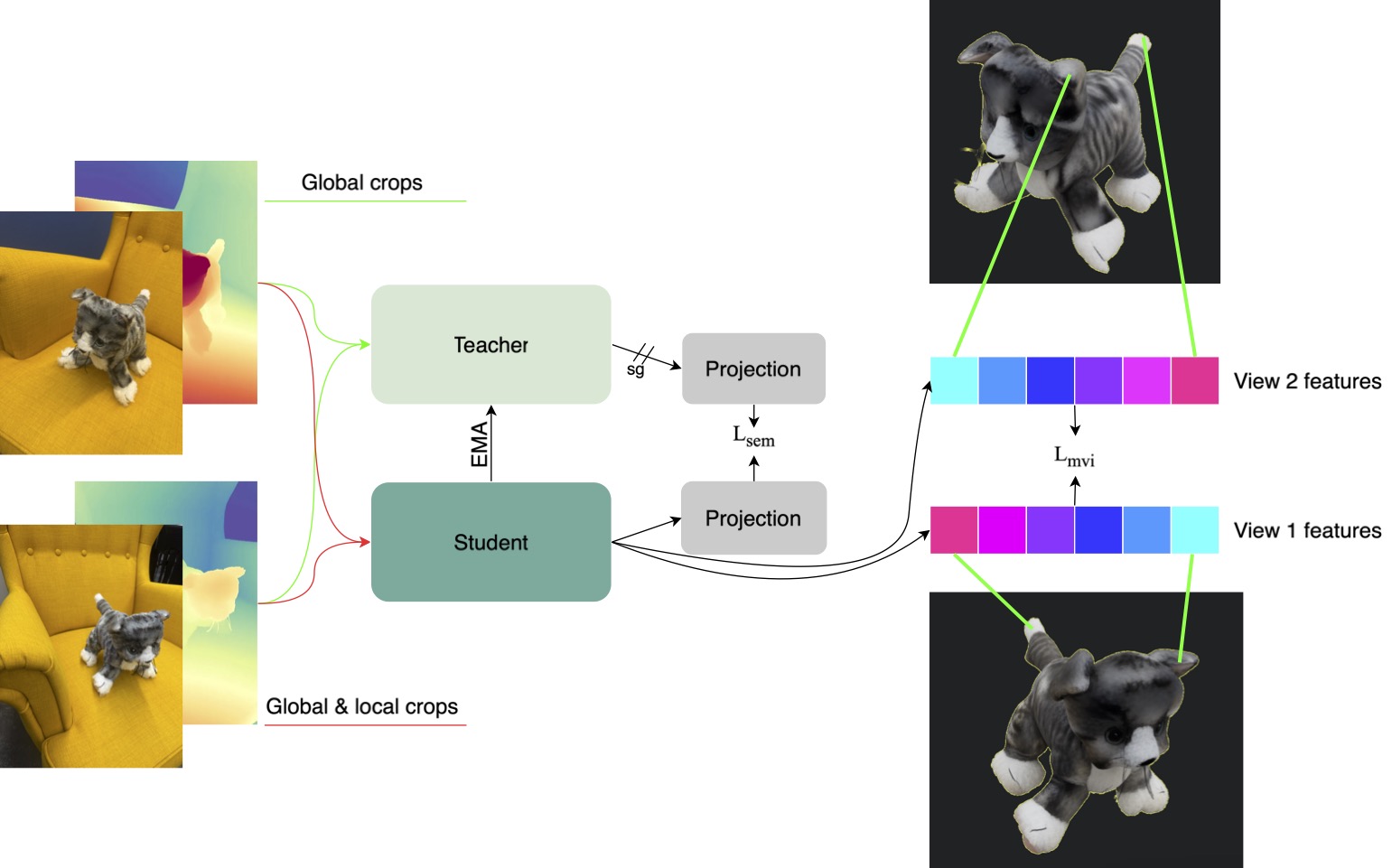}
  \caption{Multiview Pre-training setup. Teacher recieves 2 global unique views, student additionally get 4 local crops from each view. 
  $L_{sem}$ is a self-distillation loss, $L_{mvi}$ is a multiview correspondence loss calculated at randomly sampled 3D points on the object that are close in space.
  Teacher is updated through exponential moving average.}
  \label{fig:stud}
\end{figure}

To fully leverage the spatial information of depth, we combine the above loss with a spatial objective. Our reasoning is that while visual appearance may change between viewpoints, the same part of an object is more easily matchable in 3D. Given the additional spatial input to the model, we require an objective that enforces the use of this information. We therefore experiment with different formulations of a multi-view consistency loss. In all cases, two students receive two different images of the same object, but from different viewpoints. From known point maps and the depth, we find a subset of patches that is visible to both. We then enforce similarity between each pair of matched patches $(z_1, z_2)$. This is visualized in Figure~\ref{fig:stud}.

\begin{itemize}[leftmargin=1em,itemsep=0.1em,topsep=0em]
    \item \textbf{Cosine Similarity} Following~\citet{koch2025unified}, who propose a post-training refinement of DINOv2 features with this objective, we optimize patch embeddings to have low cosine distance: $\mathcal{L}_\textrm{mv-cosine} = 1 - z_1 \cdot z_2$.
    \item \textbf{Contrastive Ranking} Following~\citet{you2025multiview}, who investigate finetuning of DINOv2 and other vision foundation models for multi-view consistency, we apply a SmoothAP~\citep{brown2020smooth} ranking-based loss over the matched patches. 
    \item \textbf{Multi-view iBOT} Since the iBOT objective works on a patch level, a variant of this objective for matched patches from different views is easily conceivable. In this variant, the second image however is passed through the teacher. We bias the mask sampling to those patches with visual overlap and retrieve pairs $z_1$ and $z_2$, where $z_1$ has a masked input. Both are passed through a projection head and softmax activation, and centering (for the teacher). $\mathcal{L}_\textrm{mv-ibot}$ is then the cross-entropy between these 2 outputs.
\end{itemize}

Our ablations find empirically that $\mathcal{L}_\textrm{mv-contrastive}$ is the best compatible with $\mathcal{L}_{sem}$, leading to our full objective $\mathcal{L} = \mathcal{L}_\textrm{DINO} + \mathcal{L}_\textrm{iBOT} + \lambda_1 \mathcal{L}_\textrm{mv-contrastive} +  \lambda_2 \mathcal{L}_\textrm{KoLeo}$. We sample batches such that in every batch there are pairs of images to which the multi-view loss can be applied. 

During model development, we observed that the combination of multi-view training and added intra-patch depth embedding yielded features that were heavily biased towards spatial tasks and contained less visual information. This hints at the connection between depth input and the multi-view consistency that we hypothesize above, but yields less general features. We address this by applying dropout to the intra-patch depth embedding layers, forcing the multi-view objective to not exclusively rely on depth information.

\subsection{Technical Implementation}
The backbone architecture is a Swin hierarchical vision transformer \citep{liu2021swin} with 4 stages presented in~\citet{fan2024rmt} and further used in~\citet{yin2025dformerv2}, who added RGB-D blocks. We further completely replace mechanisms used in DFormerv2~\citep{yin2025dformerv2} with our 3D RoPE discussed in~\ref{geoprior} for inter-patch visuospatial information. And additionally provide
first RGB-D block with intra-patch geometry as lightweight depth encoding. The architecture is presented in Figure~\ref{fig:arch}, and additional training details are reported in appendix~\ref{app:training}.

\begin{figure}[t]
  \centering
  \includegraphics[width=1.0\textwidth]{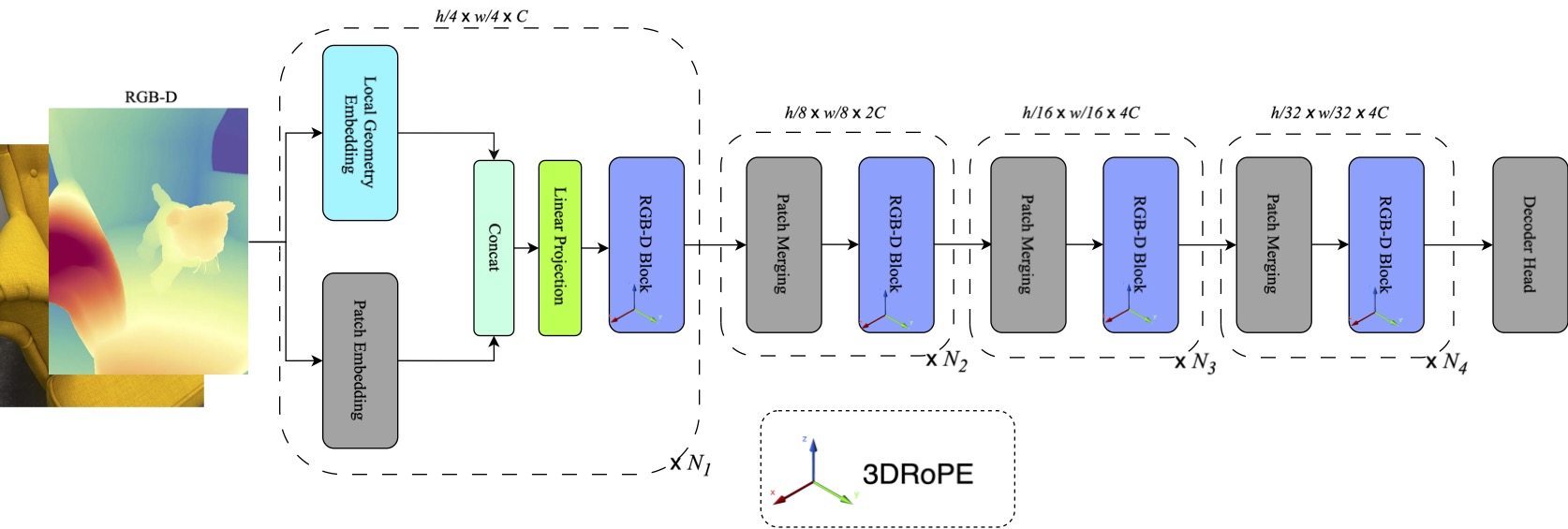}
  \caption{Architecture with 3D RoPE and local geometry encoding.}
  \label{fig:arch}
\end{figure}

\section{Experiments}
We empirically test our proposed method over a range of visuospatial tasks: RGB-D Semantic Segmentation, 3D Correspondence Estimation, and Object Pose Estimation.
We first discuss dataset preparations and chosen methods of comparison. Then we justify our method choices in ablation studies. Finally we compare our trained model with prior work on the different visuospatial tasks.
We additionally report effectiveness measurement in appendix~\ref{app:effective} and qualitative results in appendix~\ref{app:qual}.

\subsection{Training Data and Details}
We generate Depth for all our training samples using monocular depth estimation on ImageNet-1k and multi-view triangulation on MVImgNet2.0. For both, we use MapAnything~\citep{keetha2026mapanything}. Prior work~\citep{yin2025dformerv2,yin2024dformer} generated ImageNet-1K depth maps using AdaBins~\citep{bhat2021adabins}. We found that regenerating them with a newer model improved consistency with the rest of our data and performance. Our multi-view objective additionally requires point maps. We save point maps from MapAnything during depth generation. We also predict object masks using SAM3~\citep{carion2026sam}, to constrain objective's focus on the object. For ImageNet-1K, we use all \SI{1.3}{M} images. For MVImgNet2.0, we do not need the large viewpoint redundancy, so we retain 307k samples with approximately 8 views each on average.

\subsection{Baselines}
We compare against two types of methods: those learning features that combine spatial and visual-semantic information, and self-supervised visual representation learning.
\begin{itemize}[leftmargin=1em,itemsep=0.1em,topsep=0em]
    \item \textbf{Dformerv2}~\citep{yin2025dformerv2} proposes an architecture that combines RGB and Depth input to learn a general representation. They train fully supervised on image classification. This method is most directly comparable as we adapt their architecture and train the same model size on a similar scale of data.
    \item \textbf{Sonata}~\citep{wu2025sonata} is a self-supervised method to learn features for point clouds. It is the only prior work of a self-supervised method for visuospatial features. However, it is meant to be applied on merged scene-level pointclouds with more even density. Comparing it on the single-view data that our model digests is ill posed and we merely report its performance as reference.
    \item \textbf{DINO family}~\citep{caron2021emerging,oquab2023dinov2,simeoni2025dinov3} is a family of models trained in self-supervised fashion on RGB only. We directly adopt their loss. DINO trains on ImageNet-1k, same as Dformer and our method and therefore is well comparable. However, for DINOv2 and DINOv3, from where we adopt the full loss, their datasets are $\times 36$ and $\times 433$ larger and not made public. Therefore, we try to reproduce results of DINOv2 in our settings or test public checkpoints with comparable data scale~\citep{wu2025simplifying}.
    \item \textbf{DUNE}~\citep{sariyildiz2025dune} also only takes RGB as input, but distills a feature from both DINOv2 and Mast3r, therefore aiming to add more geometric information to its features~\citep{zhang20263d}. It is distilled from DINOv2 on an additional set of \SI{21}{M} images, so we report its results only for reference.
\end{itemize}

\input{tables/mlp_enc}
\input{tables/loss_abl}

\subsection{Ablation: Model Architecture}
We present an ablation of our architectural choices to encode inter- and intra-patch depth in Table~\ref{tab:ablation_enc}.
Surprisingly, even for the Dformerv2 architecture itself we find that exchanging the supervised with the DINO objective increases usefulness of the learned features with respect to segmentation probing, even though~\citet{yin2025dformerv2} specifically target segmentation as their goal task.
Our modification of their architecture with 3D RoPE instead of proximity attention bias yields the largest per-step increase in feature probing, for all 3 measured indicators.
Contrary to the results from~\citet{tziafas2023early}, the encoding of intra-patch geometry through surface normals does not show good results in our self-supervised setting. However, we find the simple MLP encoding of the depth values to be reasonably effective, most notable from the depth reconstruction. It does not surpass the setting without any intra-patch depth, but we address this aspect in the design of the full training objective.

\subsection{Ablation: Combination of Loss Objectives}
In Table~\ref{tab:ablation_multiview}, we investigate the different options for multi-view consistency losses, as introduced in Section~\ref{sec:objectives}. For all objectives, we observe a collapse in segmentation performance when finetuning purely on the multi-view objective. This is expected to a certain degree, as the objective optimizes a different goal. The effect can also, as the results show, be compensated by training on a combined objective. However, what is unexpected is the drop in 3D correspondence estimation for MV-Cosine and MV-iBOT. On closer inspection, we find that~\citet{koch2025unified} use this loss in ensemble with 3 distillation losses that likely regularize the features well and therefore can prevent such collapse. In our setting where we do not distill from a frozen teacher, the objective is not effective. The contrastive objective is also used in~\citet{you2025multiview} as an isolated objective and also in our case is the only one that achieves an improvement in 3D correspondence estimation. It also shows to be compatible with the DINO objective, as a combination of both not only compensates the loss in segmentation performance, but further enhances the 3D correspondence.

\input{tables/main}

\subsection{Semantic Segmentation}
An established downstream task for RGB-D perception is semantic segmentation. This task also allows us to evaluate over different data sources that are captured with different depth sensors: NYU Depth V2~\citep{Silberman:ECCV12} and SUN RGB-D~\citep{song2015sun} are based on the first generation of Kinect with structured light, Cityscapes~\citep{cordts2016cityscapes} has a stereo camera, and ADE20k has no actual depth available. We process ADE20k, NYU Depth v2, and SUN RGB-D with MapAnything~\citep{keetha2026mapanything} to output dense depth. For Cityscapes, we acquire dense depth from FoundationStereo~\citep{wen2025foundationstereo}. For all methods and datasets, we fit a linear probe over their training data, keeping the feature predictor completely frozen. The results are reported in Table~\ref{table:segm}.

Our features show significantly better probing results compared to both Dformerv2 and DINOv2. In most cases, DINOcular\_S even outperforms the 3 times larger DINO ViT-B, while our larger model keep advantage over models with comparable data and parameter size. However, the reference results of other large representation learners show a clear advantage of data scaling. There is also a visible tradeoff between the multi-view objective and the best possible semantic feature. While our general model still outperforms both DINOv2 and Dformverv2 at the same data scale, the features learned when leaving out the multi-view objective have even (slightly) better probing results for these semantic datasets. 
To rule out whether performance gains are coming from depth source, we report result with diverse depth inputs in appendix~\ref{app:sensordepth} to show robustness of our model.

\subsection{3D Correspondence Estimation}\label{sec:3dcorr}
To evaluate the consistency of features over different observations of the same object, we test the matching of features between pairs of images against their true correspondence based on the known object geometry. Following the evaluation setup of~\citet{el2024probing}, we test this on object-centric observations from NAVI~\citep{jampani2023navi} and scene-level observations on ScanNet~\citep{dai2017scannet}. We report results grouped by the angle change between viewpoints in Table~\ref{table:3dcorr}. Especially for large viewpoint changes, we observe a clear benefit of the  multi-view training over both datasets. Remarkably, on Probe3D-ScanNet, DINOcular-S features outperform even all reference models, including DINOv3 ViT-B. Since this stands out from the more object centric NAVI data, a possible reason is that for the cluttered and larger scale environments of ScanNet, the depth provides a more noticeable advantage.

\subsection{3D Object Pose Estimation}
3D Object Pose Estimation requires a model to predict, in the coordinate frame of the camera, where an object is and how it is oriented. We evaluate single-shot, CAD-model-free object pose estimation for low-texture objects with the different frozen backbones on the dataset~\citep{he2022onepose} and evaluation protocol of~\citet{you2025multiview}. We report recall for different thresholds of estimation error. We find that, in line with the results of Section~\ref{sec:3dcorr}, DINOcular-L features outperform all baselines and even the DINOv3 ViT-B model on the two coarser thresholds. However, there is an apparent weakness of the smaller model on the \SI{1}{cm}-\SI{1}{^\circ} threshold, potentially related to the blurriness of the features over the image plane that is also apparent in Figure~\ref{fig:qualitative}.

\section{Limitations}
Our study focuses on moderate data scale, and we have not yet validated whether the proposed method preserves the same gains when scaled to substantially larger and diverse data. When compared at this data scale, DINOcular outperforms RGB-D and RGB-only foundation models including the DINO model family. However, it is hard to fully compare models at the same data scale. This is because DINO does not make their training data available, but also because DINOcular additionally requires multi-view data, which in turn our ablation showed reduces performance of DINO models. Finally, although our learned features transfer across multiple geometric and semantic tasks, we observe a trade-off between spatial and semantic specialization: removing or weakening the multi-view objective can improve semantic performance, but at the cost of the 3D-awareness of the representations.

\section{Conclusion}
We present DINOcular, a framework for learning visuospatial features from RGB-D observations in a self-supervised manner. We demonstrate that these features generalize across semantic and geometric tasks. Our results show a significant advantage over other approaches at the same data and model scale. This is promising and motivates future research into scaling the proposed approach to larger data sources and models.
Looking further ahead, this work shows that it is possible to learn generalizing features that combine complementary information from RGB and Depth inputs. This may help increase the spatial understanding of VLMs and VLAs, eventually contributing to deploy more generalizing robots to real-world environments.

\bibliography{main}
\bibliographystyle{bibliography}

\appendix

\section{Different Depth Sources as Input}\label{app:sensordepth}
 We test inference on 3 different sources of depth: 2 monocular models - MapAnything and DepthAnything3. We do not observe strong performance variations between these depth sources. Due to the nature of the sensing setups, a controlled comparison between stereo and time-of-flight sensors is not possible. However, we present an additional comparison of raw sensor depth (some pixels without measurement) vs densified sensor depth vs monocular depth on NYU in Table~\ref{tab:robust}.
\begin{table}[h]
    \centering
    \caption{We probe our models on RGB-D Semantic Segmentation for ADE20k and NYU Depth v2, where depth needs to be predicted by monocular models or provided by sensors. We ablate MapAnything~\protect\citep{keetha2026mapanything} with DepthAnything3~\protect\citep{lin2025depth} and nearest-neighbor densified sensor depth. The probing of the resulting features appears to be robust to variations in the depth input.}
    \begin{tabular}{llr}
    \toprule
    Evaluation & Depth input & DINOcular [mIoU] \\
    \midrule
    ADE20k       & MapAnything                              & 33.36 \\
    ADE20k       & DepthAnything3                           & 32.68 \\
    NYU Depth v2 & MapAnything                              & 40.27 \\
    NYU Depth v2 & Sensor depth, nearest-neighbor densified & 39.01 \\
    NYU Depth v2 & Sensor depth, densified w/ MapAnything   & 41.05 \\
    \bottomrule
    \end{tabular}
    \label{tab:robust}
\end{table}

\section{Efficiency}\label{app:effective}
We measure performance of three models that share the same core backbone architecture in Table~\ref{tab:efficiency}. Two baselines are: RGB-only follows ~\citep{fan2024rmt} and DFormerv2~\citep{yin2025dformerv2} with added depth-based biasing to the Manhattan self-attention proposed in RGB-only backbone. DINOcular is faster and uses less memory than DFormerV2, with a modest latency and memory overhead relative to the RGB-only backbone.

\begin{table}[h]
    \centering
    \caption{Performance and efficiency comparison. Latency and memory are reported with relative differences compared to Dinocular.}
    \begin{tabular}{lrrr}
    \toprule
    Model & Parameters (M) & Average Latency (ms) & Peak Memory (MB) \\
    \midrule
    Dinocular & 26.7 & 51.07 & 249.12 \\
    DFormerV2 & 26.7 & 53.14 (+4.05\%)  & 289.35 (+16.15\%) \\
    RGB-only  & 26.7 & 47.14 (-7.70\%)  & 245.73 (-1.36\%) \\
    \bottomrule
    \end{tabular}
    \label{tab:efficiency}
\end{table}

\section{Qualitative Results}\label{app:qual}
We present a qualitative evaluation of DINOcular across multiple downstream tasks, illustrating its learned feature representations, 3D geometric awareness, and 2D semantic segmentation capabilities in diverse environments.

\begin{figure}[h]
    \centering
    \begin{subfigure}[b]{0.32\textwidth}
        \centering
        \textbf{DINOcular}\\[1ex]
        \includegraphics[width=\textwidth]{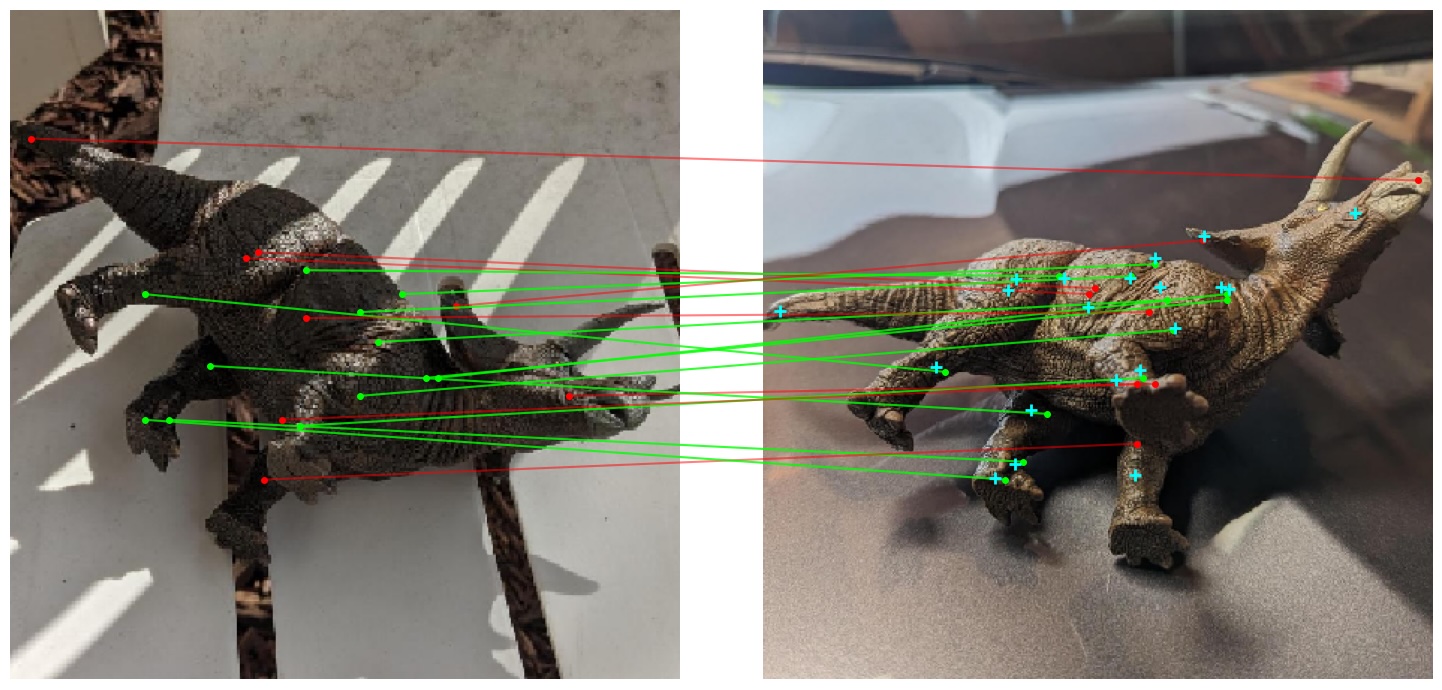}
    \end{subfigure}
    \hfill
    \begin{subfigure}[b]{0.32\textwidth}
        \centering
        \textbf{DINO}\\[1ex]
        \includegraphics[width=\textwidth]{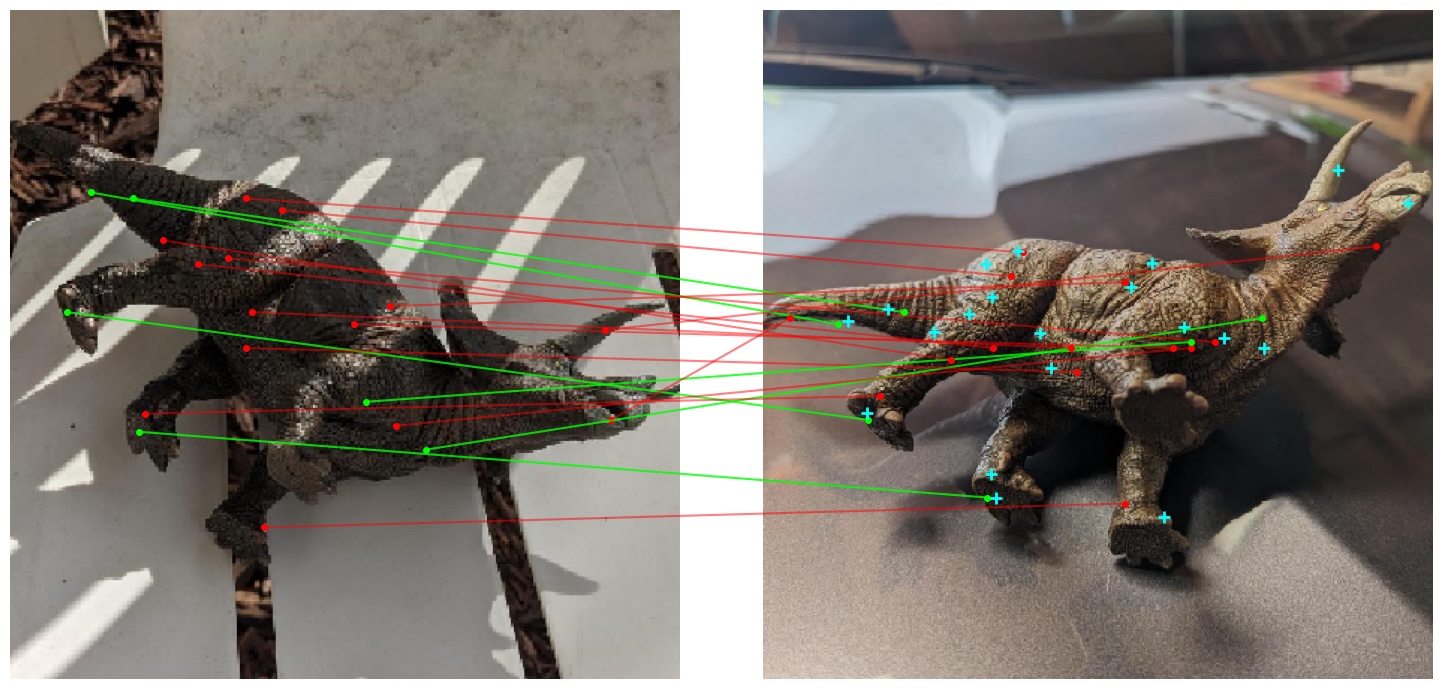}
    \end{subfigure}
    \hfill
    \begin{subfigure}[b]{0.32\textwidth}
        \centering
        \textbf{DUNE}\\[1ex]
        \includegraphics[width=\textwidth]{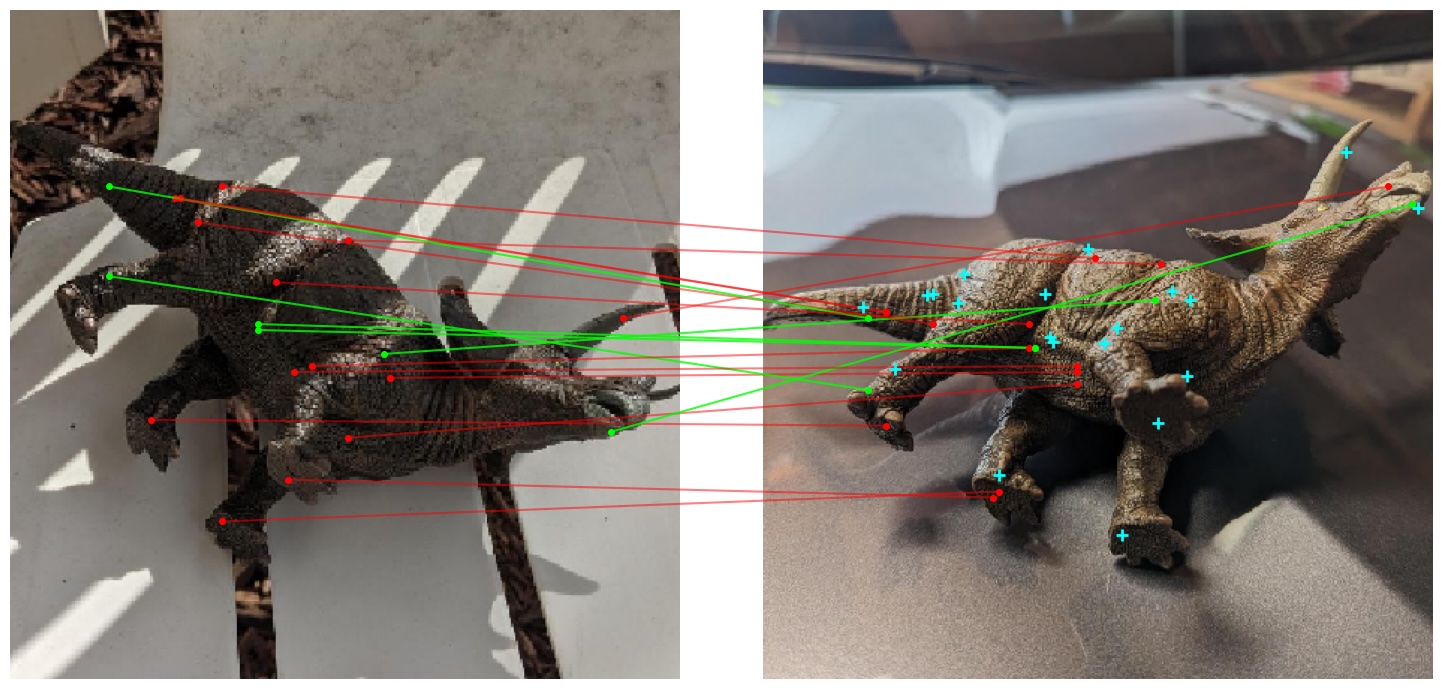}
    \end{subfigure}
    \vspace{0.2cm}

    \begin{subfigure}[b]{0.32\textwidth}
        \centering
        \includegraphics[width=\textwidth]{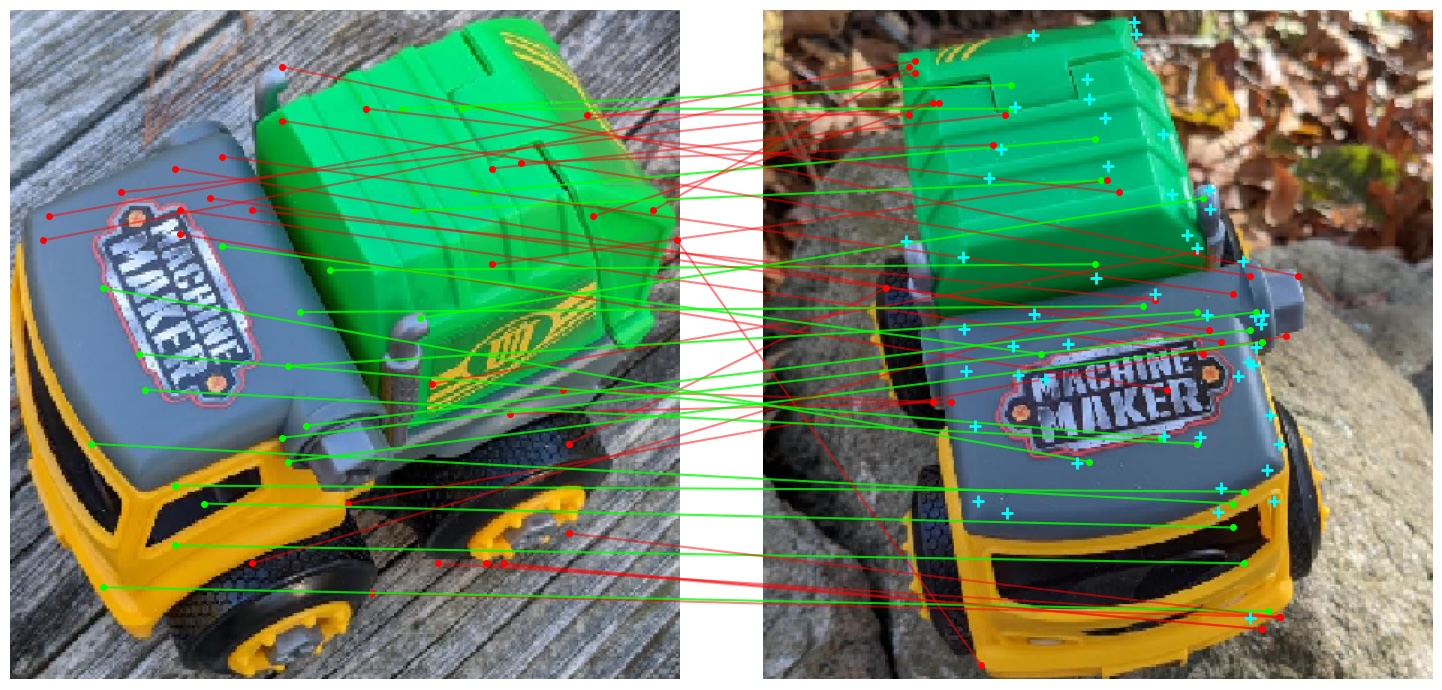}
    \end{subfigure}
    \hfill
    \begin{subfigure}[b]{0.32\textwidth}
        \centering
        \includegraphics[width=\textwidth]{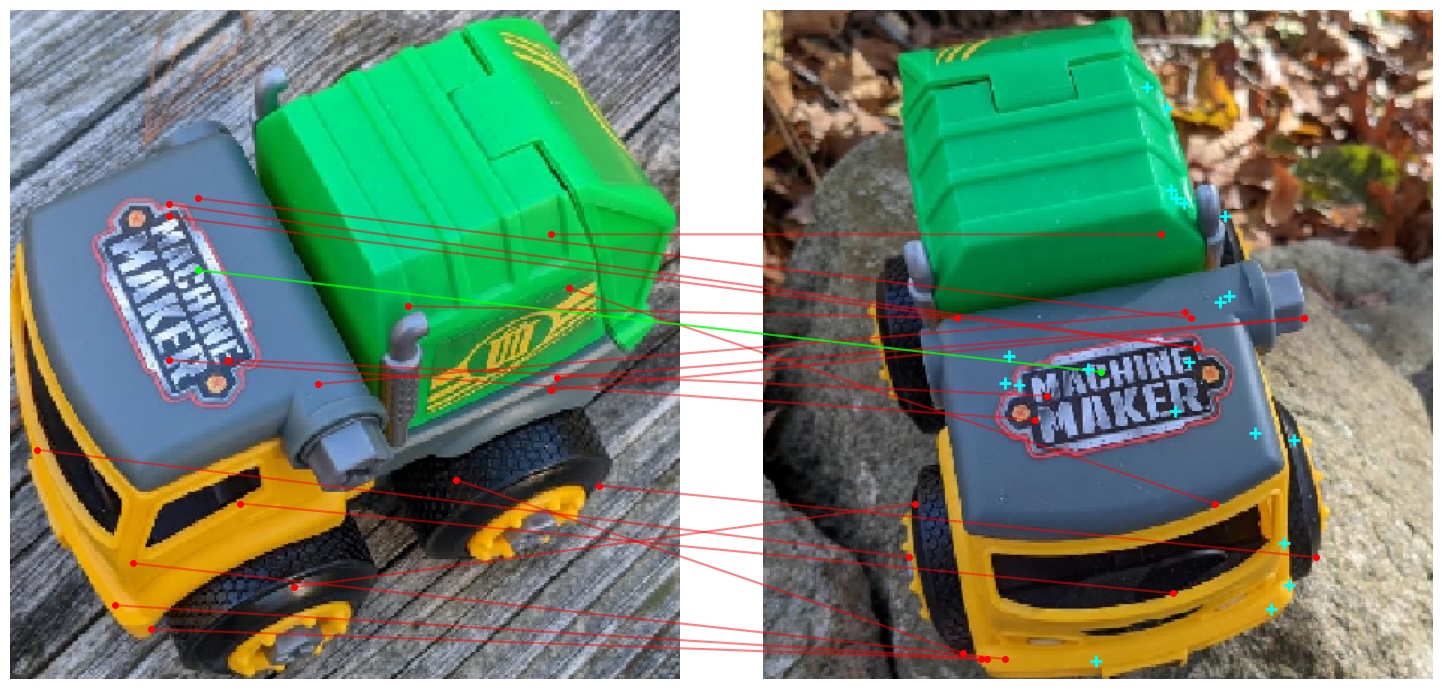}
    \end{subfigure}
    \hfill
    \begin{subfigure}[b]{0.32\textwidth}
        \centering
        \includegraphics[width=\textwidth]{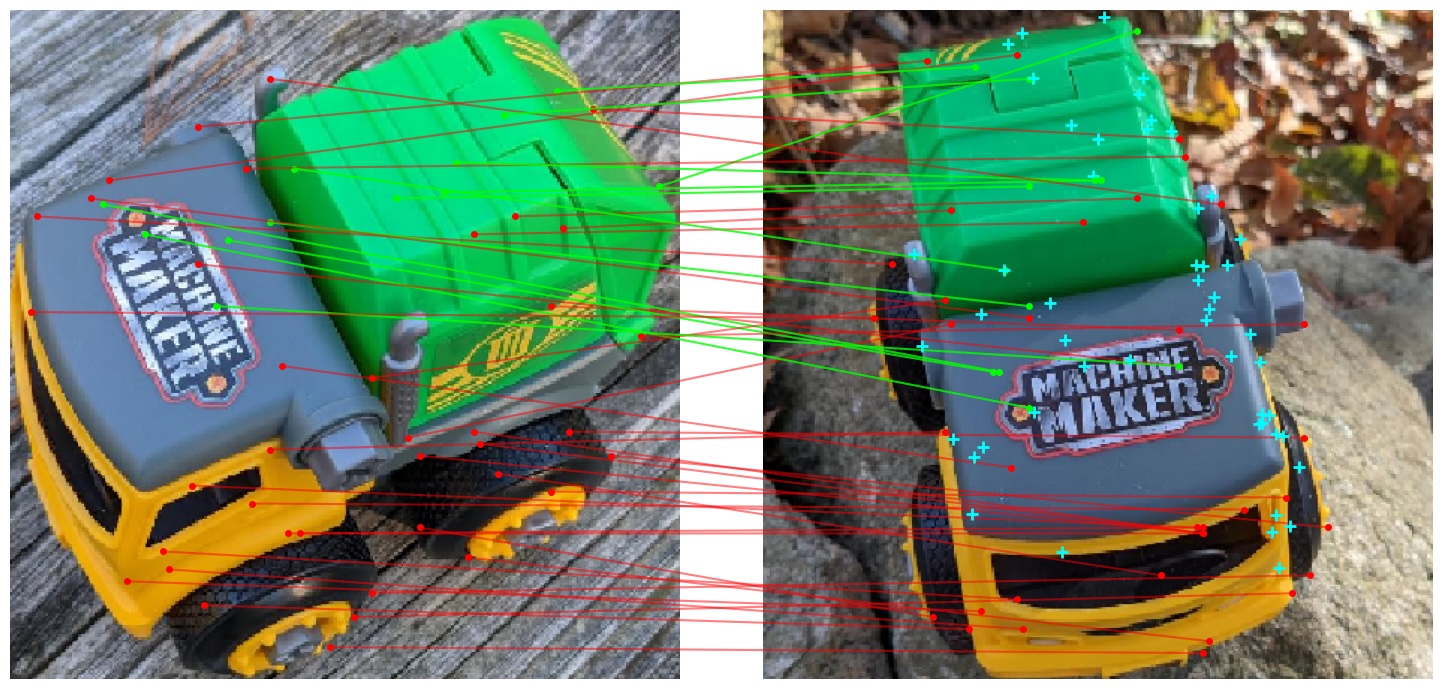}
    \end{subfigure}
    \vspace{0.2cm}

    \begin{subfigure}[b]{0.32\textwidth}
        \centering
        \includegraphics[width=\textwidth]{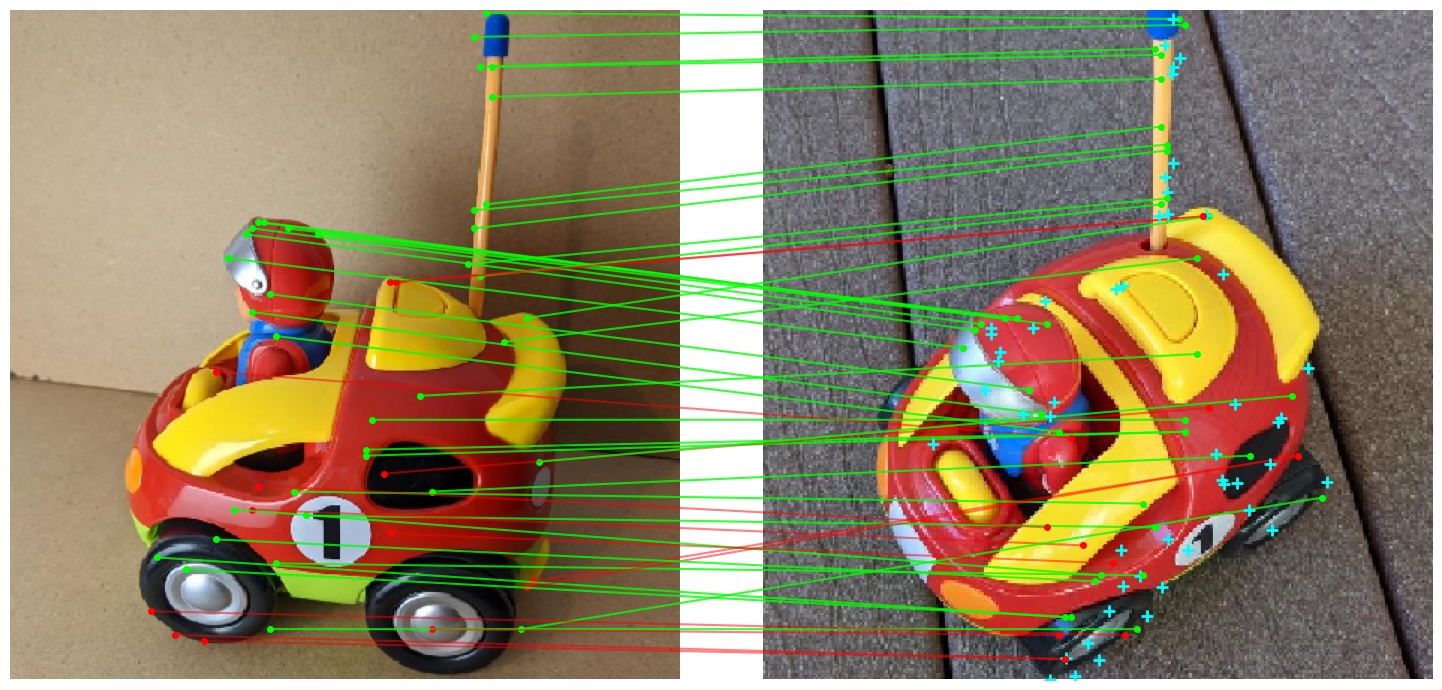}
    \end{subfigure}
    \hfill
    \begin{subfigure}[b]{0.32\textwidth}
        \centering
        \includegraphics[width=\textwidth]{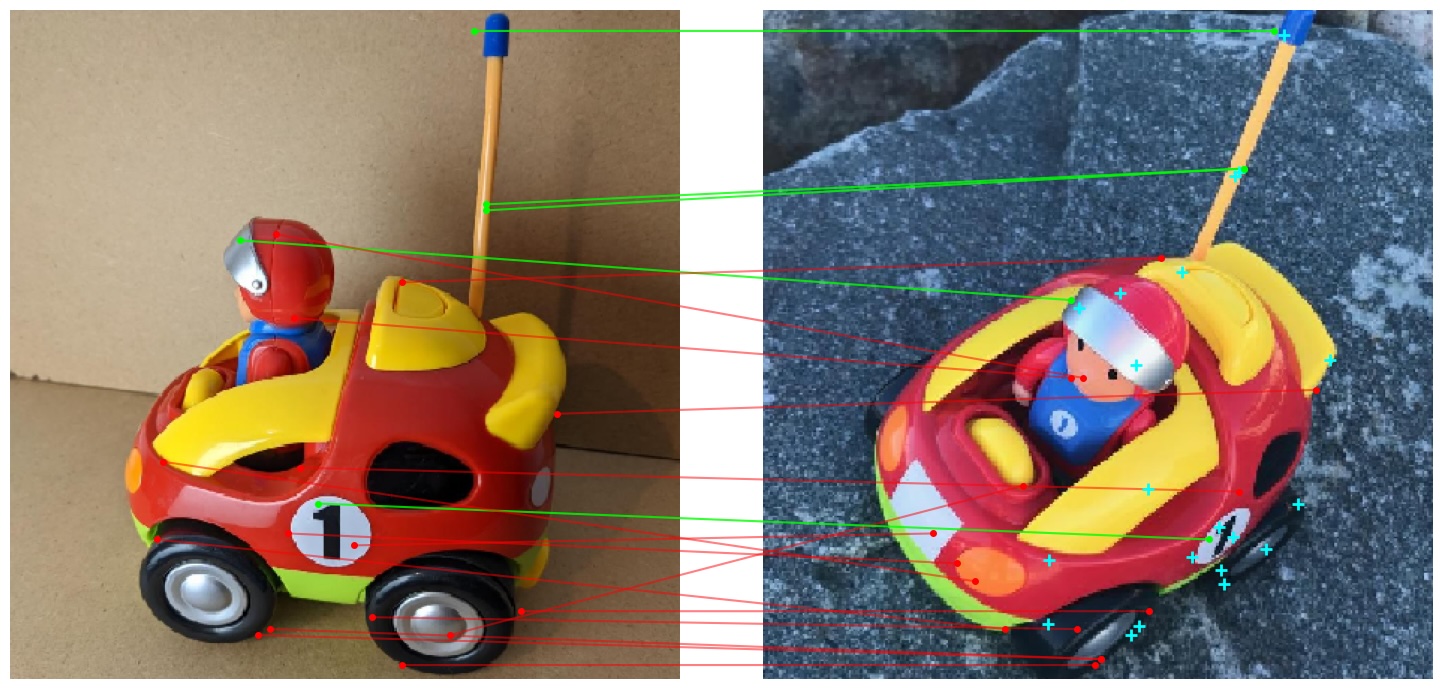}
    \end{subfigure}
    \hfill
    \begin{subfigure}[b]{0.32\textwidth}
        \centering
        \includegraphics[width=\textwidth]{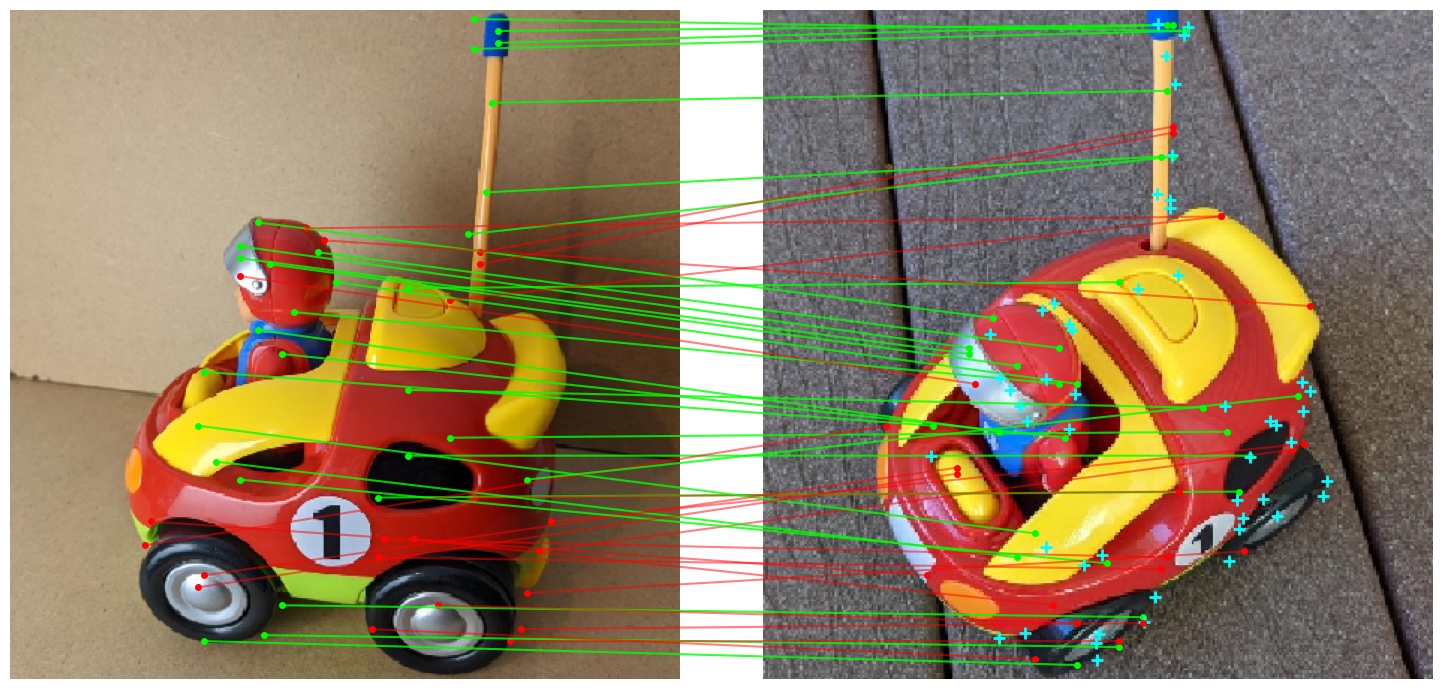}
    \end{subfigure}
    \vspace{0.2cm}

    \begin{subfigure}[b]{0.32\textwidth}
        \centering
        \includegraphics[width=\textwidth]{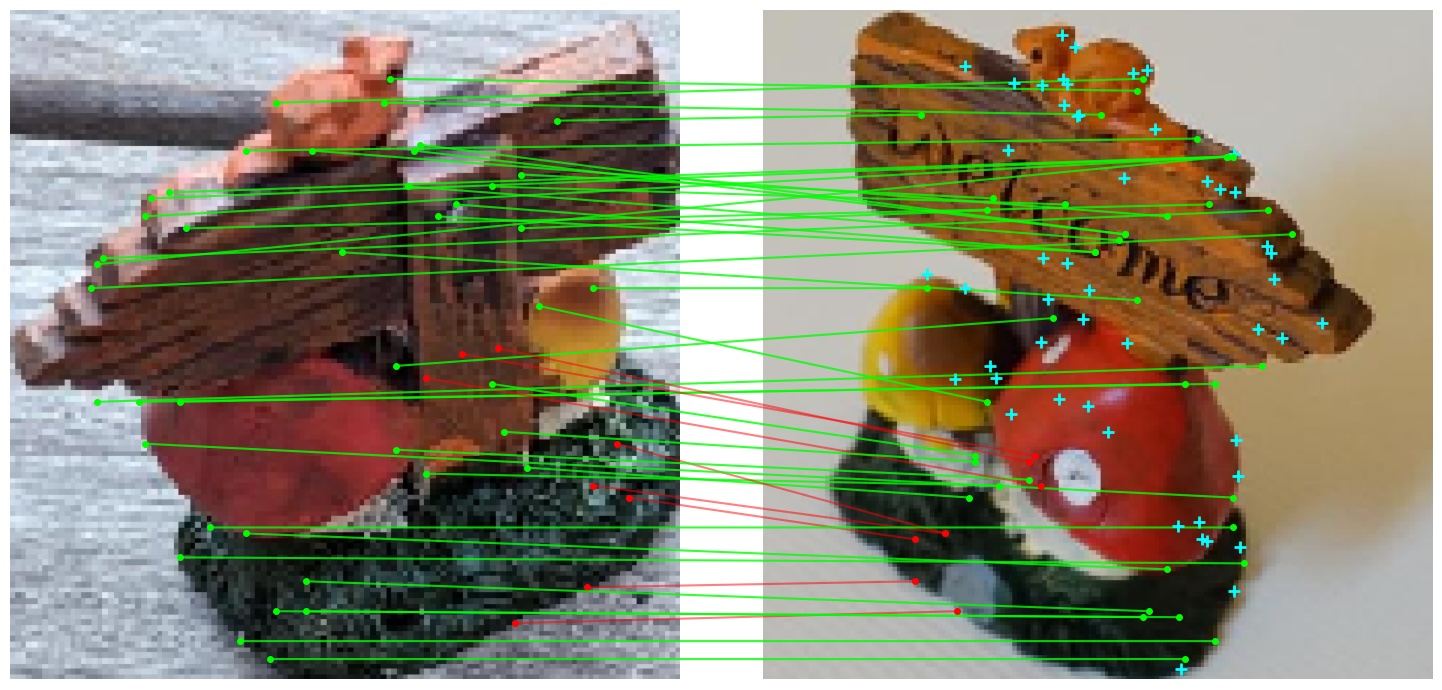}
    \end{subfigure}
    \hfill
    \begin{subfigure}[b]{0.32\textwidth}
        \centering
        \includegraphics[width=\textwidth]{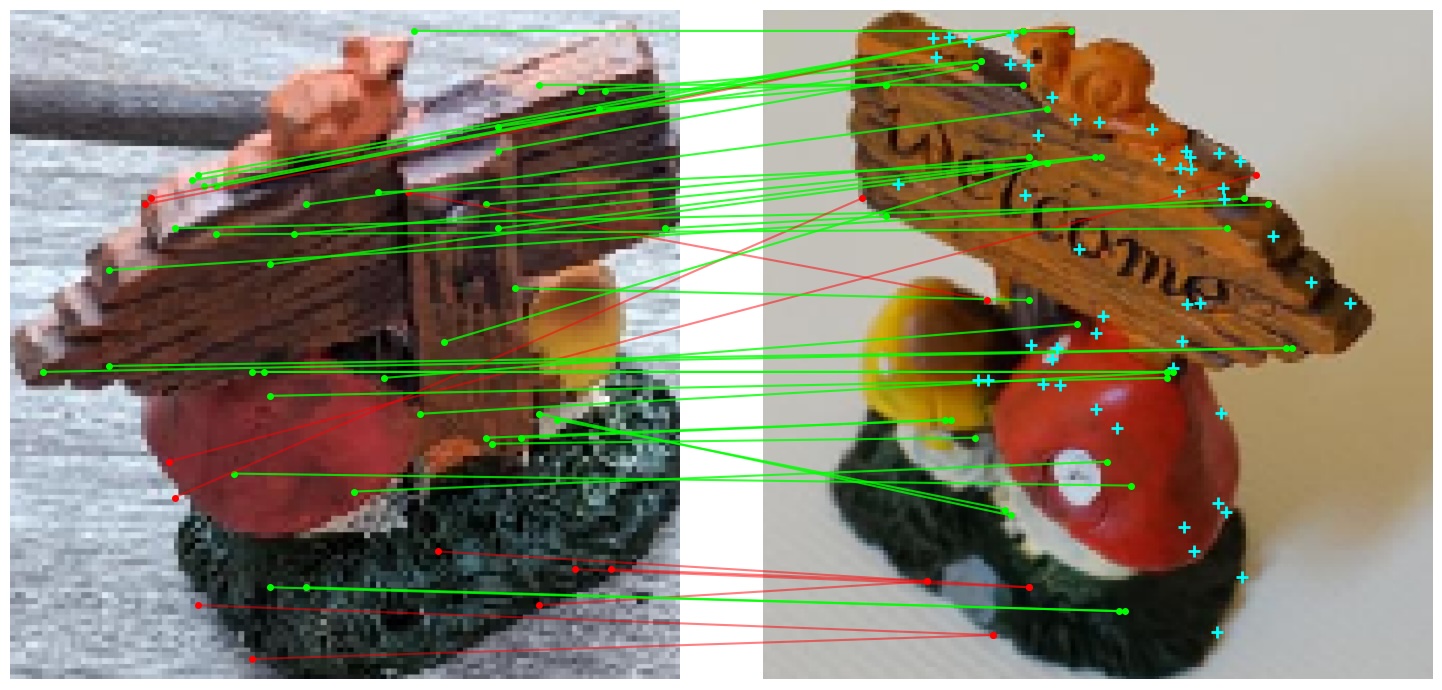}
    \end{subfigure}
    \hfill
    \begin{subfigure}[b]{0.32\textwidth}
        \centering
        \includegraphics[width=\textwidth]{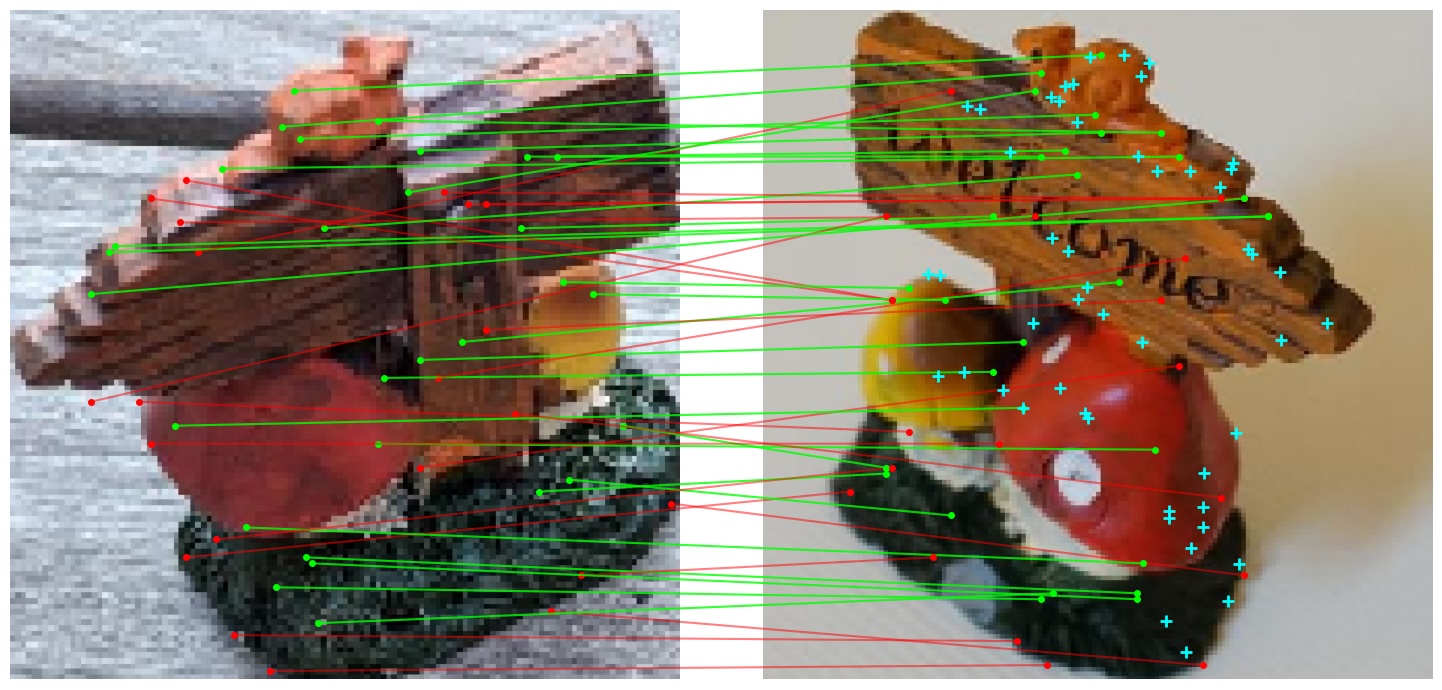}
    \end{subfigure}

    \caption{Qualitative comparison of 3D correspondences. DINOcular (ours) demonstrates superior robustness to large viewpoint variations, yielding more accurate matches (green) with significantly fewer outliers (red) than DINO and DUNE.}
    \label{fig:corr}
\end{figure}

\textbf{3D Correspondence and Feature Consistency.}
To evaluate the model's geometric awareness and cross-view consistency, we visualize 3D correspondence matching on the NAVI dataset following the Probe3D protocol (Figure~\ref{fig:corr}). DINOcular demonstrates exceptional robustness to severe viewpoint variations, establishing dense and highly accurate matches, where baseline models such as DINO ViT-B and DUNE struggle.

\begin{figure}[h]
    \centering
    \begin{minipage}{0.8\textwidth}
        \begin{subfigure}[b]{0.24\linewidth}
            \centering
            \textbf{Source}\\[1ex]
            \includegraphics[width=\textwidth]{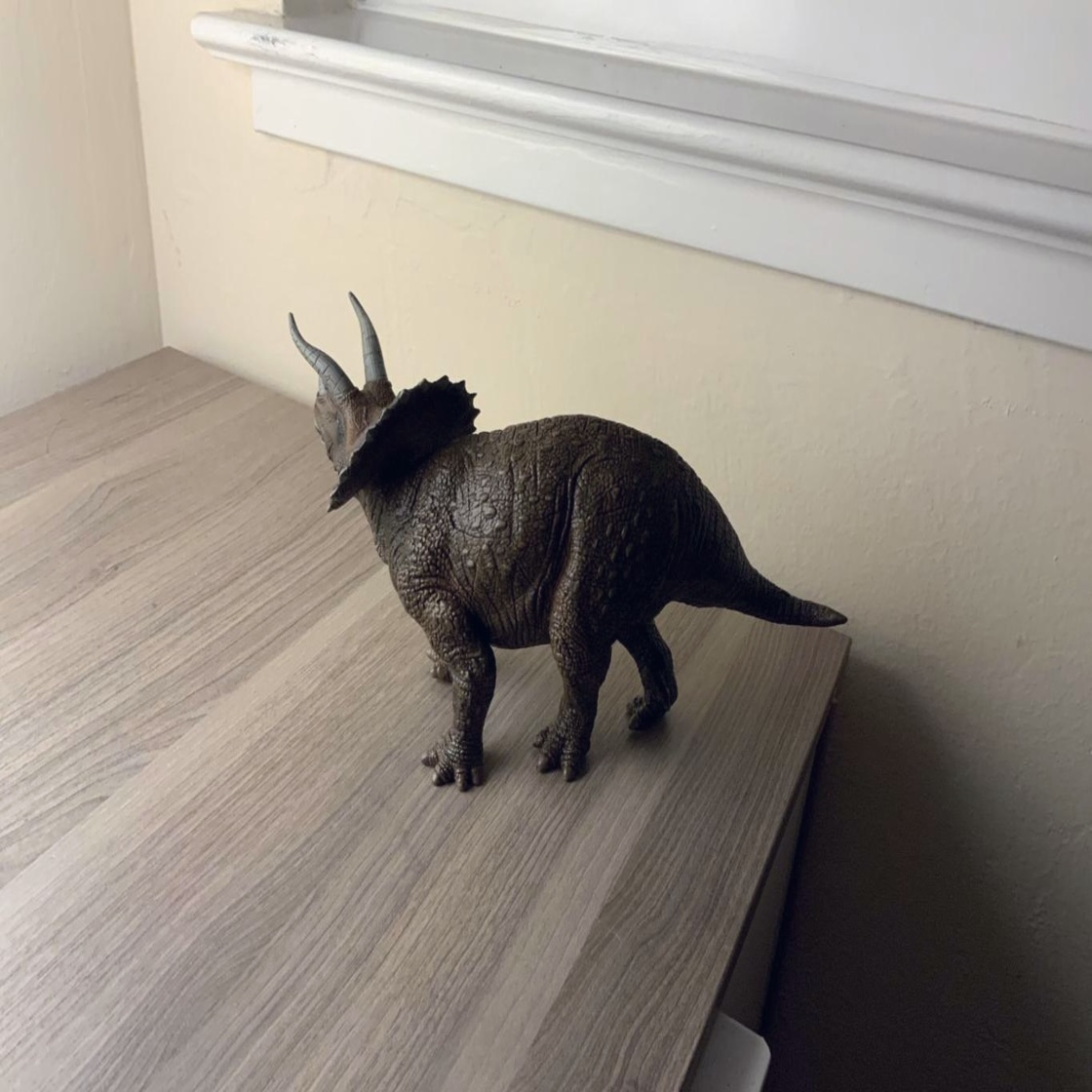}
        \end{subfigure}
        \hfill
        \begin{subfigure}[b]{0.24\linewidth}
            \centering
            \textbf{DINOcular}\\[1ex]
            \includegraphics[width=\textwidth]{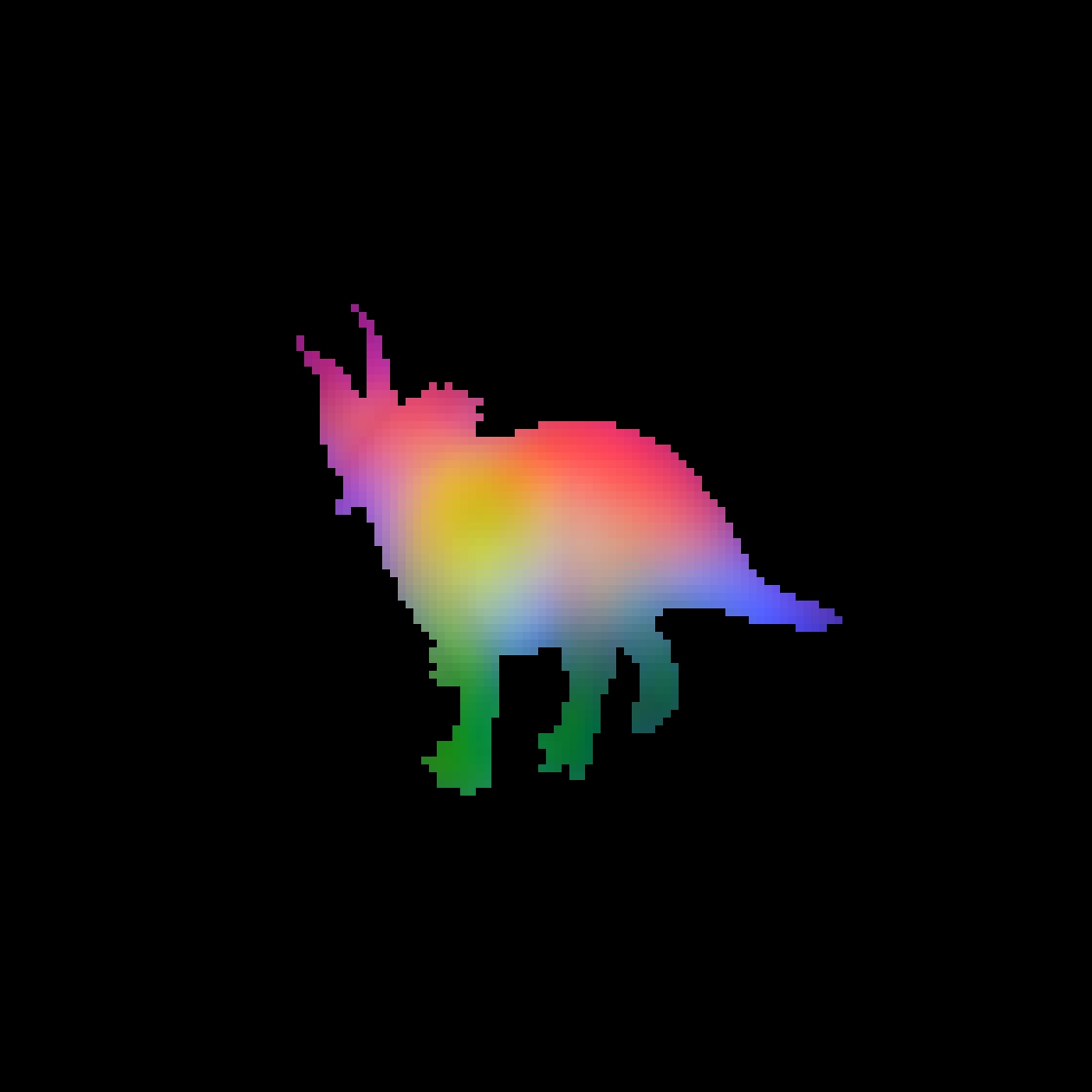}
        \end{subfigure}
        \hfill
        \begin{subfigure}[b]{0.24\linewidth}
            \centering
            \textbf{DINO}\\[1ex]
            \includegraphics[width=\textwidth]{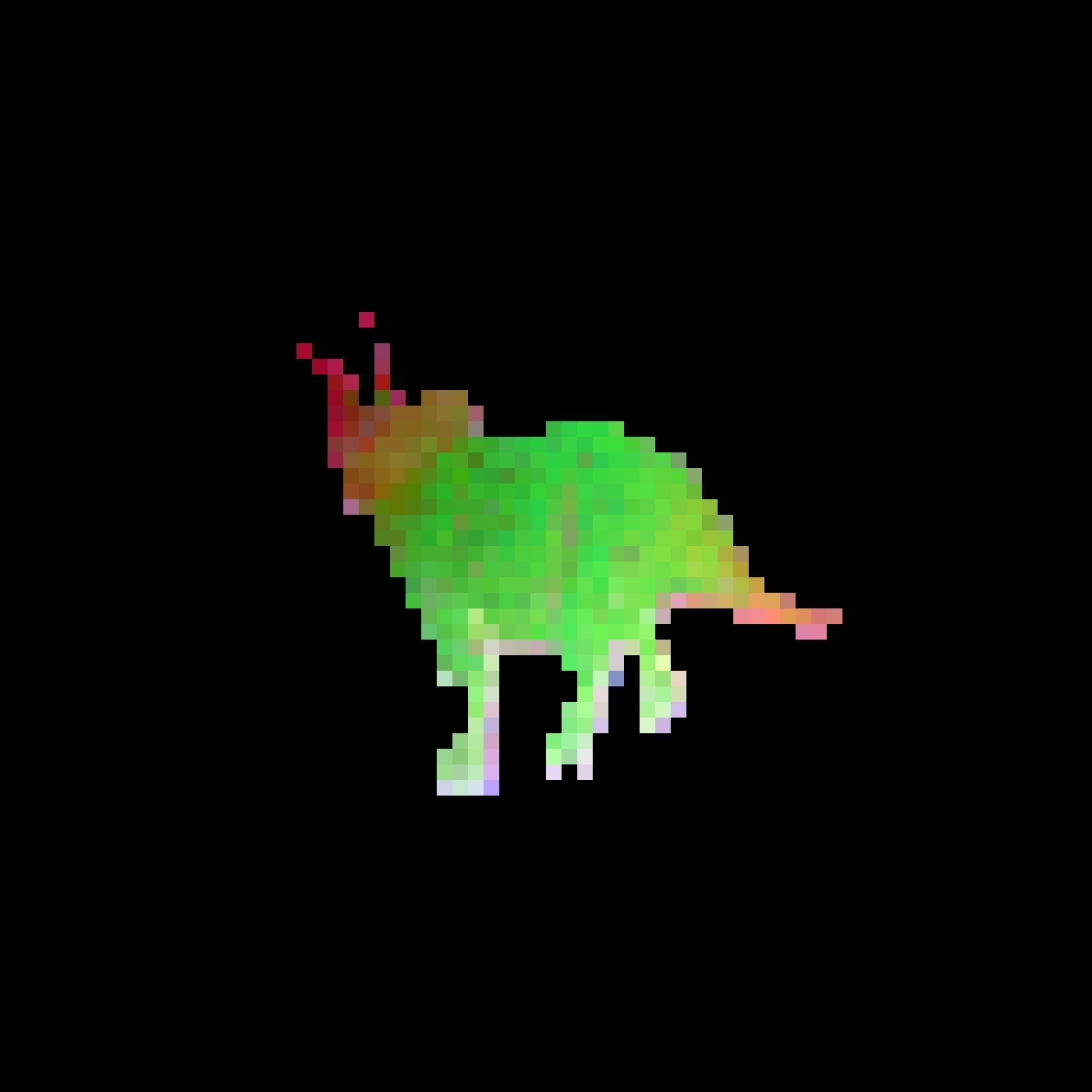}
        \end{subfigure}
        \hfill
        \begin{subfigure}[b]{0.24\linewidth}
            \centering
            \textbf{DUNE}\\[1ex]
            \includegraphics[width=\textwidth]{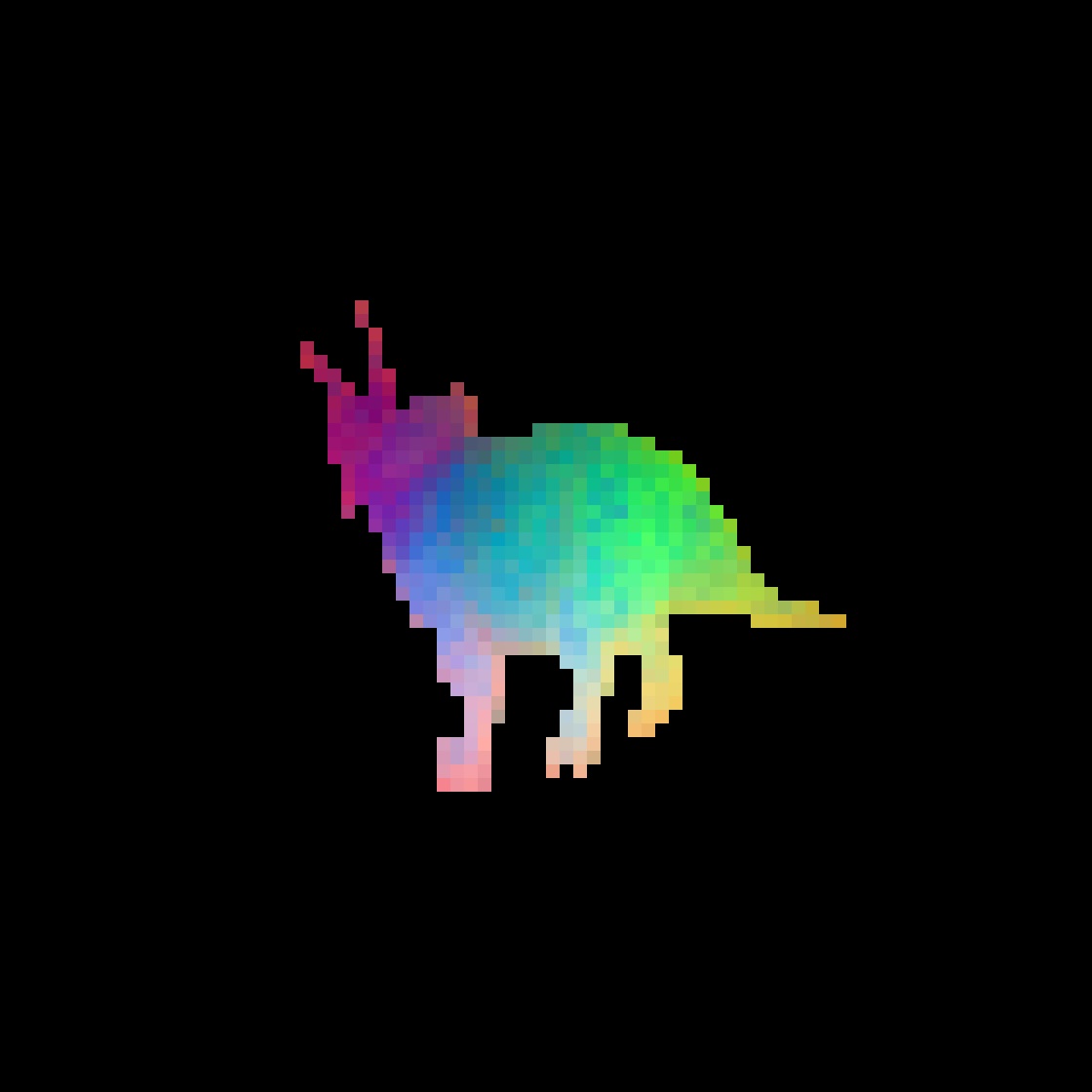}
        \end{subfigure}
        \vspace{0.1cm}

        \begin{subfigure}[b]{0.24\linewidth}
            \centering
            \includegraphics[width=\textwidth]{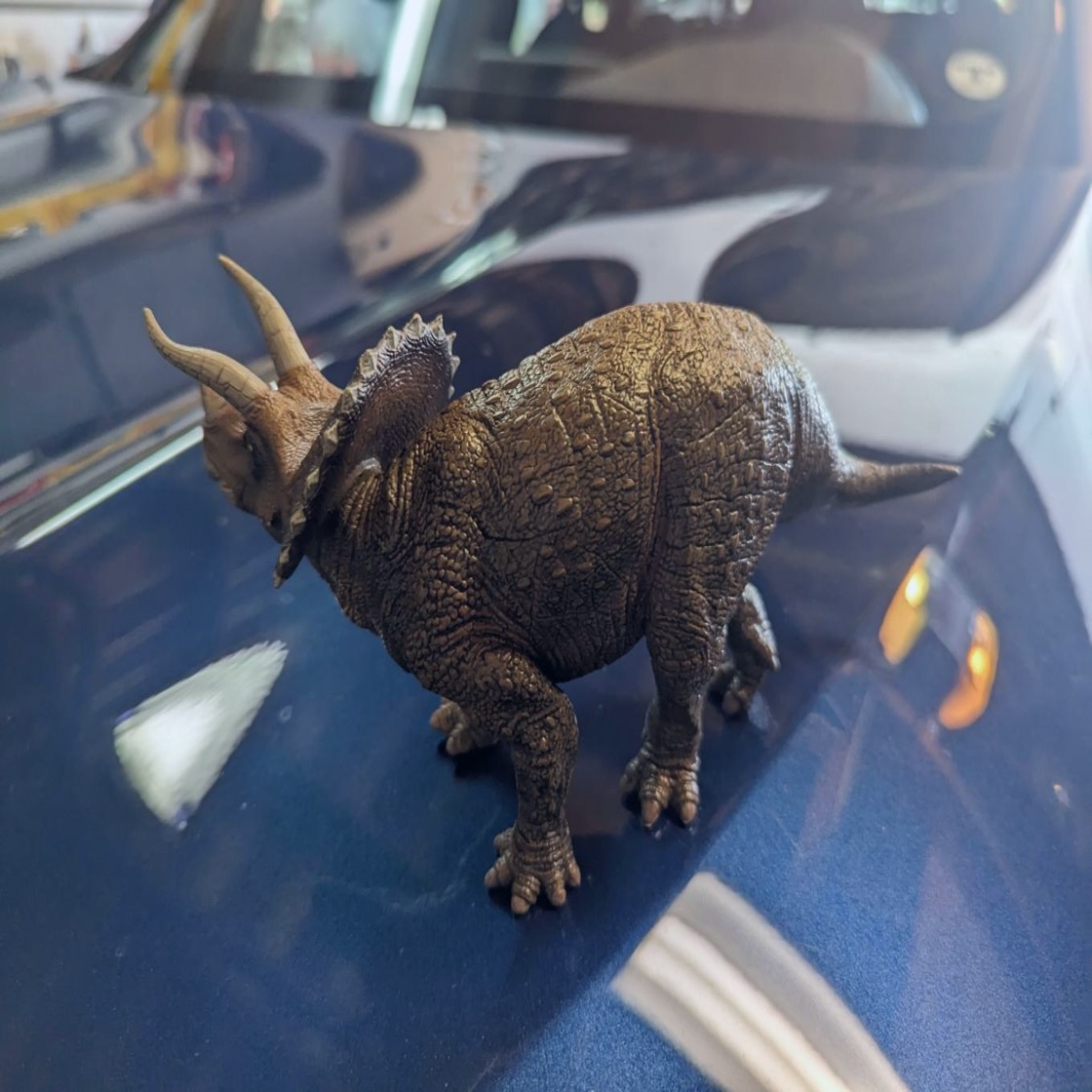}
        \end{subfigure}
        \hfill
        \begin{subfigure}[b]{0.24\linewidth}
            \centering
            \includegraphics[width=\textwidth]{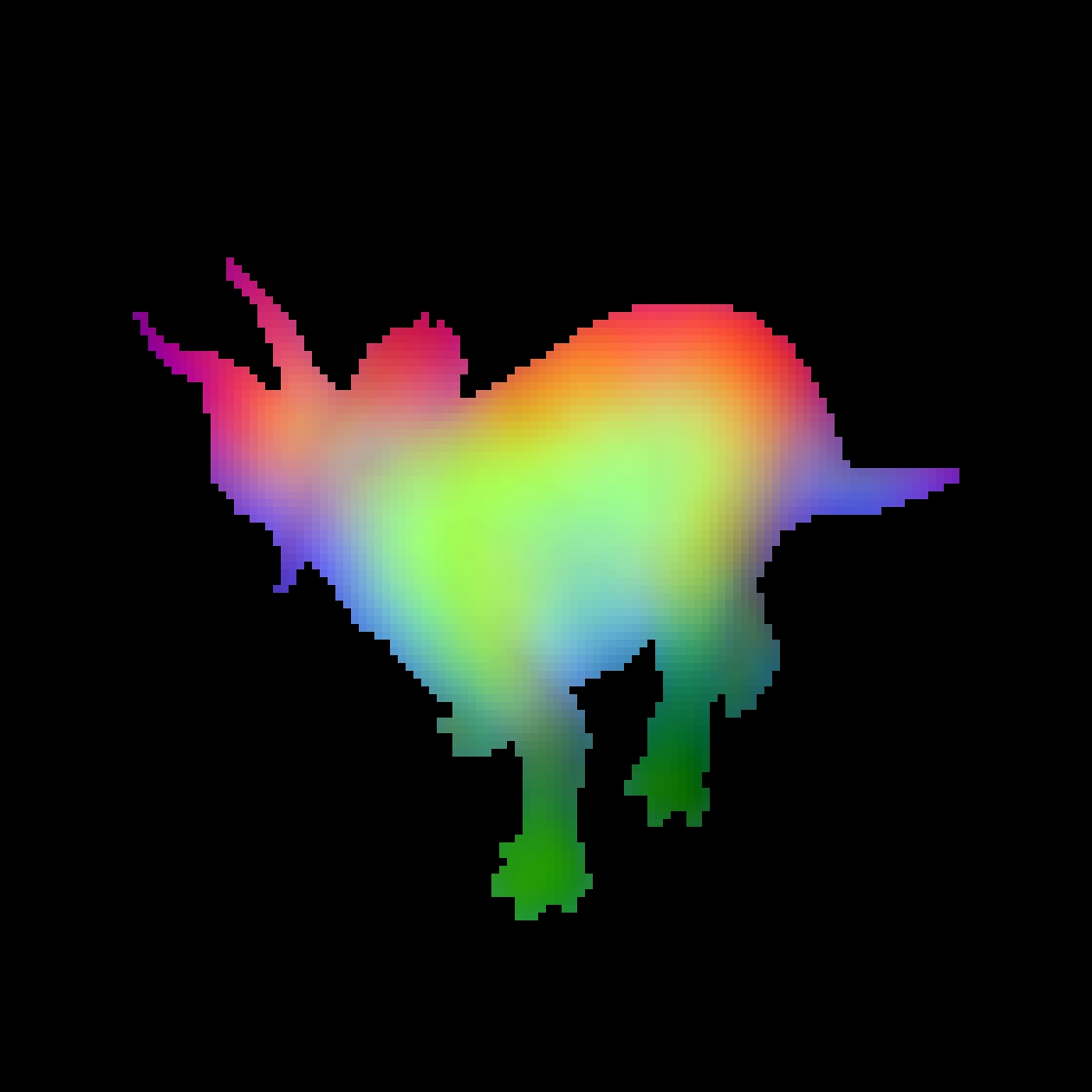}
        \end{subfigure}
        \hfill
        \begin{subfigure}[b]{0.24\linewidth}
            \centering
            \includegraphics[width=\textwidth]{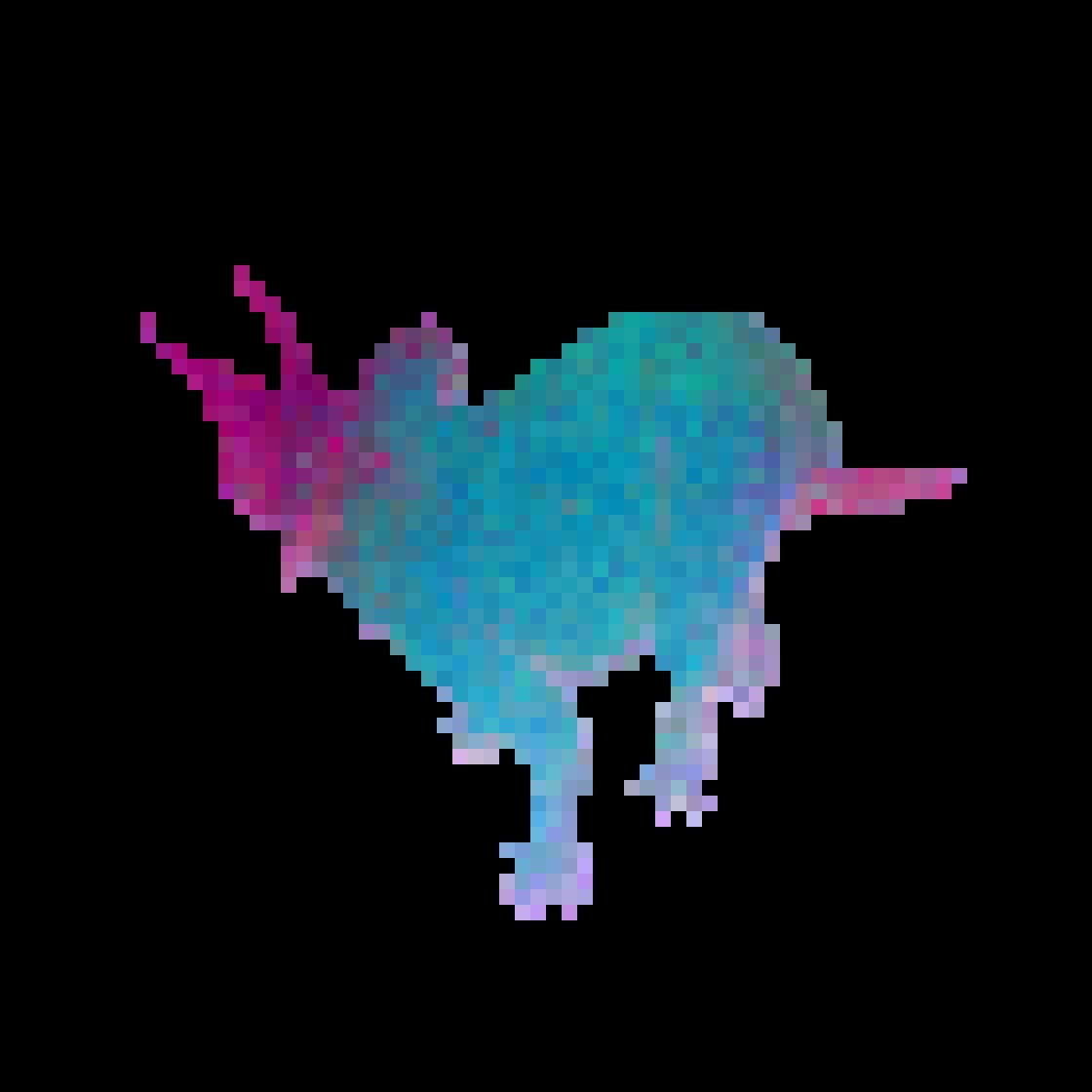}
        \end{subfigure}
        \hfill
        \begin{subfigure}[b]{0.24\linewidth}
            \centering
            \includegraphics[width=\textwidth]{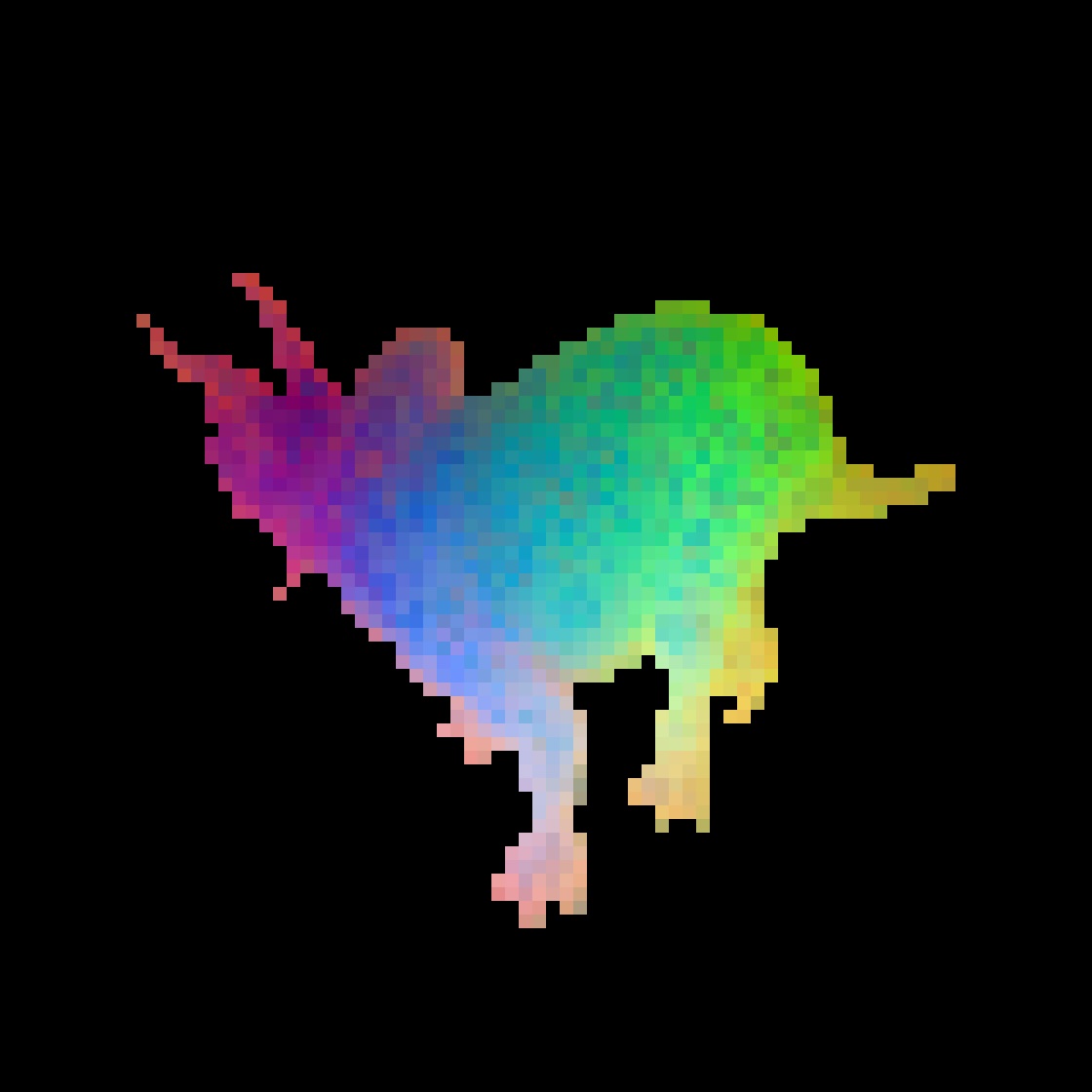}
        \end{subfigure}
        \vspace{0.1cm}

        \begin{subfigure}[b]{0.24\linewidth}
            \centering
            \includegraphics[width=\textwidth]{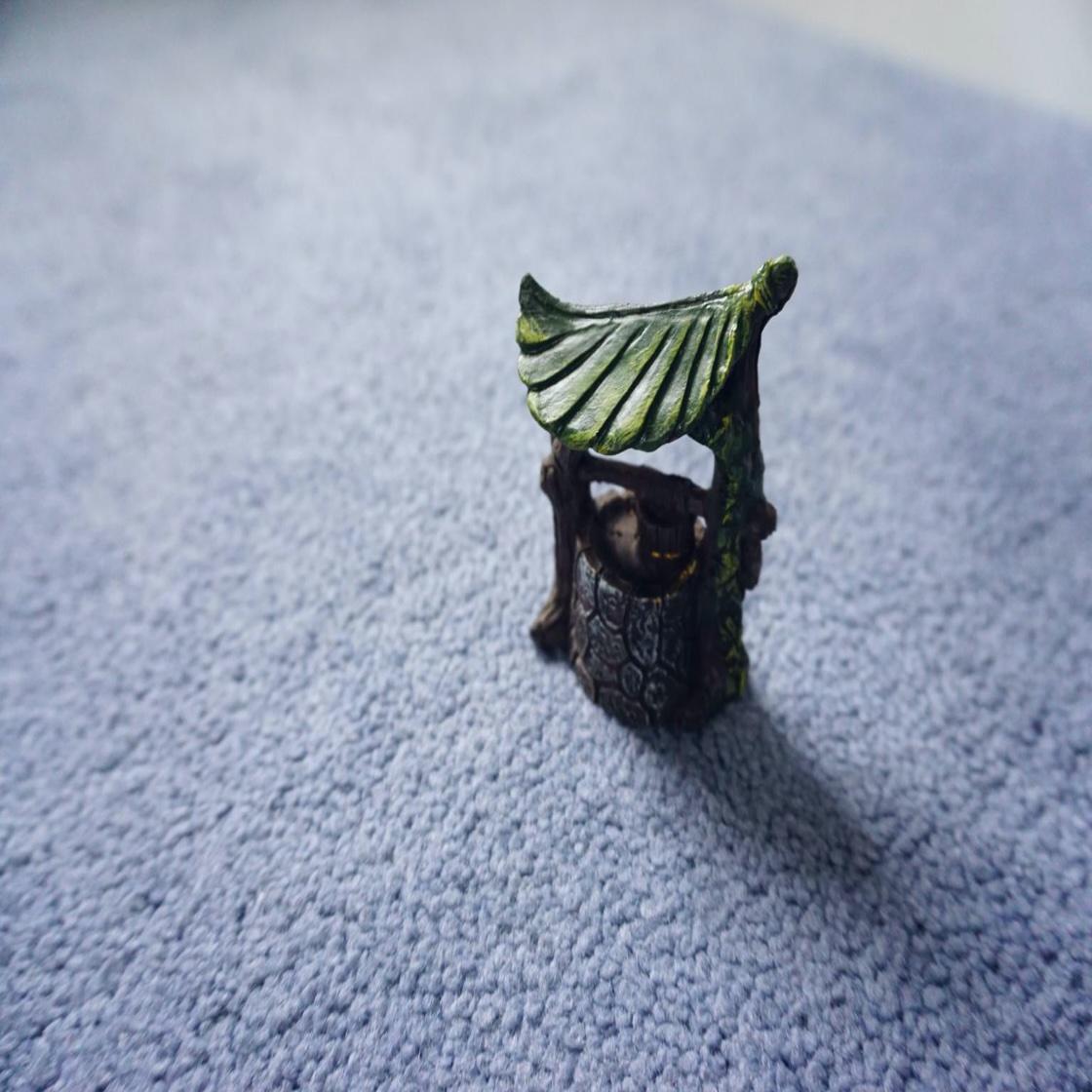}
        \end{subfigure}
        \hfill
        \begin{subfigure}[b]{0.24\linewidth}
            \centering
            \includegraphics[width=\textwidth]{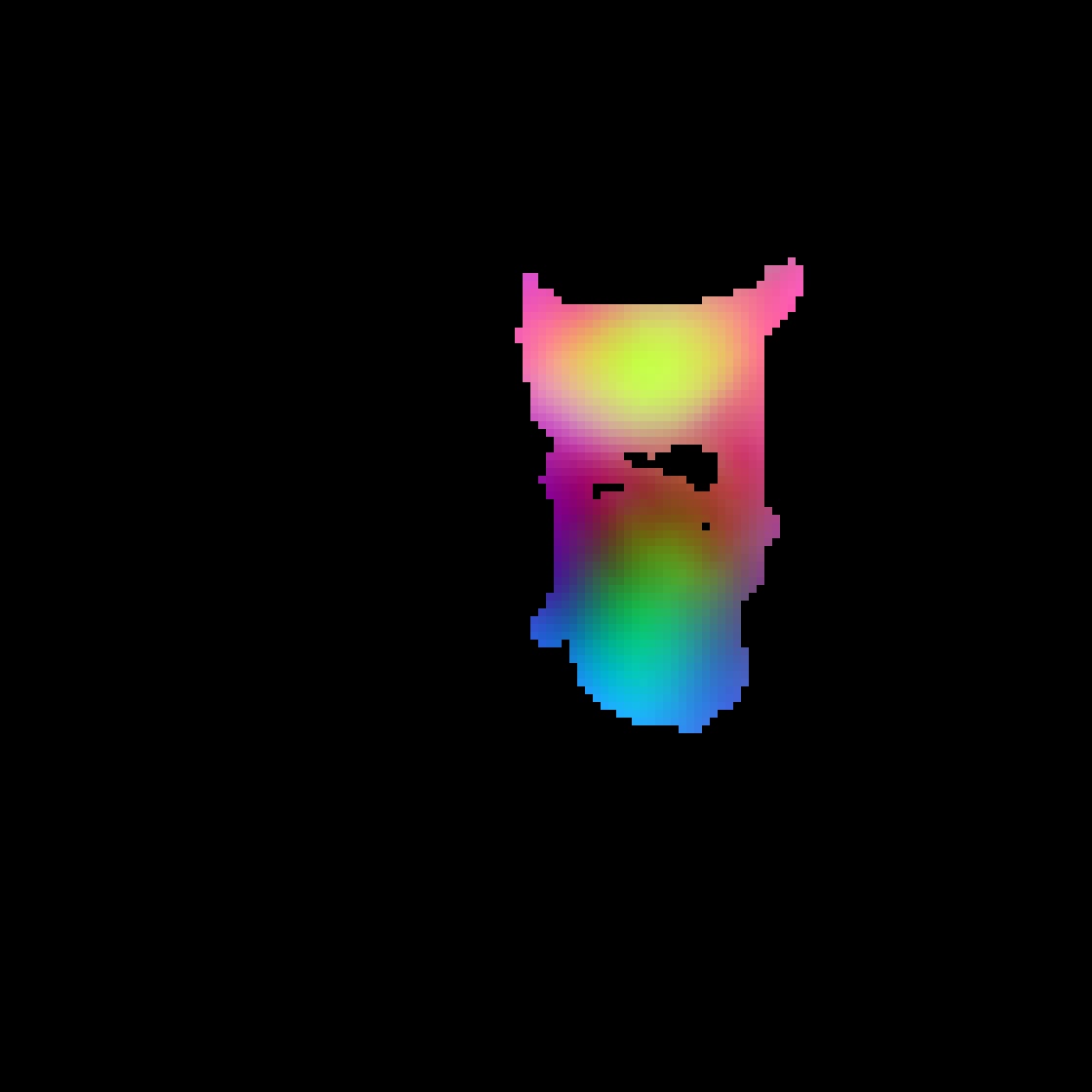}
        \end{subfigure}
        \hfill
        \begin{subfigure}[b]{0.24\linewidth}
            \centering
            \includegraphics[width=\textwidth]{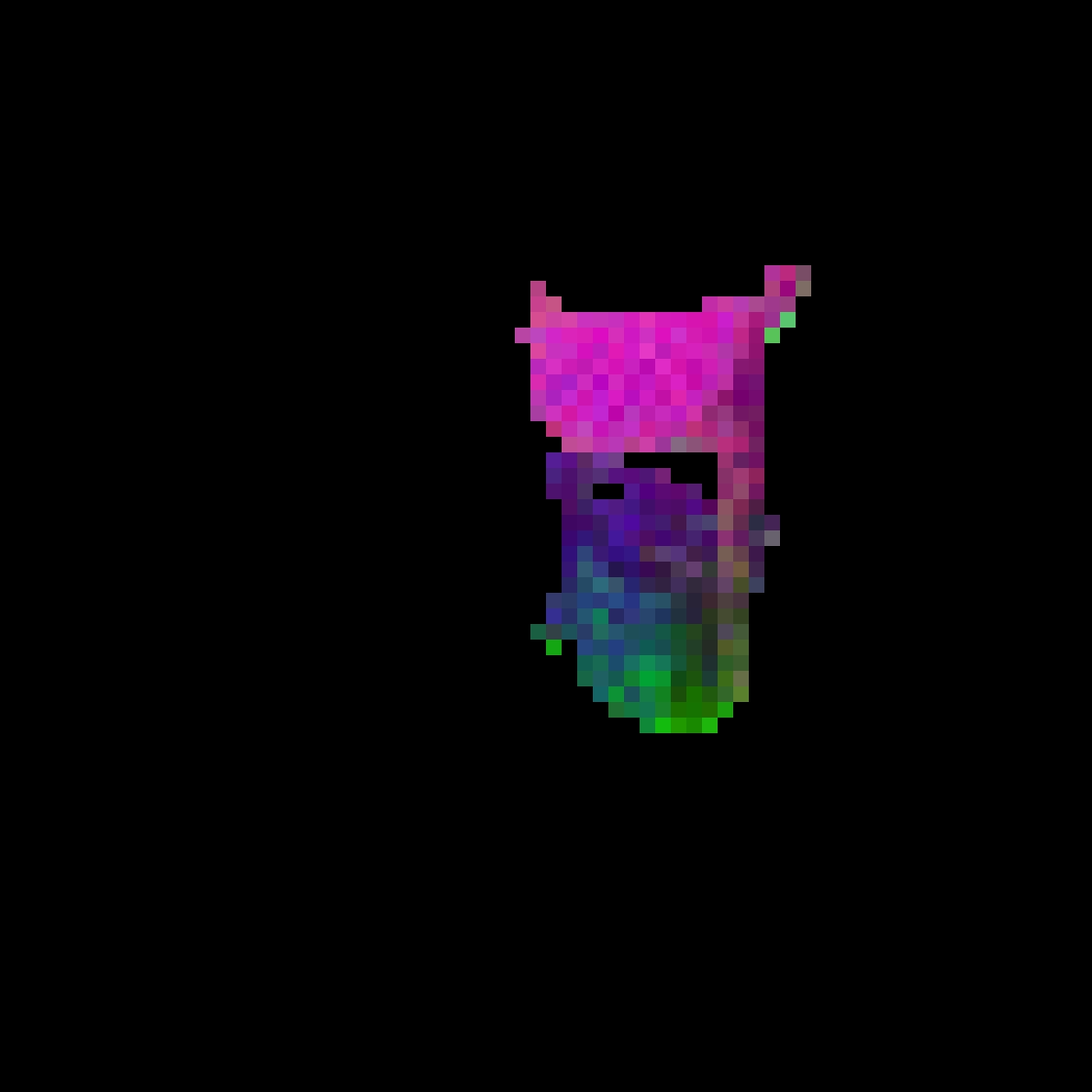}
        \end{subfigure}
        \hfill
        \begin{subfigure}[b]{0.24\linewidth}
            \centering
            \includegraphics[width=\textwidth]{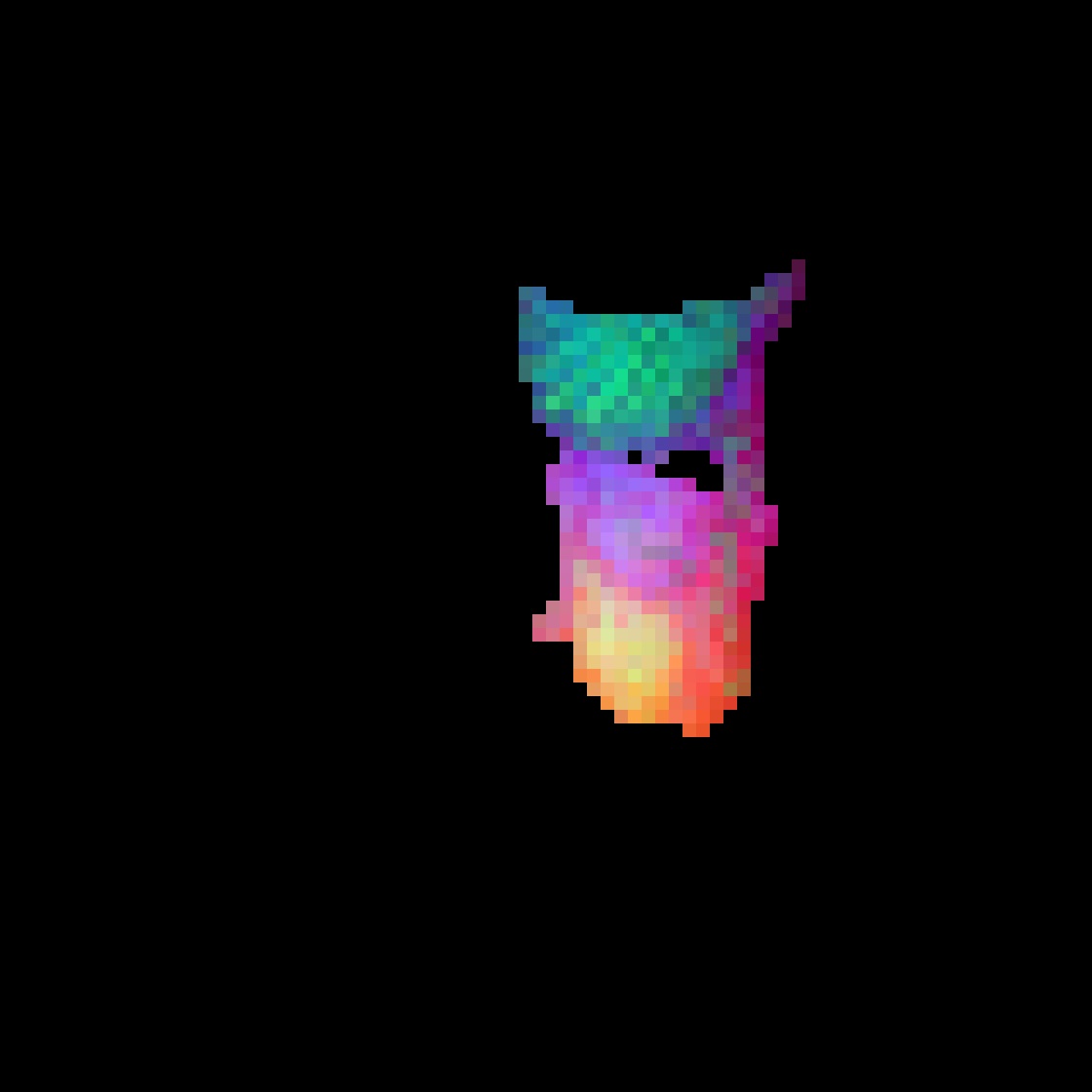}
        \end{subfigure}
        \vspace{0.1cm}

        \begin{subfigure}[b]{0.24\linewidth}
            \centering
            \includegraphics[width=\textwidth]{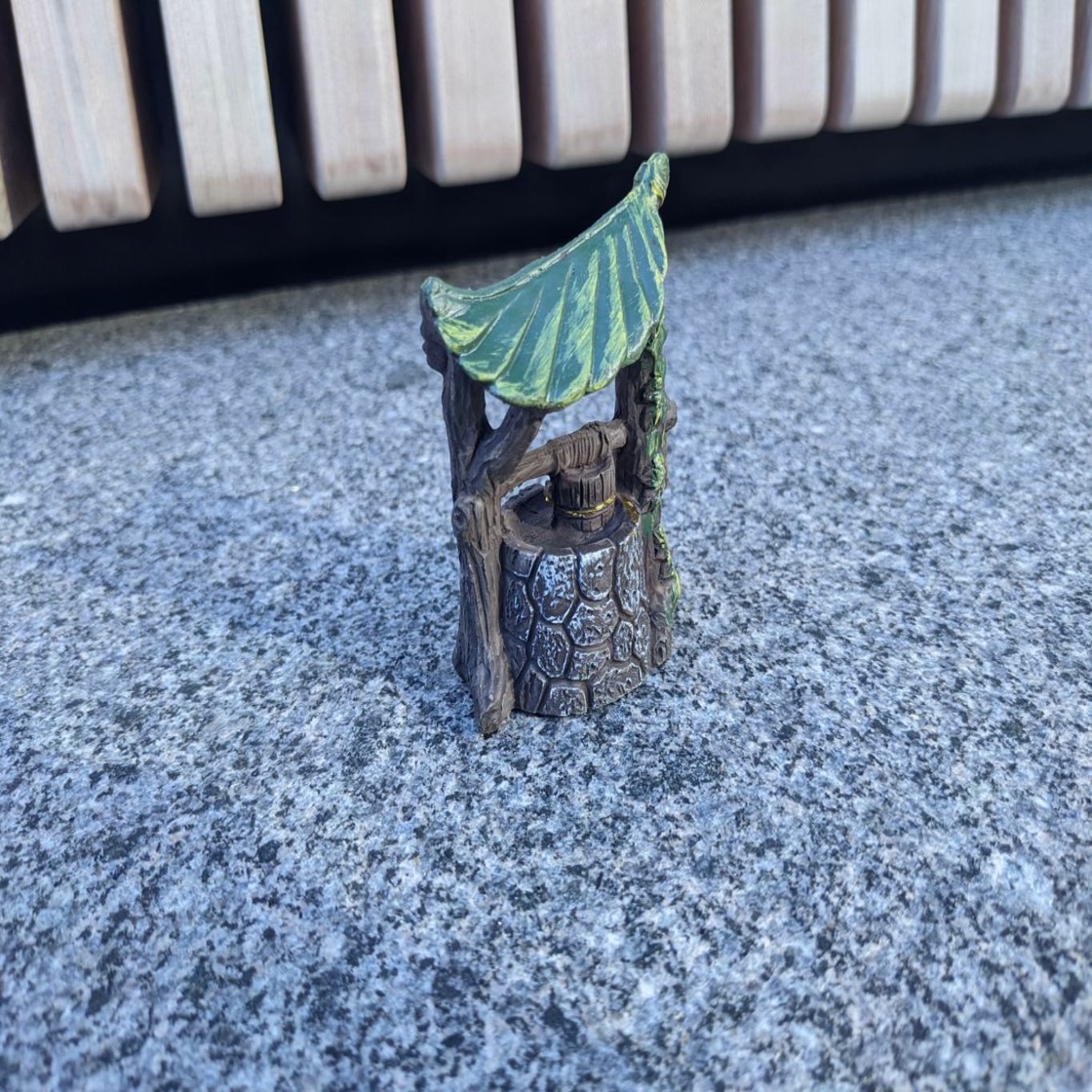}
        \end{subfigure}
        \hfill
        \begin{subfigure}[b]{0.24\linewidth}
            \centering
            \includegraphics[width=\textwidth]{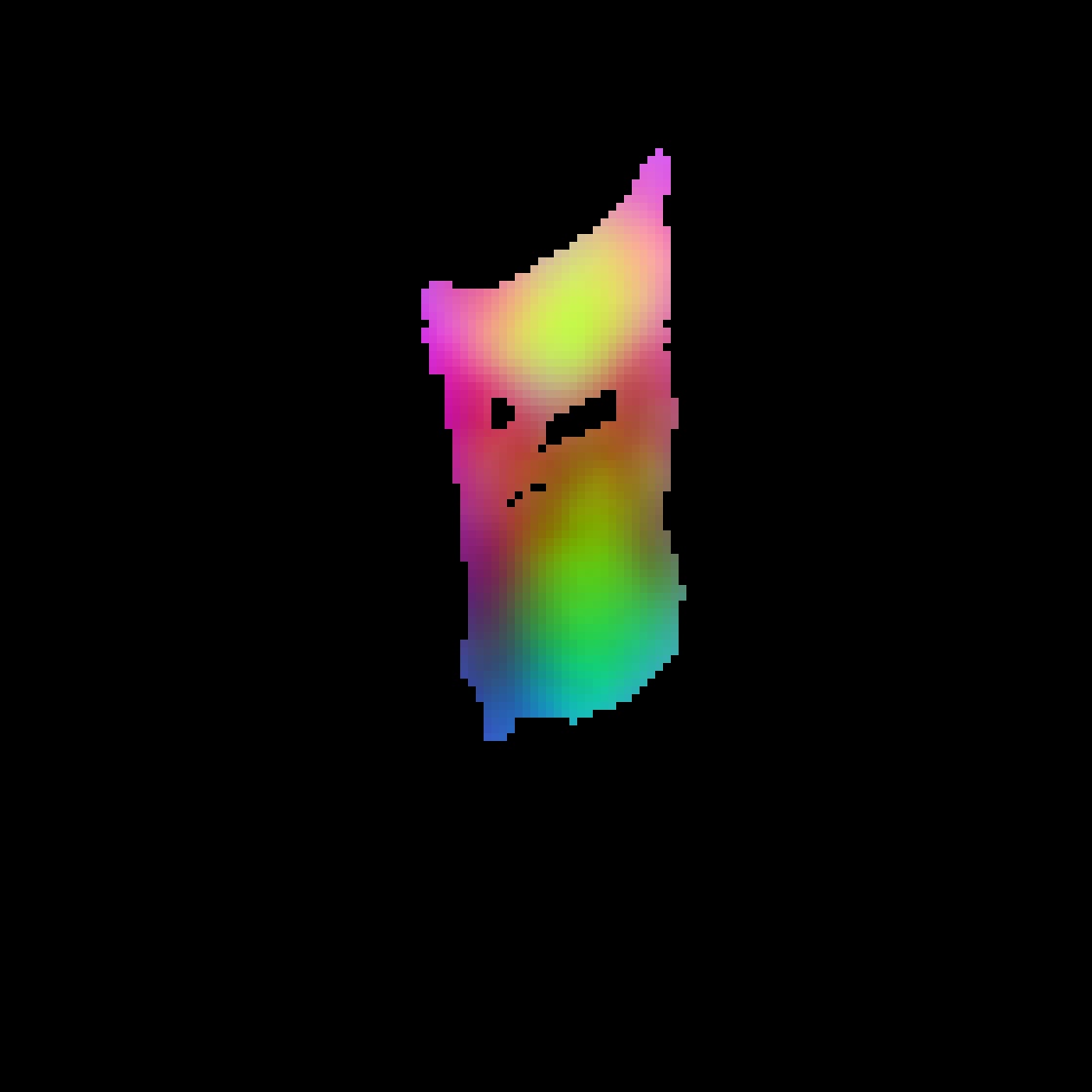}
        \end{subfigure}
        \hfill
        \begin{subfigure}[b]{0.24\linewidth}
            \centering
            \includegraphics[width=\textwidth]{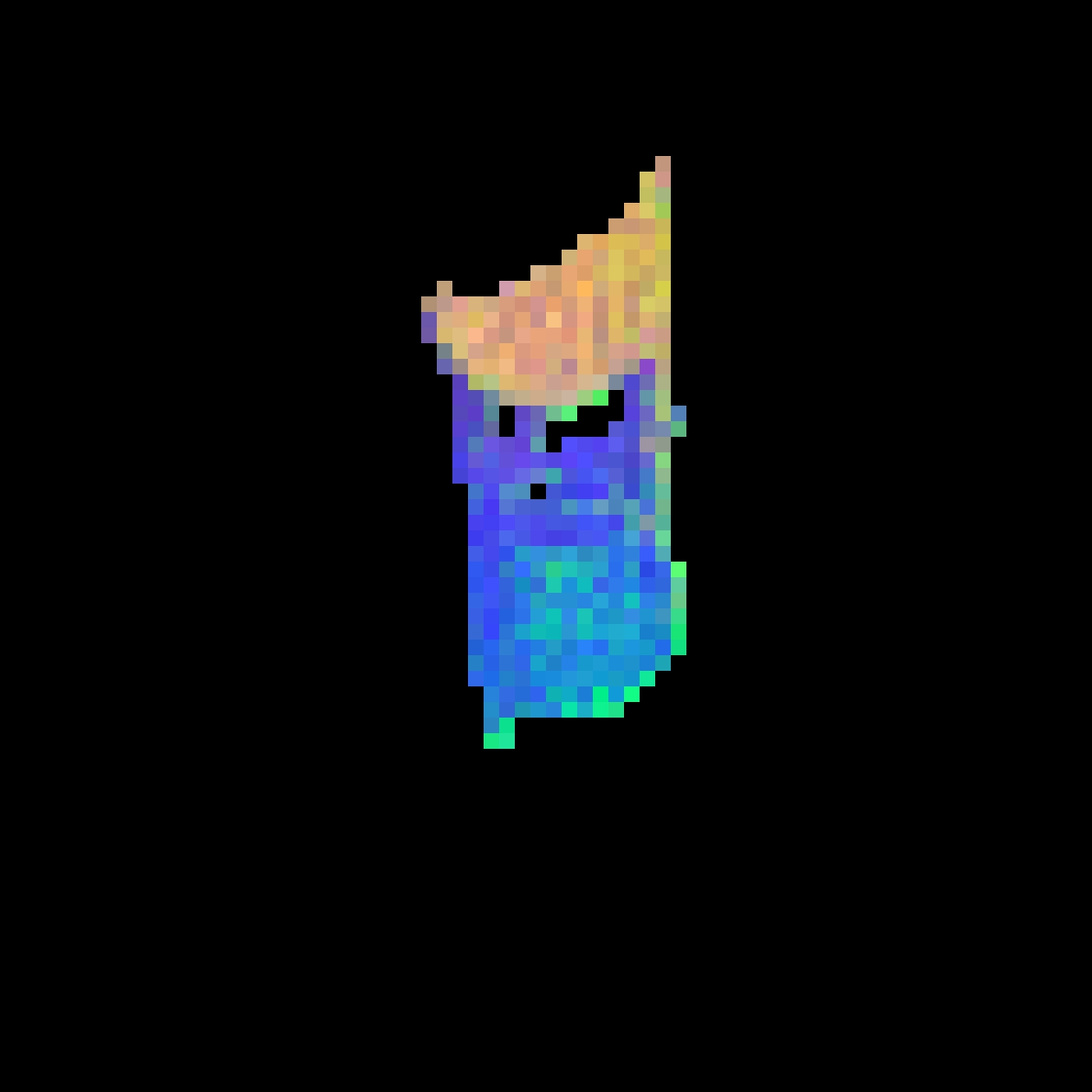}
        \end{subfigure}
        \hfill
        \begin{subfigure}[b]{0.24\linewidth}
            \centering
            \includegraphics[width=\textwidth]{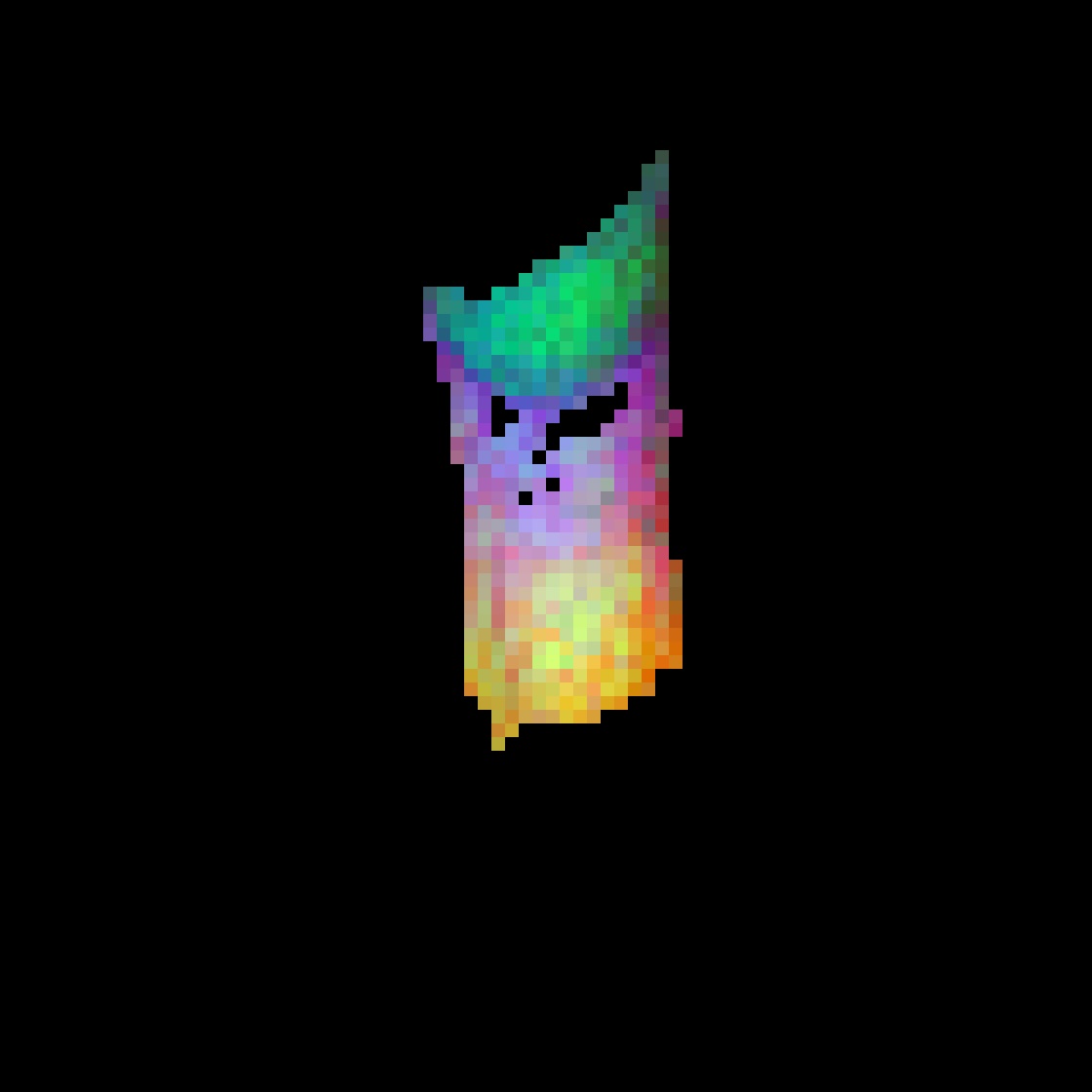}
        \end{subfigure}
        \vspace{0.1cm}

        \begin{subfigure}[b]{0.24\linewidth}
            \centering
            \includegraphics[width=\textwidth]{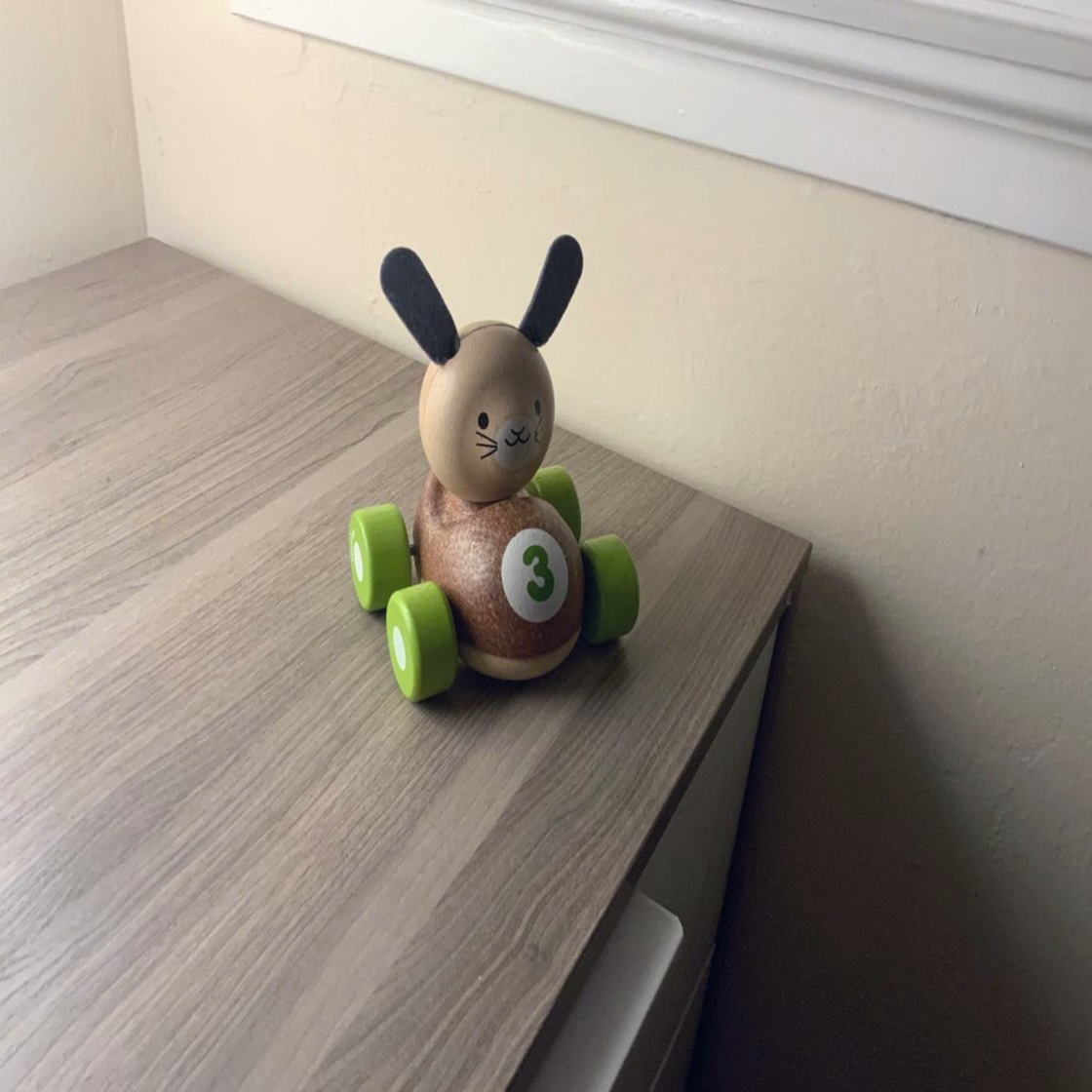}
        \end{subfigure}
        \hfill
        \begin{subfigure}[b]{0.24\linewidth}
            \centering
            \includegraphics[width=\textwidth]{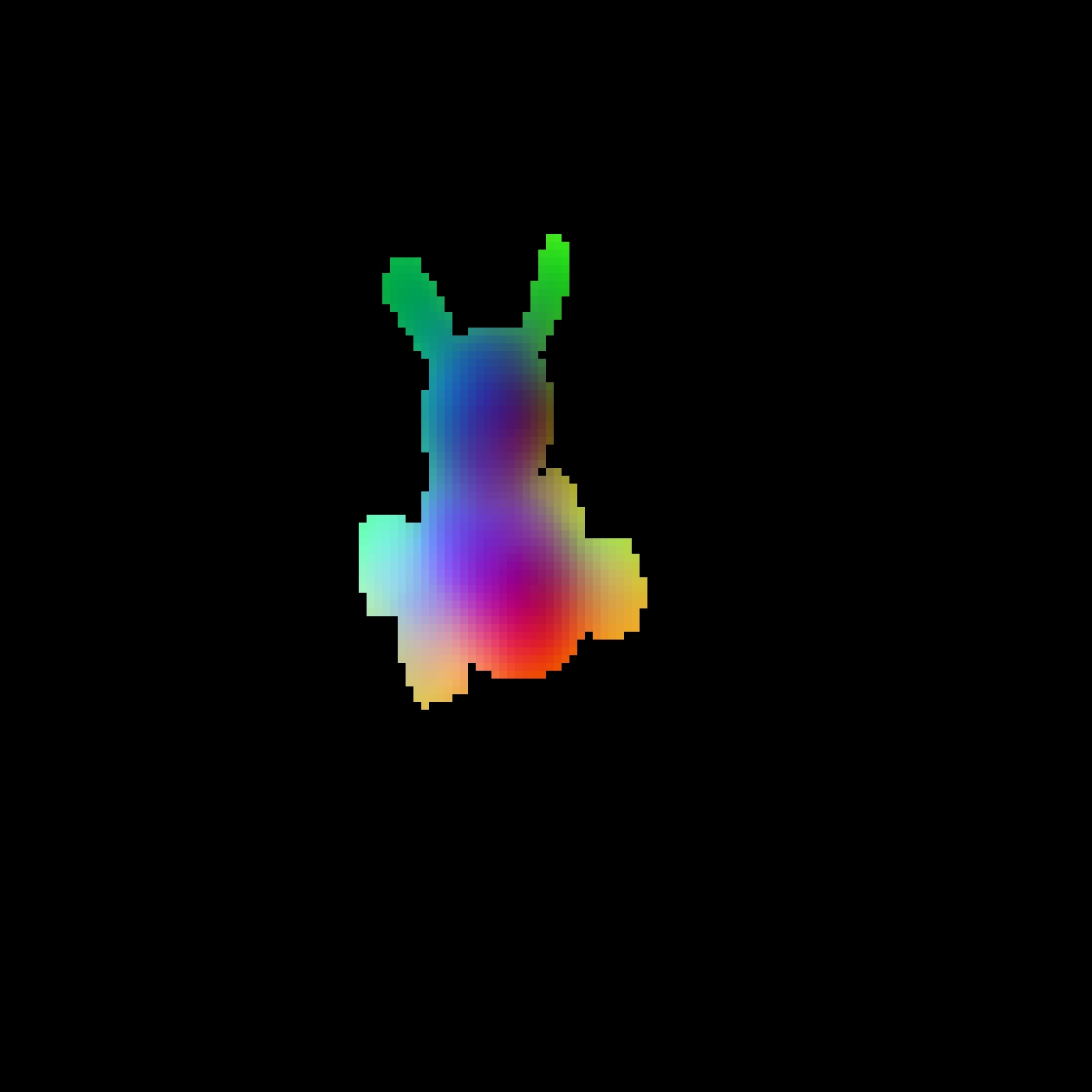}
        \end{subfigure}
        \hfill
        \begin{subfigure}[b]{0.24\linewidth}
            \centering
            \includegraphics[width=\textwidth]{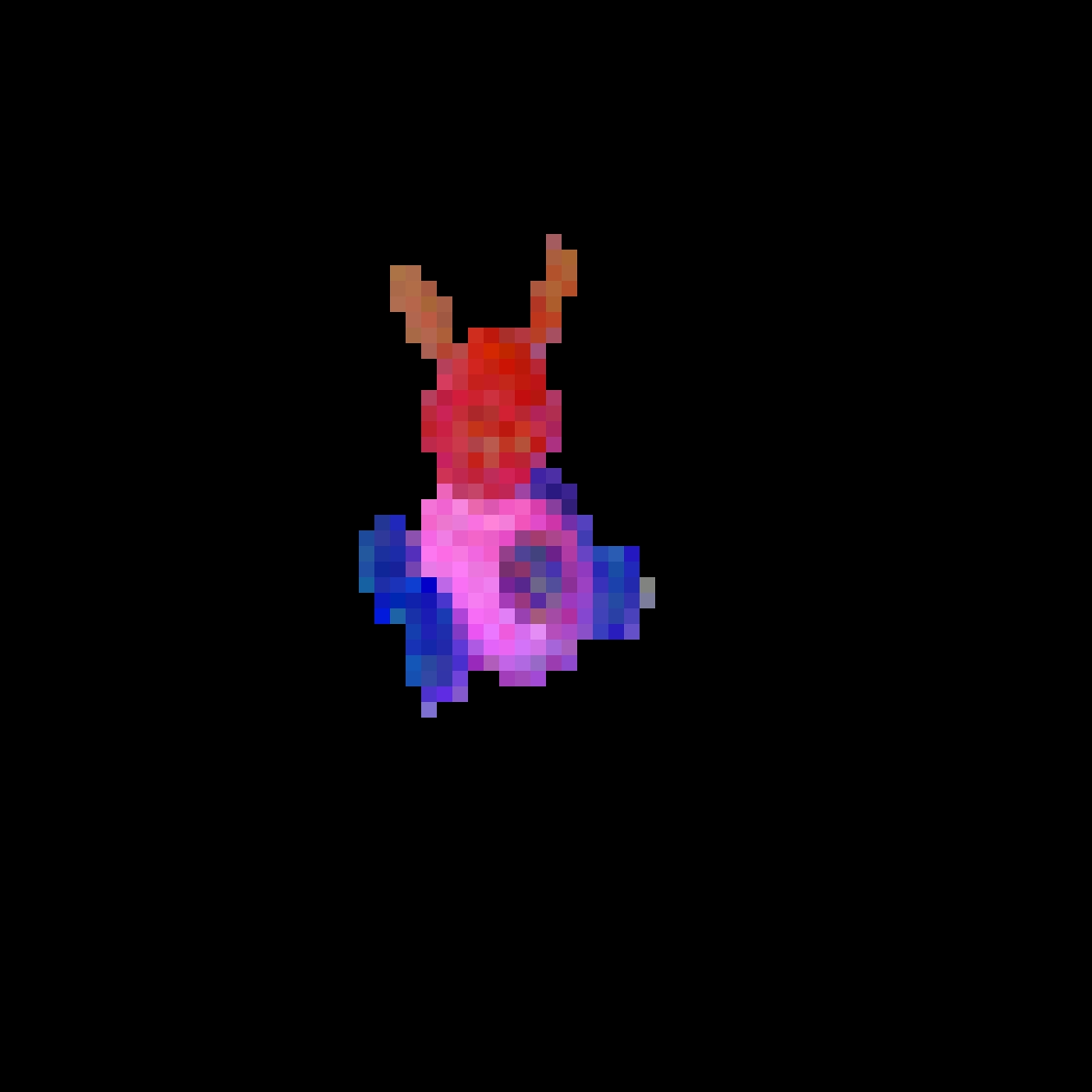}
        \end{subfigure}
        \hfill
        \begin{subfigure}[b]{0.24\linewidth}
            \centering
            \includegraphics[width=\textwidth]{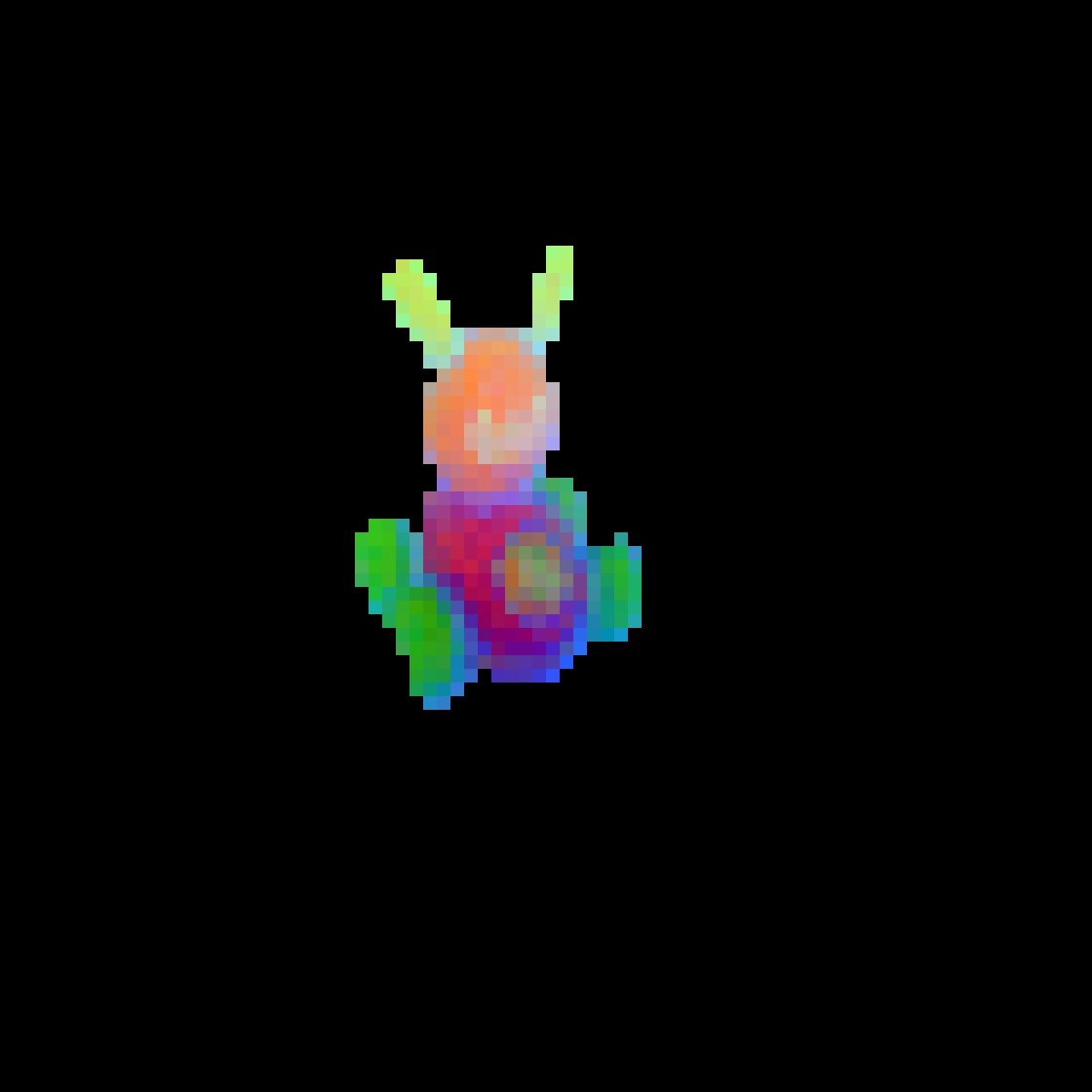}
        \end{subfigure}
        \vspace{0.1cm}

        \begin{subfigure}[b]{0.24\linewidth}
            \centering
            \includegraphics[width=\textwidth]{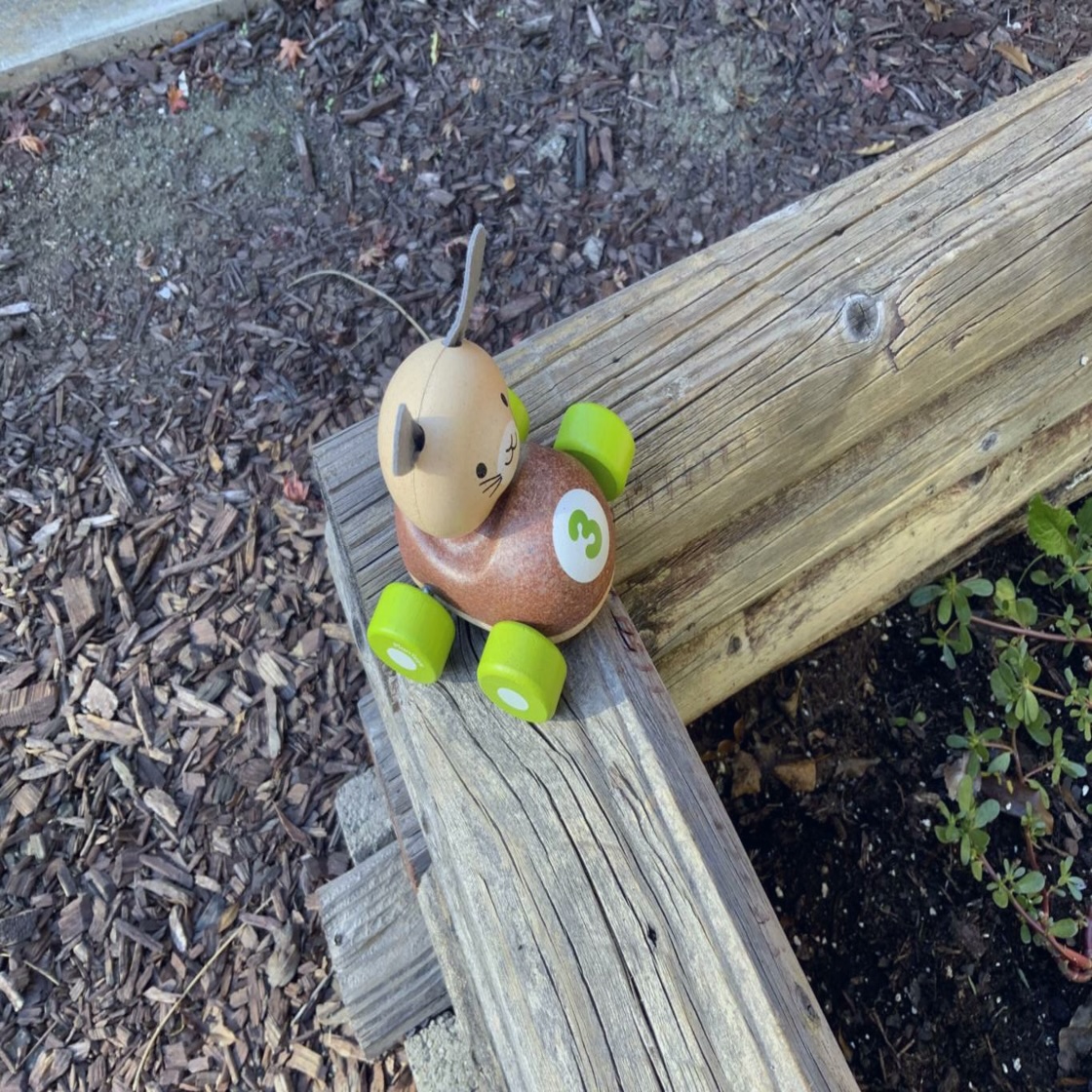}
        \end{subfigure}
        \hfill
        \begin{subfigure}[b]{0.24\linewidth}
            \centering
            \includegraphics[width=\textwidth]{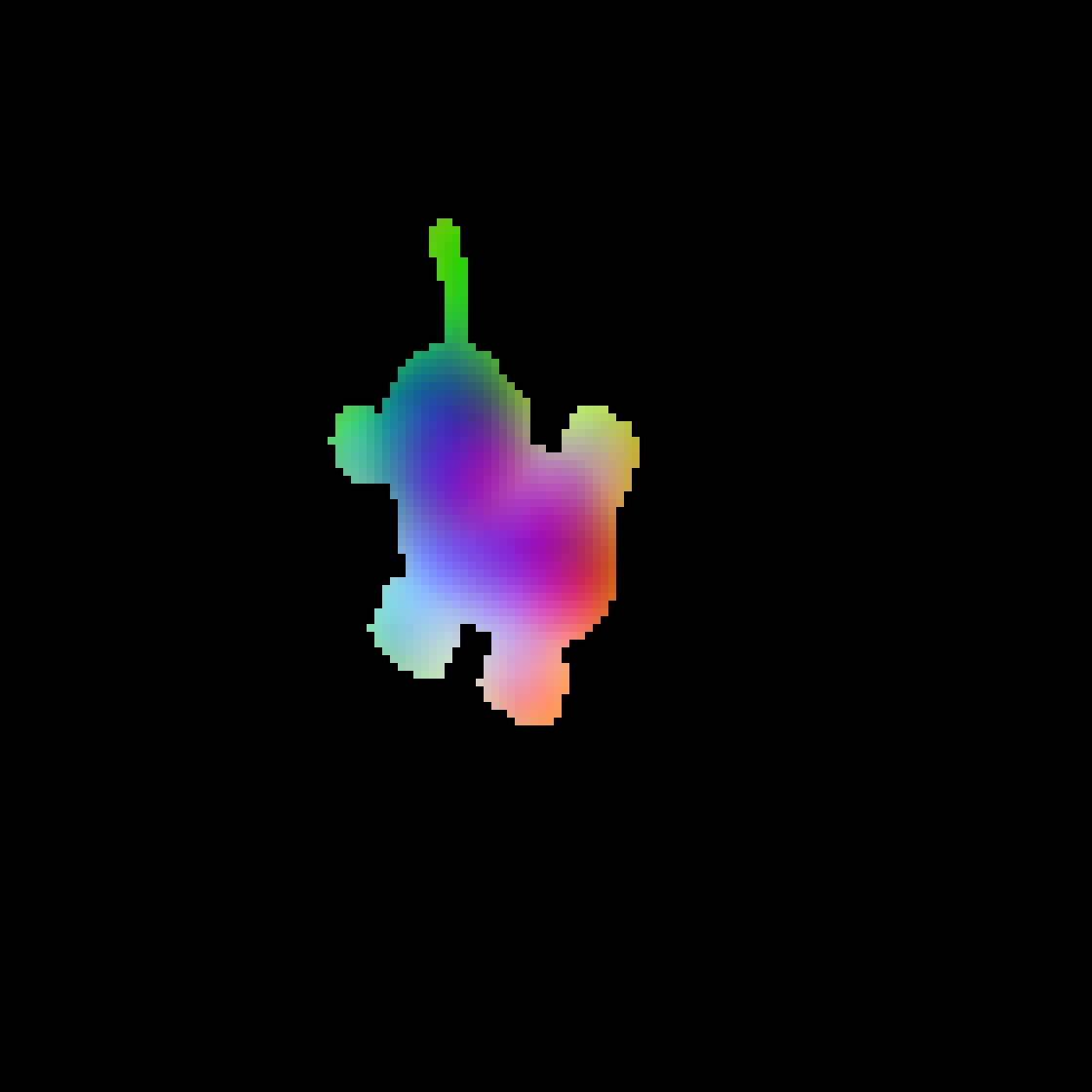}
        \end{subfigure}
        \hfill
        \begin{subfigure}[b]{0.24\linewidth}
            \centering
            \includegraphics[width=\textwidth]{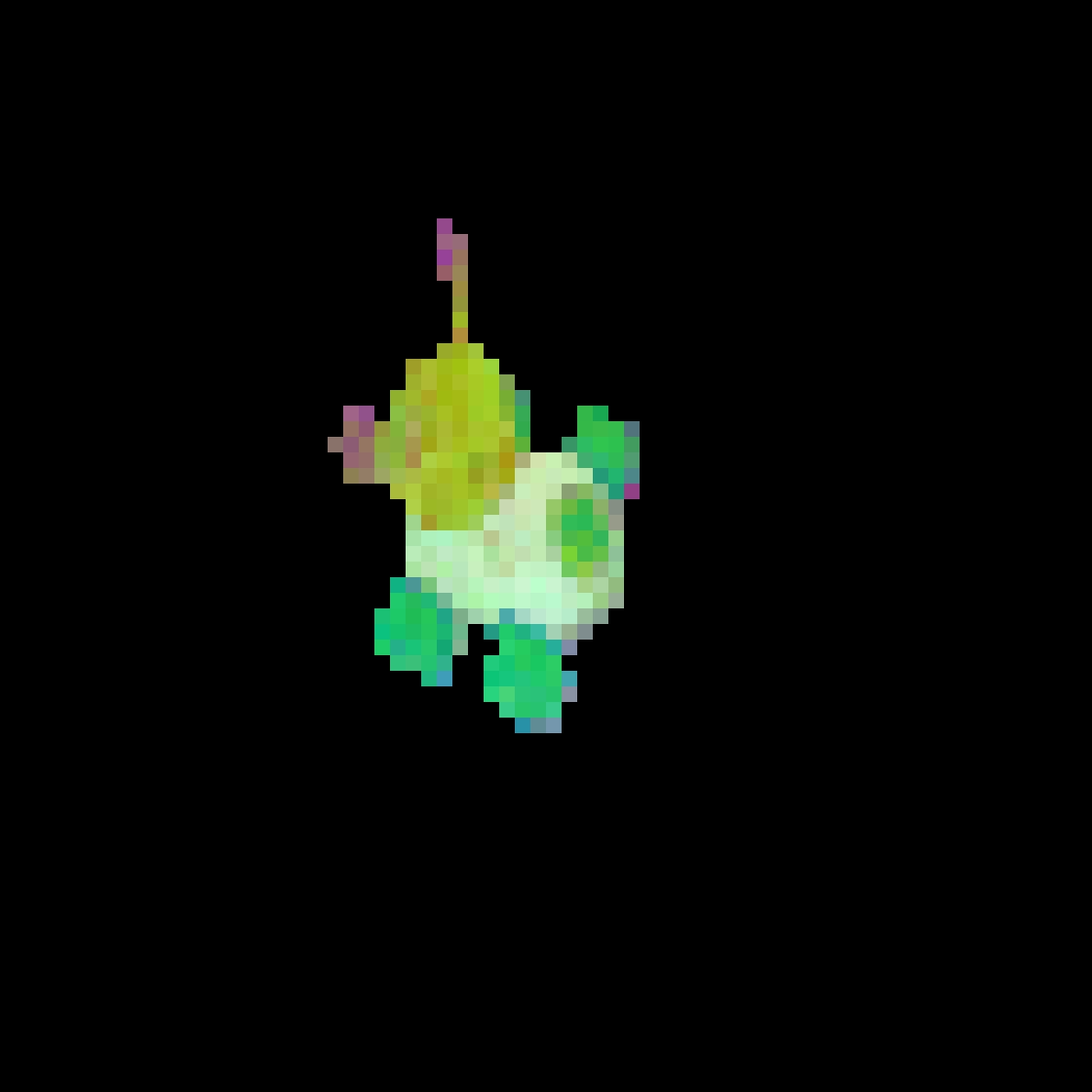}
        \end{subfigure}
        \hfill
        \begin{subfigure}[b]{0.24\linewidth}
            \centering
            \includegraphics[width=\textwidth]{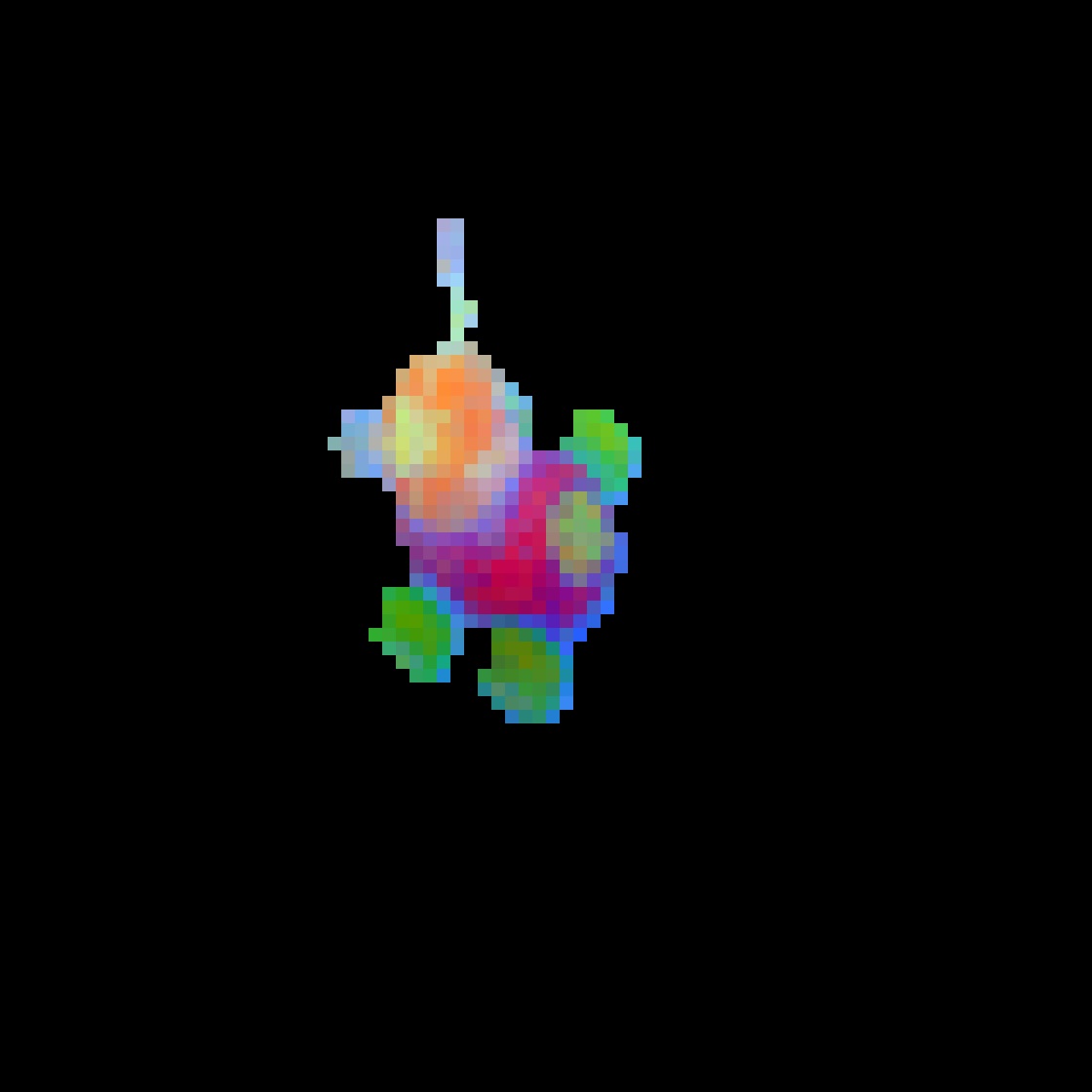}
        \end{subfigure}
        \vspace{0.1cm}

    \end{minipage}
    
    \caption{PCA visualization of extracted features. DINOcular demonstrates smoother, cross-view consistent features and clear semantic part separation.}
    \label{fig:pca_qual}
\end{figure}

This robust 3D awareness is further reflected in the feature representations. Figure~\ref{fig:pca_qual} visualizes the three strongest PCA components of the extracted features. DINOcular yields distinct, structurally aligned feature maps that remain highly stable across varying environments and viewpoints. It effectively achieves clear semantic part separation (e.g., distinguishing the horns, body, and legs of the object) while overcoming the noisy, pixelated artifacts prevalent in the feature spaces of both DINO and DUNE.

\begin{figure}[h]
    \centering
    \begin{subfigure}[b]{0.19\textwidth}
        \centering
        \textbf{Source}\\[1ex]
        \includegraphics[width=\textwidth]{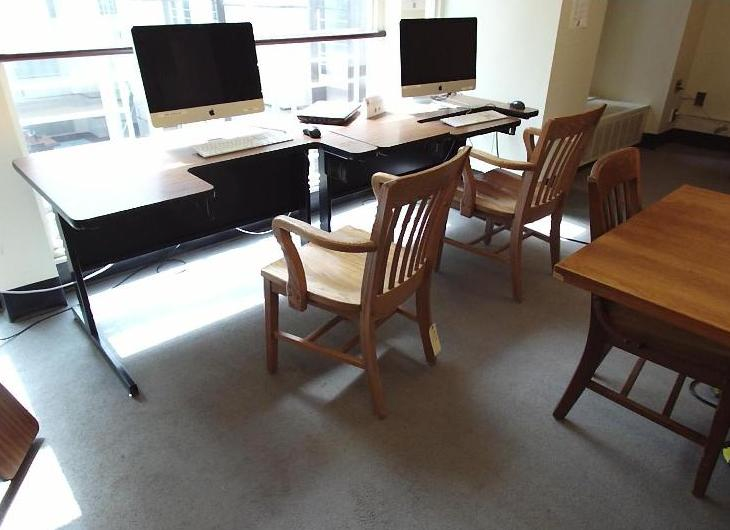}
    \end{subfigure}
    \hfill
    \begin{subfigure}[b]{0.19\textwidth}
        \centering
        \textbf{GT}\\[1ex]
        \includegraphics[width=\textwidth]{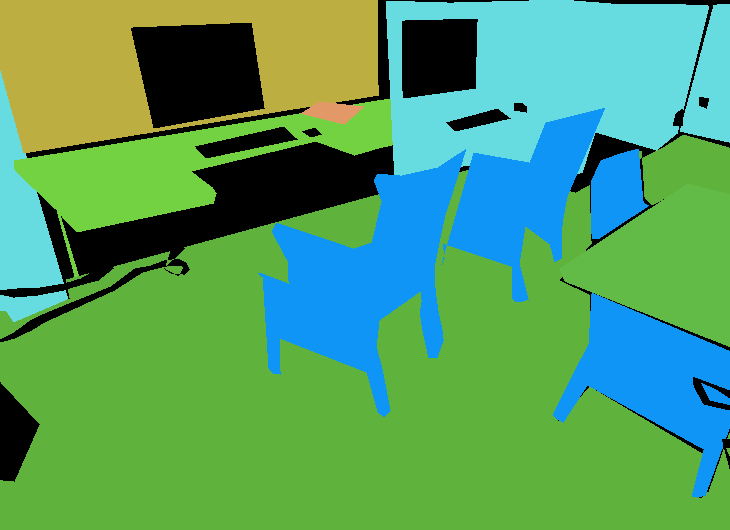}
    \end{subfigure}
    \hfill
    \begin{subfigure}[b]{0.19\textwidth}
        \centering
        \textbf{DINOcular}\\[1ex]
        \includegraphics[width=\textwidth]{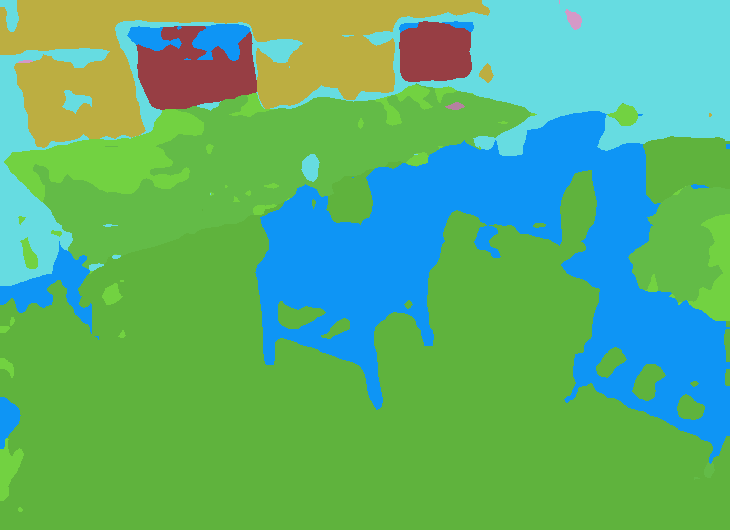}
    \end{subfigure}
    \hfill
    \begin{subfigure}[b]{0.19\textwidth}
        \centering
        \textbf{DFormerv2}\\[1ex]
        \includegraphics[width=\textwidth]{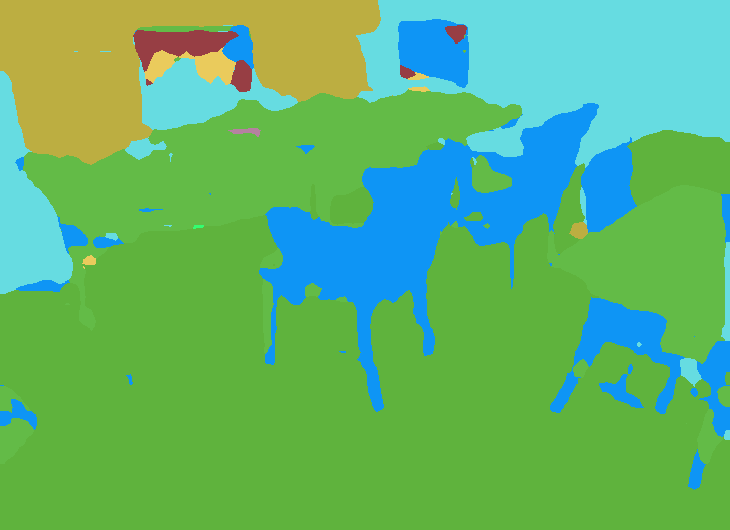}
    \end{subfigure}
    \hfill
    \begin{subfigure}[b]{0.19\textwidth}
        \centering
        \textbf{MultiMAE}\\[1ex]
        \includegraphics[width=\textwidth]{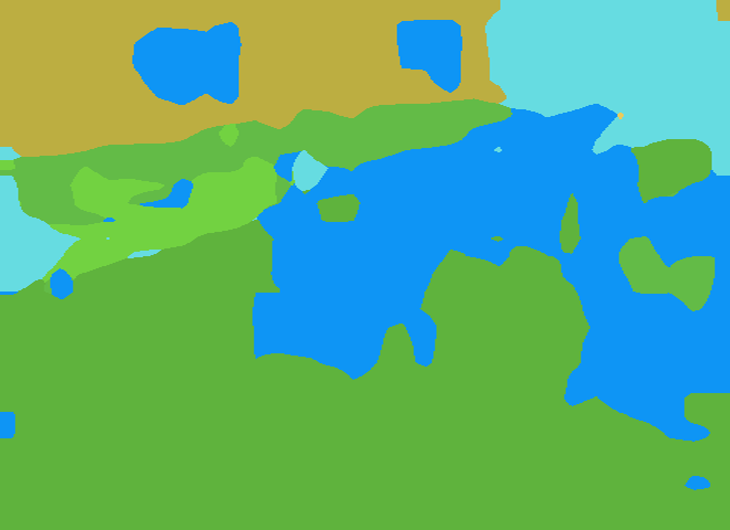}
    \end{subfigure}
    \vspace{0.2cm}

    \begin{subfigure}[b]{0.19\textwidth}
        \centering
        \includegraphics[width=\textwidth]{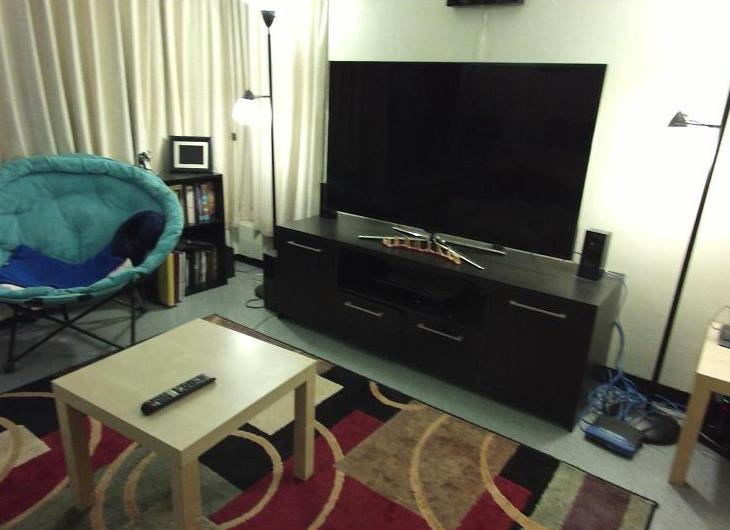}
    \end{subfigure}
    \hfill
    \begin{subfigure}[b]{0.19\textwidth}
        \centering
        \includegraphics[width=\textwidth]{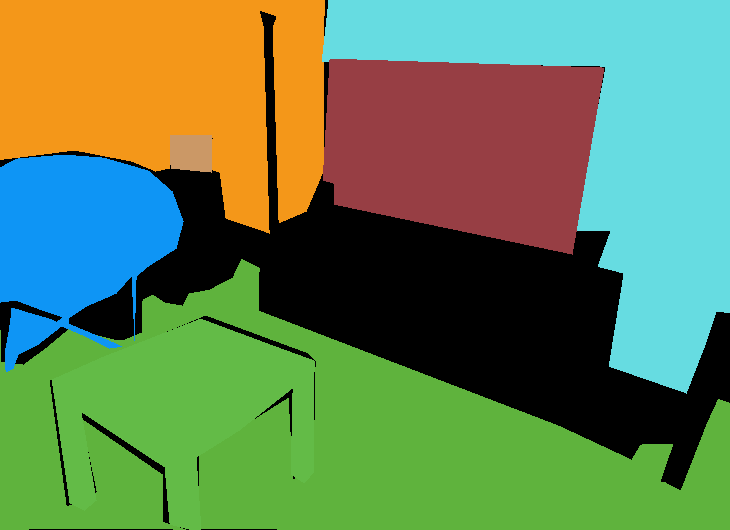}
    \end{subfigure}
    \hfill
    \begin{subfigure}[b]{0.19\textwidth}
        \centering
        \includegraphics[width=\textwidth]{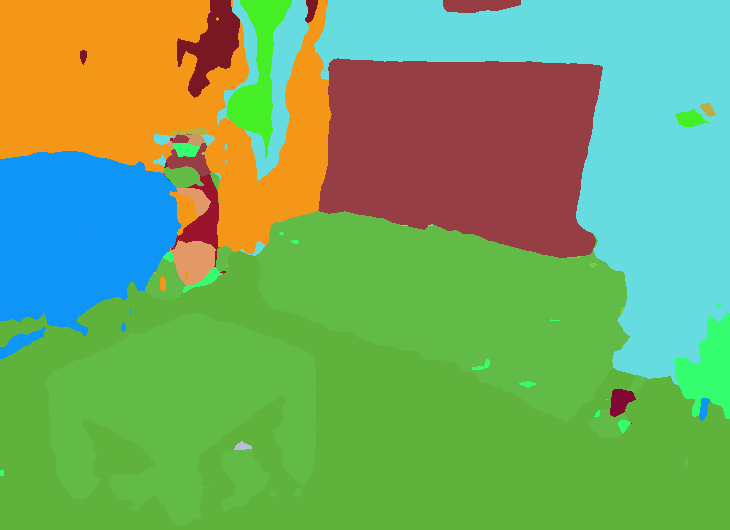}
    \end{subfigure}
    \hfill
    \begin{subfigure}[b]{0.19\textwidth}
        \centering
        \includegraphics[width=\textwidth]{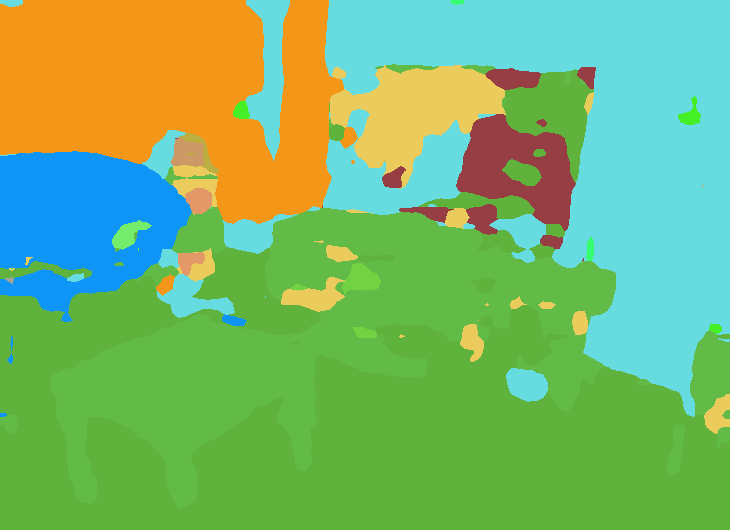}
    \end{subfigure}
    \hfill
    \begin{subfigure}[b]{0.19\textwidth}
        \centering
        \includegraphics[width=\textwidth]{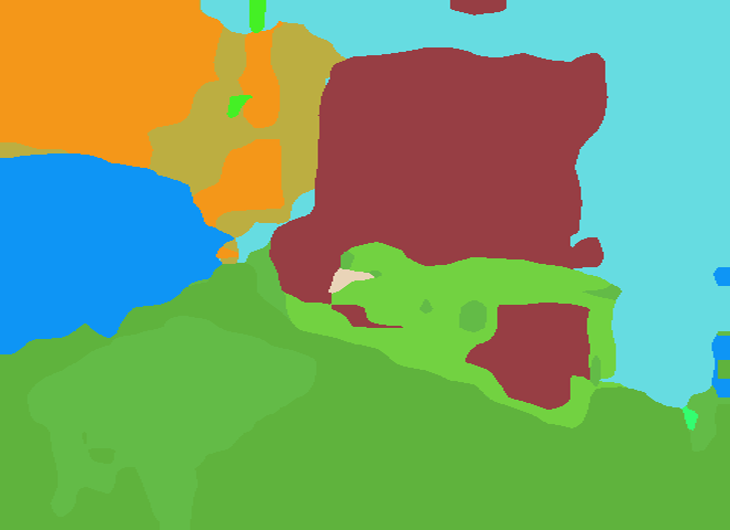}
    \end{subfigure}
    \vspace{0.2cm}

    \begin{subfigure}[b]{0.19\textwidth}
        \centering
        \includegraphics[width=\textwidth]{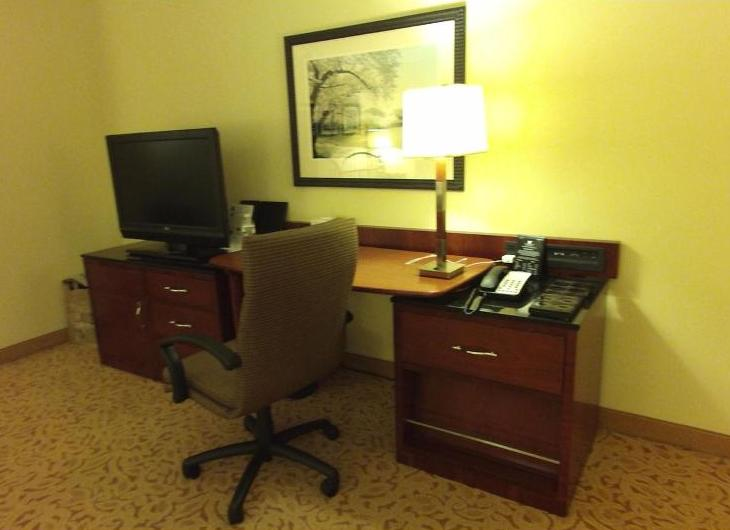}
    \end{subfigure}
    \hfill
    \begin{subfigure}[b]{0.19\textwidth}
        \centering
        \includegraphics[width=\textwidth]{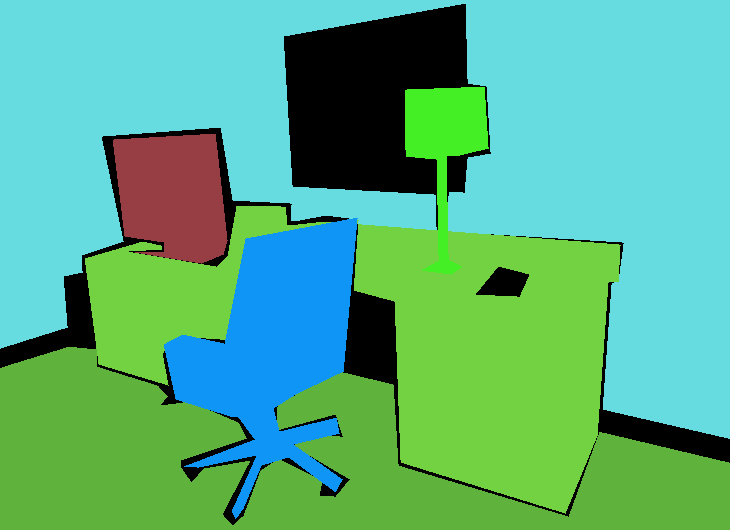}
    \end{subfigure}
    \hfill
    \begin{subfigure}[b]{0.19\textwidth}
        \centering
        \includegraphics[width=\textwidth]{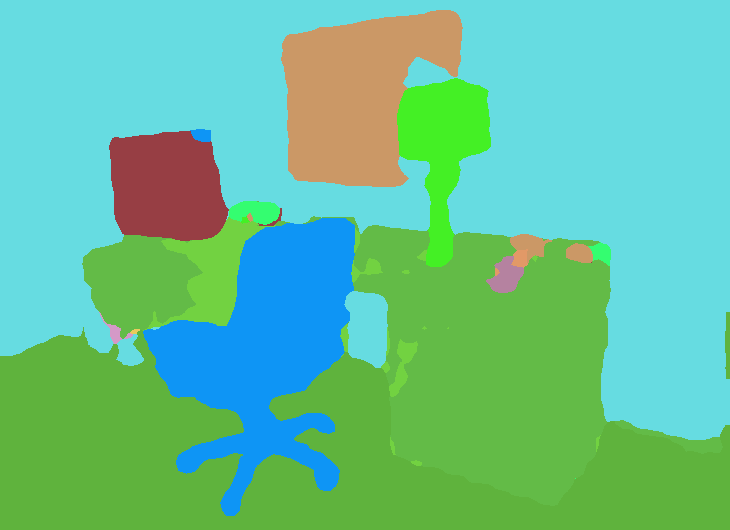}
    \end{subfigure}
    \hfill
    \begin{subfigure}[b]{0.19\textwidth}
        \centering
        \includegraphics[width=\textwidth]{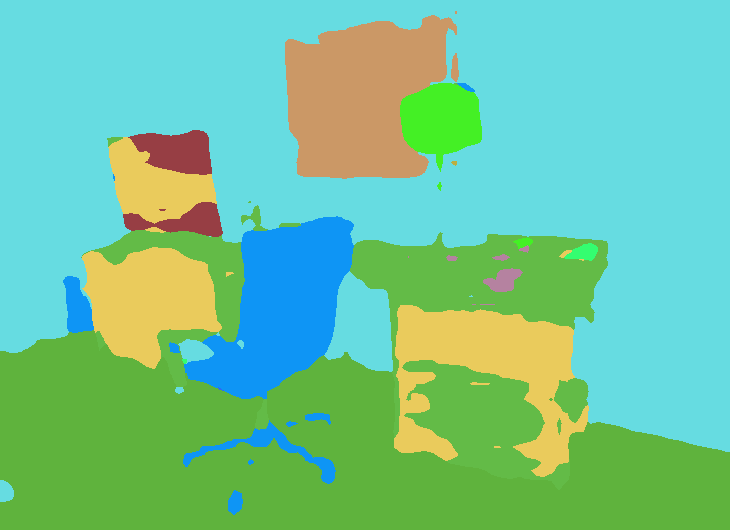}
    \end{subfigure}
    \hfill
    \begin{subfigure}[b]{0.19\textwidth}
        \centering
        \includegraphics[width=\textwidth]{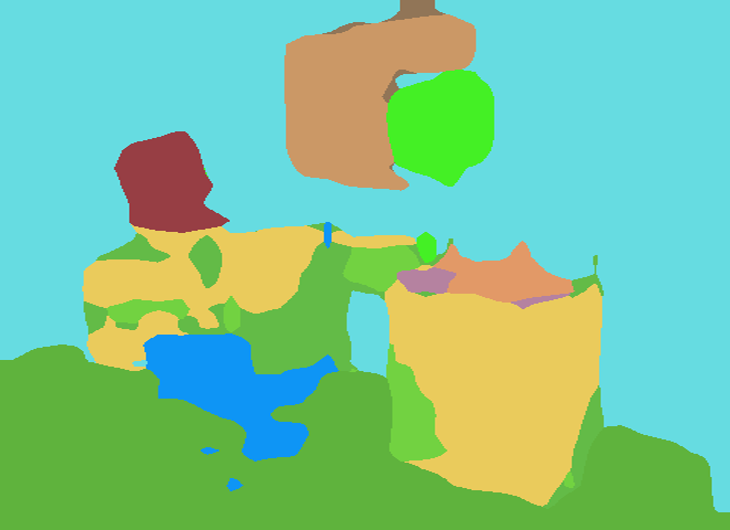}
    \end{subfigure}
    \caption{Qualitative semantic segmentation comparison. DINOcular yields highly consistent semantic masks and sharper object boundaries compared to DFormerv2 and MultiMAE.}
    \label{fig:segm_comp}
\end{figure}

\textbf{Semantic Segmentation.}
We extend our qualitative analysis to 2D semantic segmentation on complex indoor scenes shown in Figure~\ref{fig:segm_comp}. When compared to DFormerv2~\citep{yin2025dformerv2} and MultiMAE~\citep{bachmann2022multimae}, which exhibit severe spatial fragmentation and noisy patch assignments (particularly on large textured surfaces like floors and furniture), DINOcular generates clean, contiguous semantic regions with reliably delineated sharp object boundaries.

\begin{figure}[h]
    \centering
    \includegraphics[width=\textwidth]{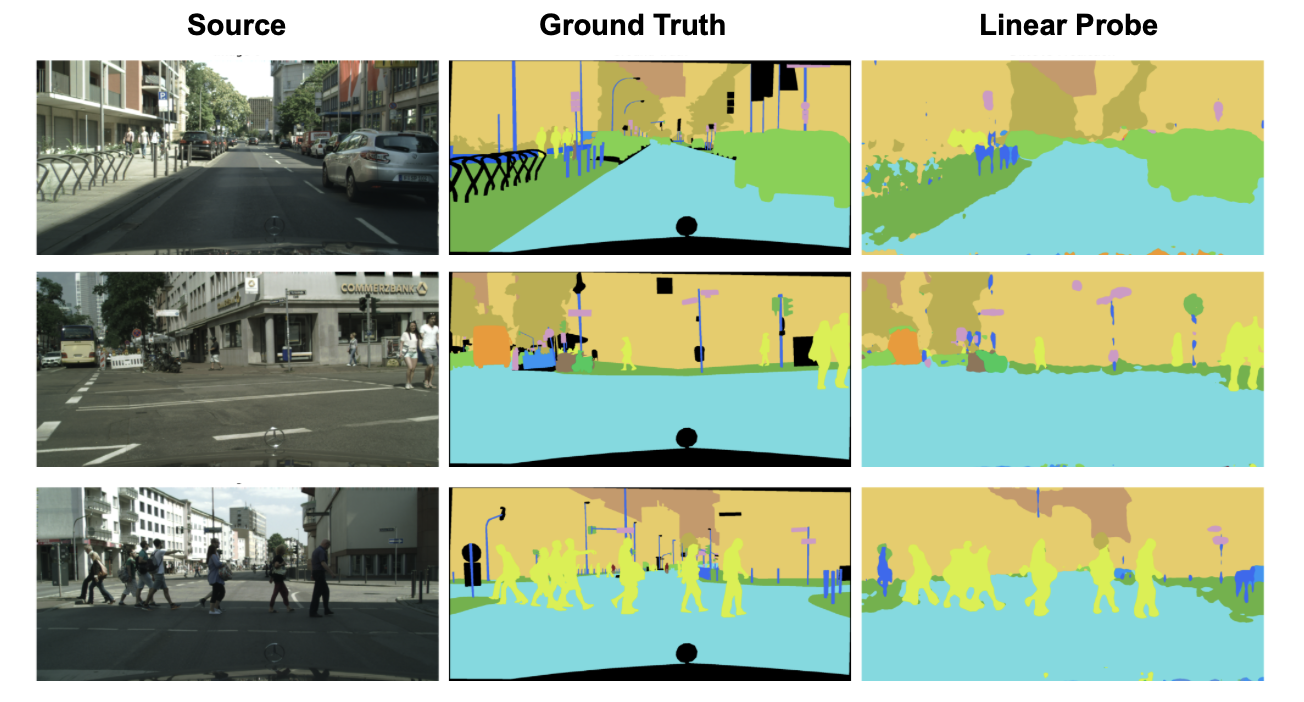}
    \caption{Qualitative semantic segmentation results on CityScapes outdoor scenes for DINOcular.
    The model reliably captures spatial layout and remains robust to lighting changes, e.g. shadows.}
    \label{fig:city}
\end{figure}

Finally, we visualize DINOcular's semantic segmentation performance on complex outdoor driving scenes in Figure~\ref{fig:city}. Model demonstrates successful generalization for outdoor urban layouts, correctly identifying primary categories. Notably, DINOcular exhibits strong robustness to challenging, real-world illumination shifts; it maintains coherent structural parsing even in the presence of lighting changes, e.g. shadows.

\section{Code}
We provide our code in project page.

\section{Training Details}\label{app:training}
By default we train all models on 4 nodes with each 4 A100 GPUs.
The details are shown in Table~\ref{tab:hyperparameters}
\begin{table}[ht]
  \centering\small
  \caption{Hyperparameters and training details used for pre-training. Highlighted rows indicate difference with model size.}
  \label{tab:hyperparameters}
  \resizebox{0.8\textwidth}{!}{%
  \begin{tabular}{lc}
    \toprule
    \textbf{Parameter} & \textbf{Value} \\
    \midrule
    \multicolumn{2}{c}{\textit{Architecture \& Data}} \\
    \midrule
    \rowcolor{lightgray} Base Architecture & DINOcular\_S/\_L \\
    Projection dimension & 65,536 \\
    Mixed precision & BF16 \\
    \rowcolor{lightgray} Batch size per GPU & 96/46 \\
    Total epochs & 150 \\
    \midrule
    \multicolumn{2}{c}{\textit{Optimization}} \\
    \midrule
    Optimizer & AdamW \\
    Peak learning rate & 0.004 \\
    Learning rate scaling & $\text{lr} \times \sqrt{\text{batch\_size}/1024}$ \\
    Min learning rate & $1 \times 10^{-6}$ \\
    Warmup epochs & 15 \\
    Weight decay & 0.04 $\rightarrow$ 0.4 \\
    Gradient clipping norm & 3.0 \\
    Freeze last layer & 1 epoch \\
    \rowcolor{lightgray}Stochastic depth (Drop path) rate & 0.1/0.2 \\
    \midrule
    \multicolumn{2}{c}{\textit{Self-Supervised Objectives}} \\
    \midrule
    \rowcolor{lightgray}Teacher momentum & 0.992/0.996 $\rightarrow$ 1.0 \\
    Teacher temperature & 0.04 $\rightarrow$ 0.07 \\
    Teacher temperature warmup & 15 epochs \\
    iBOT student temperature & 0.1 \\
    iBOT patch mask ratio & [0.1, 0.5] \\
    iBOT mask probability & 0.5 \\
    DINO loss weight & 1.0 \\
    iBOT patch loss weight & 1.0 \\
    KoLeo entropy loss weight & 0.1 \\
    Equivariance loss weight & 10.0 \\
    \midrule
    \multicolumn{2}{c}{\textit{Multi-Crop Augmentation}} \\
    \midrule
    Global crops number & 2 \\
    Global crops scale & (0.32, 1.0) \\
    Local crops number & 8 \\
    Local crops scale & (0.05, 0.32) \\
    \bottomrule
  \end{tabular}%
  }
\end{table}

\end{document}

%% file: tables/mlp_enc.tex
\begin{table}[t]
    \centering
    \caption{\textbf{Ablation on geometry encoding.} We report RGB-D semantic segmentation probing (Segm.) on NYUDepthv2~\protect\citep{silberman2012indoor} (mIoU) and correspondence estimation with viewpoint change on NAVI~\protect\citep{jampani2023navi} ($\theta_{90}^{120}$) and ScanNet ($\theta_{60}^{180}$) following Probe3D~\protect\citep{el2024probing}.$^*$ w/o SAM3 object mask sampling.}
    \label{tab:ablation_enc}
    \resizebox{\textwidth}{!}{%
    \begin{tabular}{@{} llllc rrr @{}}
    \toprule
    \begin{tabular}{@{}l@{}}Geometric\\Attention\end{tabular} & 
    \begin{tabular}{@{}l@{}}Intra-Patch\\Geometry\end{tabular} & 
    Objective & 
    Dataset & 
    \begin{tabular}{@{}c@{}}Depth\\Dropout\end{tabular} & 
    \begin{tabular}{@{}r@{}}Segm.\\\scriptsize{[mIoU]}\end{tabular} & 
    \begin{tabular}{@{}r@{}}NAVI$\theta_{90}^{120}$\\\scriptsize{[R]}\end{tabular} & 
    \begin{tabular}{@{}r@{}}ScanNet $\theta_{60}^{180}$\\\scriptsize{[R]}\end{tabular} \\
    \midrule
    
    Proximity~\citep{yin2025dformerv2} & -- & Supervised & IN1k & -- & 24.86 & 17.40 & 7.30 \\
    Proximity~\citep{yin2025dformerv2} & -- & DINO       & IN1k & -- & 25.78 & 16.40 & 4.50 \\
    \midrule
    
    2D RoPE & --        & DINO & IN1k & -- & 33.50 & 20.71 & 11.35 \\
    2D RoPE & Depth MLP & DINO+iBOT+MVI & IN1k+MVI2.0 & Yes & 35.54 & 23.25 & 14.58 \\
    \midrule
    
    3D RoPE & --            & DINO & IN1k & -- & 39.92 & 22.74 & 12.00 \\
    3D RoPE & --            & DINO+iBOT+MVI & IN1k+MVI2.0 & -- & 40.92 & 23.15 & 11.73 \\
    3D RoPE & Surf. Normals & DINO & IN1k & -- & 38.38 & 12.80 & 5.80 \\
    \midrule
    
    3D RoPE & Depth MLP & DINO          & IN1k        & No  & 40.16 & 17.90 & 10.50 \\
    3D RoPE & Depth MLP & MVI           & IN1k+MVI2.0 & No  & 28.54 & 19.99 & 10.80 \\
    3D RoPE & Depth MLP & DINO+MVI      & IN1k+MVI2.0 & No  & 31.20 & 20.58 & 11.40 \\
    3D RoPE & Depth MLP & DINO+MVI      & IN1k+MVI2.0 & Yes & 31.92 & 19.87 & 13.60 \\
    3D RoPE & Depth MLP & DINO+iBOT     & IN1k+MVI2.0 & No  & 34.87 & 15.55 & 4.88 \\
    3D RoPE & Depth MLP & DINO+iBOT     & IN1k        & No  & 40.45 & 16.52 & 8.40 \\
    3D RoPE & Depth MLP & DINO+iBOT+MVI$^*$ & IN1k+MVI2.0 & Yes & 39.89 & 22.50 & 18.80 \\
    3D RoPE & Depth MLP & DINO+iBOT+MVI & IN1k+MVI2.0 & No  & 38.58 & 25.29 & 13.19 \\
    3D RoPE & Depth MLP & DINO+iBOT+MVI & IN1k+MVI2.0 & Yes & 40.27 & 25.25 & 16.52 \\
    \bottomrule
    \end{tabular}%
    }
\end{table}

%% file: tables/loss_abl.tex
\begin{table}[t]
    \centering
    \caption{\textbf{Ablation of the multi-view loss objective.} We finetune a baseline model that was trained with the DINO objective with different multi-view objectives. We report RGB-D semantic segmentation probing (Segm.) on NYUDepthv2~\protect\citep{silberman2012indoor} (mIoU) and correspondence estimation with viewpoint change (3D Corresp.) on NAVI~\protect\citep{jampani2023navi} ($\theta_{90}^{120}$) following Probe3D~\protect\citep{el2024probing}.}
    \resizebox{.7\textwidth}{!}{%
    \begin{tabular}{lrr}
        \toprule
\textbf{Finetuning Objective} & Segm. \scriptsize{[mIoU]} & 3D Corresp. \scriptsize{[R]} \\
        \midrule
        Baseline (trained w/ DINO objective) & 40.16 & 17.90 \\
        \midrule
        MV-Cosine & 11.10  & 11.41 \\
        DINO + MV-Cosine & 24.40 & 7.45 \\
        \midrule
        MV-iBot    & 30.30  & 15.85 \\
        DINO + MV-iBot   & 34.00  & 16.86 \\
        \midrule
        MV-Contrastive       & 28.54 & 19.99 \\
        DINO + MV-Contrastive  & 31.20  & 20.58\\
        \bottomrule
    \end{tabular}%
    }
    \label{tab:ablation_multiview}
\end{table}

%% file: tables/main.tex
\newcommand{\best}[1]{\colorbox{blue!15}{\textbf{#1}}\hspace*{-3pt}}
\newcommand{\snd}[1]{\colorbox{blue!10}{#1}\hspace*{-3pt}}

\begin{table}[t]
    \centering
    \caption{\textbf{RGB-D semantic segmentation} results (mIOU) using linear probing on the respective features. All results are reported in [\% mIoU]. \dag\ marks models trained with supervised objectives, all other methods are self-supervised. \depth marks models with RGB and depth as input, \nodepth only with RGB. $^\star$Sonata is not intended to be tested on these evaluation protocols.}
    \label{table:segm}
    \resizebox{\textwidth}{!}{%
    \begin{tabular}{lrrrrrr}
        \toprule
        \textbf{Method} & \textbf{Param.} & \textbf{Data} &\textbf{ADE20k} & \textbf{NYUDepthv2} & \textbf{SUN RGB-D} & \textbf{Cityscapes} \\ 
        \midrule
        \multicolumn{7}{l}{\emph{SoTA ViT-B / PTv3 models}}\\
        \nodepth DINOv2 (ViT-B) & \SI{86}{M} & \SI{142}{M} & 47.30 & 56.04 & 52.26 & 69.40\\
        \nodepth DINOv3 (ViT-B) & \SI{86}{M} & \SI{1689}{M} & 51.80 & 60.78 & 54.28 & 71.53\\
        \nodepth DUNE (ViT-B) & \SI{85}{M} & \SI{21}{M} (dist.) & 44.90 & 68.20 & 50.22 & 70.60\\
        \depth Sonata$^\star$ (PTv3) & \SI{108}{M} & \SI{140}{k} (pcd) & 11.53 & 26.62 & 27.83 & - \\
        \midrule
        \multicolumn{7}{l}{\emph{ViT-B / RMT-L models at comparable data size}}\\
        \nodepth DINO (ViT-B) & \SI{85}{M} & \SI{1.3}{M} & 31.80 & 34.49 & 37.13 & 56.90\\
        \depth MultiMAE (ViT-B) & \SI{85}{M} & \SI{1.3}{M} & 17.37 & \snd{39.29} & 31.88 & 28.63\\
        \depth DFormerv2-L\dag & \SI{95.5}{M} & \SI{1.3}{M} & \snd{33.84} & 38.29 & \snd{37.69} & \best{60.25}\\
        \nodepth DINOv2 (ViT-B repr., only ImgNet) & \SI{85}{M} & \SI{1.3}{M} & 29.83 & 36.16 & 35.90 & 52.14\\
        \depth DINOcular-L \emph{(Ours)} & \SI{94}{M} & \SI{3.9}{M} & \best{40.23} & \best{47.46} & \best{42.25} & \snd{59.19}\\
        \midrule
        \multicolumn{7}{l}{\emph{ViT-S / RMT-S models at comparable data size}}\\
        \nodepth DINO (ViT-S) & \SI{21}{M} & \SI{1.3}{M} & 22.82 & 35.97 & 35.35 & 45.84\\
        \depth DFormerv2-S\dag & \SI{27}{M} & \SI{1.3}{M} & 30.60 & 24.86 & 29.34 & 51.41\\
        \nodepth DINOv2 (ViT-S repr., only ImgNet) & \SI{21}{M} & \SI{1.3}{M} & 24.22 & 33.43 & 33.91 & 47.03\\
        \nodepth DINOv2 (ViT-S repr.) & \SI{21}{M} & \SI{3.9}{M} & 20.64 & 29.74 & 31.67 & 44.14\\
        \nodepth Ours w/ RGB-only & \SI{27}{M} & \SI{3.9}{M} & 18.47 & 28.15 & 28.98 & 39.85\\
        \depth Ours w/o multi-view & \SI{27}{M} & \SI{1.3}{M} & \best{36.02} & \best{40.45} & \snd{39.19} & \best{56.31}\\
        \depth DINOcular-S \emph{(Ours)} & \SI{27}{M} & \SI{3.9}{M} & \snd{33.88} & \snd{40.27} & \best{39.99} & \snd{55.35}\\
        \bottomrule
    \end{tabular}%
    }
\end{table}

\vspace{2pt}

\begin{table}[t]
    \centering
    \caption{\textbf{Geometric Tasks:} We evaluate features through the Probe3D protocol on 3D Correspondence Estimation (NAVI and ScanNet) over different angles between viewpoints, as well as through one-shot Object Pose Estimation. To analyze correlation of feature information with the depth input, we report results on linear probing for depth regression/reconstruction (NYU Depth v2). \dag\ marks models trained with supervised objectives, all other methods are self-supervised. \depth marks models with RGB and depth as input, \nodepth only with RGB. $^\star$Sonata is not intended to be tested on these evaluation protocols.}
    \label{table:3dcorr}
    \resizebox{\textwidth}{!}{%
    \begin{tabular}{lrrrrrrrrrrrrr}
        \toprule
        & \multicolumn{4}{c}{\textbf{NAVI} \scriptsize{[\% R]}} & &\multicolumn{4}{c}{\textbf{ScanNet}  \scriptsize{[\% R]}} & \multicolumn{3}{c}{\textbf{OnePose-LowTex} \scriptsize{[\% R]}} & \textbf{NYUDv2}\\
        \cmidrule(lr){2-5} \cmidrule(lr){7-10} \cmidrule(lr){11-13}
        \textbf{Model} & $\theta_{0}^{30}$ & $\theta_{30}^{60}$ & $\theta_{60}^{90}$ & $\theta_{90}^{120}$ & & $\theta_{0}^{15}$ & $\theta_{15}^{30}$ & $\theta_{30}^{60}$ & $\theta_{60}^{180}$ & $\sfrac{\SI{1}{cm}}{\SI{1}{^\circ}}$ & $\sfrac{\SI{3}{cm}}{\SI{3}{^\circ}}$ & $\sfrac{\SI{5}{cm}}{\SI{5}{^\circ}}$ & \scriptsize{[m RMSE]} $\downarrow$\\
        \midrule
        \multicolumn{14}{l}{\emph{SoTA ViT-B / PTv3 models}}\\
        \nodepth DINOv2 (ViT-B) & 92.3 & 67.9 & 45.4 & 32.0 & & 37.0 & 27.5 & 19.7 & 11.2 & 10 & 51 & 70 & .39\\
        \nodepth DINOv3 (ViT-B) & 96.3 & 71.8 & 41.9 & 25.4 & & 54.9 & 39.2 & 26.6 & 15.7 & 4 & 32 & 54 & .38\\
        \nodepth DUNE (ViT-B) & 94.8 & 60.4 & 29.4 & 17.2 & & 34.6 & 25.9 & 17.9 & 7.6 & 10 & 50 & 66 & .36\\
        \depth Sonata$^\star$ (PTv3) & 73.9 & 41.4 & 27.5 & 23.8 & & 37.3 & 24.0 & 11.4 & 4.2 & 0 & 3 & 8 & -\\
        \midrule
        \multicolumn{14}{l}{\emph{ViT-B / RMT-L models at comparable data size}}\\
        \nodepth DINO (ViT-B) & \snd{89.0} & \snd{54.0} & \snd{30.7} & 21.0 & & 45.0 & 34.4 & 22.6 & 10.7 & 6 & 34 & 50 & .64\\
        \depth MultiMAE (ViT-B) & 85.1 & 39.3 & 18.7 & 11.1 & & 7.5 & 6.9 & 6.1 & 2.5 & 3 & 7 & 18 & .19\\
        \depth DFormerv2-L\dag & 74.7 & 42.9 & 26.0 & 19.4 & & 47.0 & 38.0 & 24.4 & \snd{11.0} & \snd{8} & 32 & 43 & .72\\
        \nodepth DINOv2 (ViT-B repr., only ImgNet) & \best{93.1} & \best{57.4} & \best{31.2} & \snd{19.5} & & \snd{51.5} & \snd{40.2} & \snd{24.8} & 9.8 & 7 & \snd{40} & \snd{57} & .63\\
        \depth DINOcular-L \emph{(Ours)} & 88.2 & 52.7 & 30.3 & \best{23.2} & & \best{62.7} & \best{52.2} & \best{32.7} & \best{13.1} & \best{10} & \best{46} & \best{63} & .26\\
        \midrule
        \multicolumn{14}{l}{\emph{ViT-S / RMT-S models at comparable data size}}\\
        \nodepth DINO (ViT-S) & 86.7 & 52.9 & 31.9 & 21.3 & & 41.5 & 30.9 & 19.0 & 9.1 & 6 & 32 & 47 & .70\\
        \depth DFormerv2-S\dag & 80.0 & 41.4 & 24.0 & 17.4 & & 37.2 & 26.9 & 16.8 & 7.3 & \best{9} & 32 & 44 & .70\\
        \nodepth DINOv2 (ViT-S repr., only ImgNet) & \snd{88.1} & 54.3 & 29.4 & 19.9 & & \snd{50.8} & 39.8 & 24.5 & 9.9 & \snd{7} & \best{40} & \best{56} & .70\\
        \nodepth DINOv2 (ViT-S repr.) & 84.3 & 52.3 & 29.5 & 17.2 & & 51.4 & \snd{40.4} & \snd{25.7} & 11.7 & 6 & 30 & 46 & .73\\
        \nodepth Ours w/ RGB-only & 81.2 & \best{58.5} & \best{37.0} & \snd{23.4} & & 47.0 & 33.5 & 22.5 & \snd{11.9} & 6 & 33 & 48 & .72\\
        \depth Ours w/o multi-view & 87.5 & 49.3 & 25.8 & 16.5 & & 44.0 & 35.4 & 22.0 & 8.4 & \snd{7} & 35 & 51 & .26\\
        \depth DINOcular-S \emph{(Ours)} & \best{89.1} & \snd{57.3} & \snd{33.9} & \best{25.3} & & \best{58.3} & \best{46.4} & \best{30.3} & \best{16.5} & 6 & \snd{37} & \snd{55} & .24\\
        \bottomrule
    \end{tabular}%
    }
\end{table}